\documentclass[]{servicenow}

\usepackage[utf8]{inputenc}
\usepackage{amsfonts}
\usepackage{nicefrac}
\usepackage{enumitem}
\usepackage{amsmath}
\usepackage{listings}
\usepackage{multirow}
\usepackage{tabularx}
\usepackage{makecell}
\usepackage{pifont}
\usepackage{adjustbox}
\usepackage{float}
\usepackage{etoolbox}
\usepackage{tikz}
\usepackage[cachedir=_minted]{minted}

\newif\ifshowtodo
\showtodotrue

\ifshowtodo
  \newcommand{\TODO}[1]{\textcolor{red}{\textbf{TODO:} #1}}
\else
  \newcommand{\TODO}[1]{}
\fi

\definecolor{myteal}{HTML}{62D4C5}
\definecolor{myblue}{HTML}{3A86FF}
\definecolor{mydarkblue}{rgb}{0,0.08,0.45}

\definecolor{myblue}{RGB}{147,204,255}
\definecolor{mylightblue}{HTML}{E6F0FA}
\definecolor{stagegreen}{HTML}{488D8A}
\definecolor{stagegreenbg}{HTML}{E4F0EF}

\tcbuselibrary{minted,skins, breakable}

\DeclareRobustCommand{\stagecircle}[1]{\tikz[baseline=-0.55ex]{
    \node[
      circle,
      fill=stagegreen,
      text=white,
      font=\bfseries\scriptsize,
      inner sep=0pt,
      minimum size=1.em
    ] {#1};
  }}

\tcbset{
  enhanced,
  colback=white!200!black,
  colframe=stagegreenbg,
  colbacktitle=stagegreenbg,
  title filled,
  coltitle=stagegreen,
  fonttitle=\bfseries,
  arc=3mm,
  outer arc=3mm,
  boxrule=0.5mm,
  toprule=0.5mm,
  bottomrule=0.5mm,
  titlerule=0.5mm,
  drop fuzzy shadow,
}
\setlist[itemize]{topsep=1pt, itemsep=0.5pt}

\hypersetup{
  colorlinks,
  linkcolor=stagegreen,
  citecolor=stagegreen,
  urlcolor=stagegreen
}

\newcommand{\paragraphtight}[1]{\par\vspace{-0.75\parskip}\vspace{0.3em}\noindent\textbf{#1}\hspace{0.4em}\ignorespaces
}

\newcommand{\rev}[1]{\textcolor{black}{#1}}

\renewcommand{\email}[1]{\href{mailto:#1}{#1}}
\DeclareRobustCommand{\titlelinkicon}[1]{\makebox[1.2em][c]{\raisebox{-0.1em}{#1}}\hspace{0.4em}}
\renewcommand{\snowdataformat}[2][]{{\small\titlelinkicon{#1}#2}}
\DeclareRobustCommand{\huggingfaceicon}{\raisebox{-0.05em}{\includegraphics[height=1em]{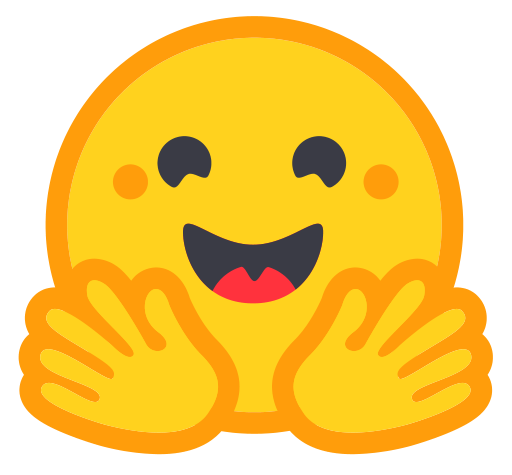}}}
\renewcommand{\correspondence}[1]{\snowdata[\textcolor{stagegreen}{\faEnvelopeSquare}]{#1}}
\renewcommand{\github}[1]{\snowdata[\textcolor{black}{\faGithub}]{#1}}

\title{Dr-CiK: A Benchmark testing \\Deep Research for Context-Aided Forecasting}

\author[1,2,3]{Yihong Tang}
\author[2,3,4]{Andrew Robert Williams}
\author[2,3,4]{Arjun Ashok}
\author[1]{Vincent Zhihao Zheng}
\author[2,3,5]{Shubham Gupta}
\author[1]{Lijun Sun}
\author[2,3,5]{Alexandre Drouin}
\author[2,6]{Issam H.~Laradji}
\author[2]{\'Etienne Marcotte}
\author[2,5]{Valentina Zantedeschi}

\affiliation[1]{McGill University}
\affiliation[2]{ServiceNow Research}
\affiliation[3]{Mila -- Quebec AI Institute}
\affiliation[4]{Universit\'e de Montr\'eal}
\affiliation[5]{Universit\'e Laval}
\affiliation[6]{University of British Columbia}

\github{\href{https://github.com/ServiceNow/Dr-CiK}{github.com/ServiceNow/Dr-CiK}\quad\titlelinkicon{\huggingfaceicon}\href{https://huggingface.co/datasets/ServiceNow/Dr-CiK}{huggingface.co/datasets/ServiceNow/Dr-CiK}}
\correspondence{Valentina Zantedeschi (\email{vzantedeschi@gmail.com})}

\abstract{
Time series forecasting in real-world settings often depends not only on historical observations, but also on external context that must be actively discovered from noisy, heterogeneous information sources.
Yet existing context-aided forecasting benchmarks typically assume that the supporting context is already provided, leaving open whether agents can identify it on their own.
Therefore, we introduce Dr-CiK, a benchmark for evaluating whether agents can retrieve forecasting-relevant supporting context from a document corpus, filter out distractors, distill the retrieved context into forecast-useful evidence, and generate forecasts supported by that evidence. Through context ablations and evaluations of state-of-the-art deep research and forecasting methods paired together, we show that high-quality context substantially improves forecasting performance in Dr-CiK. However, existing DR methods recover limited ground-truth supporting evidence, are frequently misled by distractors, and can cause forecasters to perform worse with retrieved context than without context. Our results motivate research on agents that search for the right context to predict the future.
}

\begin{document}

\maketitle

\section{Introduction}

Traditional time-series forecasters predict future values by extrapolating trend and seasonality from historical observations~\citep{hyndman2018forecasting,box2015time}.
Recent deep learning models~\citep{lim2021time,chen2023long} and time-series foundation models~\citep{rasul2023lag,garza2023timegpt,ansari2024chronos,woo2024unified,das2023decoder} have widened applicability, but they still largely operate on numerical history alone.
In real-world settings, this history-only paradigm is insufficient when future values are shaped by exogenous factors.
For example, anticipating a demand spike during a heat wave, a drop in ridership during a transit strike, or a shift in disease incidence after a policy change requires context beyond the time series itself.

A growing body of work has explored \emph{context-aided forecasting} (CAF), where models condition forecasts on textual context alongside numerical history~\citep{xue2023promptcast,gruver2024large,liu2024time,xu2024beyond,emami2024syscaps,merrill2024language,kim2024multi,wu2025dualforecaster,liu2026rethinking}.
These methods typically use Large Language Models (LLMs), alone~\citep{requeima2024llm,jin2024timellm,williams2024context} or in multimodal architectures~\citep{liu2024unitime}, to leverage world knowledge and textual side-information for forecasting.
However, existing CAF methods assume that useful (i.e., supporting) context is available, whereas real-world forecasting often requires context discovery from heterogeneous sources.
Recent advances in \emph{deep research} (DR) systems suggest a path toward this setting: DR agents can search across sources, synthesize context into evidence, and use synthesized evidence to support downstream tasks~\citep{abaskohi2025drbench,bosse2025deep,du2025deepresearch}.
We therefore study the \textbf{context-aided forecasting via deep research (CAF via DR)} problem, evaluating whether agents can identify anomalies and patterns in the historical series, hypothesize plausible triggers, discover drivers and covariates likely to affect future values, and search for the context needed to make accurate predictions.

Existing benchmarks do not directly evaluate this setting. CAF benchmarks, such as Context is Key (CiK)~\citep{williams2024context}, assume that supporting evidence is provided, whereas DR benchmarks~\citep{abaskohi2025drbench,du2025deepresearch} evaluate retrieval intrinsically rather than by downstream forecasting utility.
Given this lack of benchmarks for CAF via DR, we introduce \textbf{Dr-CiK}, a benchmark for testing whether agents can discover predictive context, reject forecast-dependent distractors, distill the retrieved context into forecast-useful evidence, and use that evidence to improve grounded forecasts.
A differentiating feature of Dr-CiK w.r.t. DR benchmarks is that distractor rejection is forecast-dependent: distractors may be semantically compatible with the variable to forecast (e.g. by mentioning the target entity)
but would lead to an incorrect forecast if used as context.
Indeed, some distractors are temporally misaligned with the forecast horizon and require \emph{temporal reasoning}~\citep{beniwal2024remember}, others contradict the observed history (e.g., by describing unobserved patterns) and require \emph{time-series reasoning}~\citep{merrill2024language,chow2024towards,kong2025position,potosnak2024implicit,ye2024ts} to rule them out. Thus, Dr-CiK tests whether agents retrieve evidence that is not merely related, but forecast-useful.

Our central contribution is a generation pipeline that converts any CAF task into CAF via DR one.
For each instance, it generates documents that contain the supporting evidence and categorized distractors, with labels indicating which documents carry the evidence.
We release 279 CAF-via-DR tasks created with this pipeline: 199 from CiK and GIFT-CTX~\citep{salesforce2026giftctx}, and 80 from expert-curated tasks spanning environment, economics, transportation, healthcare, and observability domains.
We evaluate DR agents paired with context-aided forecasters, measuring DR quality, forecasting quality and sensitivity to context quality.
Results show that DR quality is a primary driver of forecasting performance, confirming the validity of the generation pipeline.
More precisely, supporting context substantially improves forecasting accuracy over forecasting without context, while DR-gathered context often yields little benefit and can even degrade performance due to incomplete evidence recovery and the presence of distractors, which existing forecasters are vulnerable to.
Overall, Dr-CiK exposes a large gap between current DR agents and ground-truth supporting evidence, underscoring the need for methods that can reliably identify, filter, and synthesize external evidence specialized for forecasting.
To summarize, our contributions are:
\vspace{-0.4em}
\begin{itemize}[leftmargin=*, itemsep=0.1em, topsep=0.1em, parsep=0pt, partopsep=0pt]
\item We introduce the \emph{Context-aided Forecasting via Deep Research} problem, where an agent must actively search a document corpus to find forecasting-relevant context, avoid distractors, distill the retrieved context into forecast-useful evidence, and produce evidence-grounded forecasts.
\item We develop \textbf{Dr-CiK}, a benchmark for evaluating CAF via DR, and release $279$ tasks with ground-truth annotations for supporting evidence retrieval and CAF. This design enables stage-wise evaluation of failures from DR to CAF.
\item We benchmark existing DR agents with forecasters and show that DR quality shapes forecasting performance: supporting evidence substantially lowers the scaled Continuous Ranked Probability Score relative to both no-context and DR-retrieved-context settings, yet most DR agents recover less than $5\%$ of supporting evidence, while more than $70\%$ of their citations point to distractors.
\end{itemize}

\section{Related Work}

\paragraphtight{Deep Research Agents and Benchmarks}
DR benchmarks evaluate agents on multi-step search, retrieval, synthesis, and citation, treating retrieved information or synthesized reports as final outputs. ReAct~\citep{yao2022react} introduced the reason-and-act paradigm underlying many tool-using agents, such as STORM~\citep{shao2024assisting} and GPT-Researcher~\citep{elovic2023gptresearcher}, which conduct multi-step web research and report synthesis, while GAIA~\citep{mialon2024gaia} evaluates general-purpose agents on real-world question answering with reasoning, web browsing, and tool use. Recent benchmarks target large-scale DR settings: DR Bench~\citep{bosse2025deep} provides a frozen retrieval environment. DeepResearch Bench~\citep{du2025deepresearch} evaluates PhD-level reports, DRBench~\citep{abaskohi2025drbench} focuses on enterprise DR, DRACO~\citep{zhong2026draco} measures factuality, comprehensiveness, presentation, and citation quality, and DeepSearchQA~\citep{gupta2026deepsearchqa} evaluates agentic retrieval for open-ended questions. These benchmarks evaluate retrieved context, reports, or citations as final outputs, whereas Dr-CiK evaluates whether retrieved context improves downstream forecasting.

\paragraphtight{Context-Aided Forecasting}
Context-aided forecasting (CAF) studies how textual context can improve time-series prediction. CiK~\citep{williams2024context} evaluates models on human-crafted text and time-series pairs; TimeMMD~\citep{liu2024time} pairs time series with relevant textual reports; TimesX~\citep{liu2026rethinking} offers a larger, more diverse, leakage-controlled benchmark, but its noisy context lacks the evidence/distractor labels needed to evaluate DR. Aurora~\citep{wu2026aurora} and DoubleCast~\citep{zheng2026overcoming} synthesize textual context for time series, and TemporalBench~\citep{weng2026temporalbench} evaluates temporal reasoning under increasingly rich natural-language contexts. Other work studies reasoning failures in multimodal forecasting: \citet{merrill2024language} shows that strong LLMs struggle on synthetic time-series and context tasks, while \citet{ashok2025beyond} elicits rationales before context-aided forecasts to diagnose failures. These works test whether models can use provided context. Dr-CiK instead tests whether agents can discover forecasting-relevant context from a noisy corpus and use it effectively for forecasting.

\paragraphtight{Event-Forecasting Agents}
Some agentic forecasting benchmarks do use downstream predictive performance as the evaluation target, but they focus mainly on categorical rather than numerical time-series forecasts.
ForecastBench~\citep{karger2024forecastbench} and ProphetArena~\citep{yang2025llm} release live event-forecasting questions for real-time agentic evaluation with web retrieval, but provide no controlled setting for isolating DR agent capabilities.
FutureX~\citep{zeng2025futurex} evaluates agentic reasoning, search, and tool use for future prediction, again primarily for categorical forecasts.
\citet{wang2024news} integrates textual analysis into a multi-LLM forecasting pipeline with multi-turn reflection, but uses a deterministic workflow with pre-filtered news.
The closest work to ours is Bench to the Future (BTF)~\citep{wildman2025bench}, a pastcasting benchmark with known-resolution questions and large offline web-page snapshots.
However, it covers categorical rather than numerical time-series forecasting.

Table~\ref{tab:benchmark_comparison} situates Dr-CiK against related CAF and event-forecasting benchmarks.
Dr-CiK differs from these works by enabling controlled analysis of forecasting accuracy when supporting evidence is not readily available. Its labeled and controllable document corpus and forecast-dependent distractors make it possible to evaluate under intervention not only whether a method succeeds, but \textit{why}.

\begin{table}[t]
\centering
\caption{
Comparison with related context-aided forecasting and event-forecasting benchmarks.
}
\vspace{-5pt}
\label{tab:benchmark_comparison}
\footnotesize
\setlength{\tabcolsep}{4pt}
\resizebox{\linewidth}{!}{
\begin{tabular}{lccccccc}
\toprule
\textbf{Benchmark} & \textbf{\#Tasks} & \textbf{Horizon} & \makecell{\textbf{Includes}\\\textbf{TS}} &
\makecell{\textbf{Context}\\\textbf{access}} & \makecell{\textbf{Forecast-helpful}\\\textbf{context}} & \makecell{\textbf{DR}\\\textbf{GT}} & \makecell{\textbf{Controlled}\\\textbf{distractors}} \\
\midrule
BTF v1~\citep{wildman2025bench} & 299 & Event & \ding{55} & Web search & \ding{55} & \ding{55} & \ding{55} \\
OpenForesight~\citep{chandak2025scaling} & 302 & Event & \ding{55} & Given & \ding{55} & -- & -- \\
FutureBench~\citep{bianchi2025back} & 156 & Event & \ding{55} & Web search & \ding{55} & \ding{55} & \ding{55} \\
LEAF-event~\citep{tan2026leaf} & 30 & Event & \ding{55} & Given & \ding{55} & -- & -- \\
CiK~\citep{williams2024context} & 71 & 6--99 steps & \ding{51} & Given & \ding{51} & -- & -- \\
GIFT-CTX~\citep{salesforce2026giftctx} & 245 & 50--150 steps & \ding{51} & Given & \ding{51} & -- & -- \\
TemporalBench-T4~\citep{weng2026temporalbench} & 271 & 20--168 steps & \ding{51} & Given & \ding{55} & -- & -- \\
TimesX~\citep{liu2026rethinking} & 190 & 12 steps & \ding{51} & Given & \ding{55} & -- & -- \\
\textbf{Dr-CiK (ours)} & \textbf{279} & \textbf{15--2,798 steps} & \ding{51} & \textbf{Deep research} & \ding{51} & \ding{51} & \ding{51} \\
\bottomrule
\end{tabular}
}
\vspace{-1.2em}
\end{table}

\section{Context-Aided Forecasting via Deep Research}

\begin{figure}[t]
    \centering
    \vspace{-5pt}
    \includegraphics[width=\linewidth]{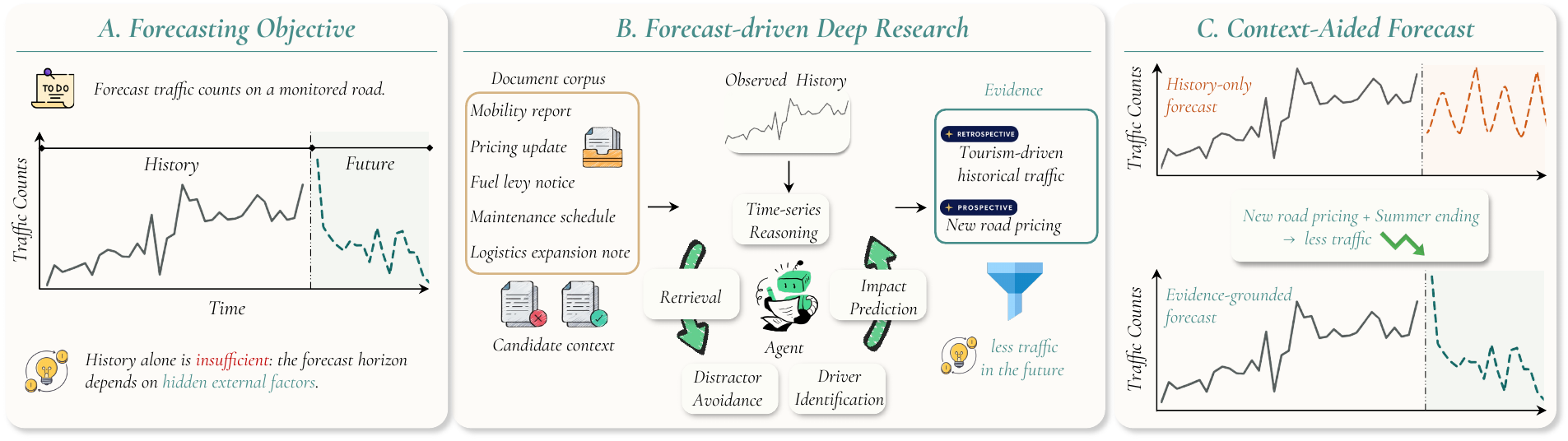}
    \vspace{-1.3em}
\caption{
\textbf{Context-Aided Forecasting via Deep Research}.
(A) The task provides historical observations and asks for a future trajectory that depends on external factors.
(B) Given a document corpus and the observed series, the agent reasons about the query, retrieves relevant context while avoiding distractors, and extracts and synthesizes the retrieved context into forecast-useful evidence.
(C) A forecast conditioned on the extracted evidence substantially improves from a history-only forecast.
}
    \label{fig:dr4f}
    \vspace{-.7em}
\end{figure}

Let $\mathbf{X}_H=[X_1,\dots,X_t]$ and $\mathbf{X}_F=[X_{t+1},\dots,X_T]$ denote historical and future observations of a univariate time series, with $X_\tau \in \mathbb{R}$.
Traditional probabilistic forecasting estimates $P(\mathbf{X}_F \mid \mathbf{X}_H)$, while Context-aided Forecasting (CAF) estimates $P(\mathbf{X}_F \mid \mathbf{X}_H,\mathbf{C})$, where $\mathbf{C}$ denotes the textual \emph{context} which the forecast is conditioned on.
We introduce a new setting, where the context $\mathbf{C}$ is not readily available but must be extracted from a document corpus $\mathbf{D}=\{D_1,\dots,D_n\}$, as pictured in Figure~\ref{fig:dr4f}.
In what follows, we use \emph{context} to refer broadly to any textual information used for forecasting, and \emph{evidence} to refer to the ground-truth context that should be recovered and used for optimally forecasting.
Let $\mathbf{E}_s$ be the set of evidence that is spread across the set of supporting documents from the corpus $\mathbf{D}_s \subset \mathbf{D}$.
A good research process should retrieve the supporting documents, extract all forecast-useful evidence $\mathbf{E}_s$, while avoiding distractor documents $\mathbf{D} \setminus \mathbf{D}_s $.

Given a loss function $\mathcal{L}$\footnote{The inequality holds uniquely and tightly only if the loss function  evaluates the full predictive distribution and is a proper scoring rule \citep{gneiting2007strictly}.} on the predictive distribution over $\mathbf{X}_F$ and a realization $\mathbf{x}_F \sim \mathbf{X}_F|\mathbf{X}_H,\mathbf{E}_s$, we expect that good forecasting methods will achieve optimal performance when conditioned on the evidence $\mathbf{E}_s$, compared to any other context $\mathbf{C}$ derived from the corpus:
\begin{equation}
    \mathop{\mathbb{E}}_{\mathbf{x}_F} \mathcal{L}\left(
        P(\mathbf{X}_F|\mathbf{X}_H,\mathbf{E}_s), \mathbf{x}_F
    \right)
    \le
    \mathop{\mathbb{E}}_{\mathbf{x}_F} \mathcal{L}\left(
        P(\mathbf{X}_F|\mathbf{X}_H,\mathbf{C}), \mathbf{x}_F
    \right)
    \quad \forall \mathbf{C} \in \mathcal{C}(\mathbf{D}),
\end{equation}
where $\mathcal{C}(\mathbf{D})$ denotes the space of all possible textual contexts extracted from $\mathbf{D}$.

\section{Dr-CiK: A Testbed for Deep Research for Forecasting}

\label{sec:method}

Evaluating CAF via DR requires a challenging environment where an agent must actively discover forecast-relevant evidence while rejecting distractors.
Existing CAF resources cannot supply that environment directly: their context is paired one-to-one with the target series, hence relevance is trivially decidable by surface-level semantic association. We confirm this in \S\ref{app:corpus_validation}, where running DR agents over the aggregated CiK contexts yields near-ceiling evidence recovery.
We therefore generate the DR environment starting from existing CAF datasets: we generate a document corpus in which the ground-truth context (the evidence) is fragmented across multiple supporting documents, alongside categorized distractors, together with the labels needed to evaluate recovery.
Dr-CiK is designed around three scientific pillars to ensure that every task demands true foresight-driven DR:
\vspace{-0.3em}

\begin{enumerate}[leftmargin=*, itemsep=0.1em, topsep=0.1em, parsep=0pt, partopsep=0pt]
    \item \textbf{Scale and Realism:} The generation pipeline can be applied to any CAF dataset with minimal effort. We run it on three sources: CiK~\citep{williams2024context} and GIFT-CTX~\citep{salesforce2026giftctx}, which together yield 199 tasks and provide the sample size necessary for \emph{statistical robustness}, and a newly expert-curated CAF set contributing 80 tasks whose real-world time series capture the complexity of physical dynamics (details are provided in \S\ref{app:benchmark-composition}).

    \item \textbf{Controlled Difficulty:}
    The DR environment is constructed by jointly controlling the supporting and distracting context. On the supporting documents side, we fragment the ground-truth evidence into multi-hop reasoning steps with controllable depth, while applying anonymization and time-shifting to mitigate forecasting-stage memorization shortcuts. On the distractor side, we introduce a five-class taxonomy of distractors (Off-variable, Off-entity, Off-window, Counterfactual history, and Irrelevant) forcing agents to rely on time-series and text reasoning rather than semantic relevance.

    \item \textbf{Fine-grained Evaluation:} Ground truth is automatically provided for both evidence recovery and forecasting, along with fine-grained difficulty annotations (e.g., temporal complexity, evidence explicitness, see \S\ref{appx:difficulty_annotations}), so evidence recovery and forecasting can be scored independently and jointly end-to-end, isolating where an agent fails and what that failure costs downstream.
\end{enumerate}

\subsection{Task Generation Pipeline}
\begin{figure}[t]
    \centering
    \vspace{-5pt}
    \includegraphics[width=\linewidth]{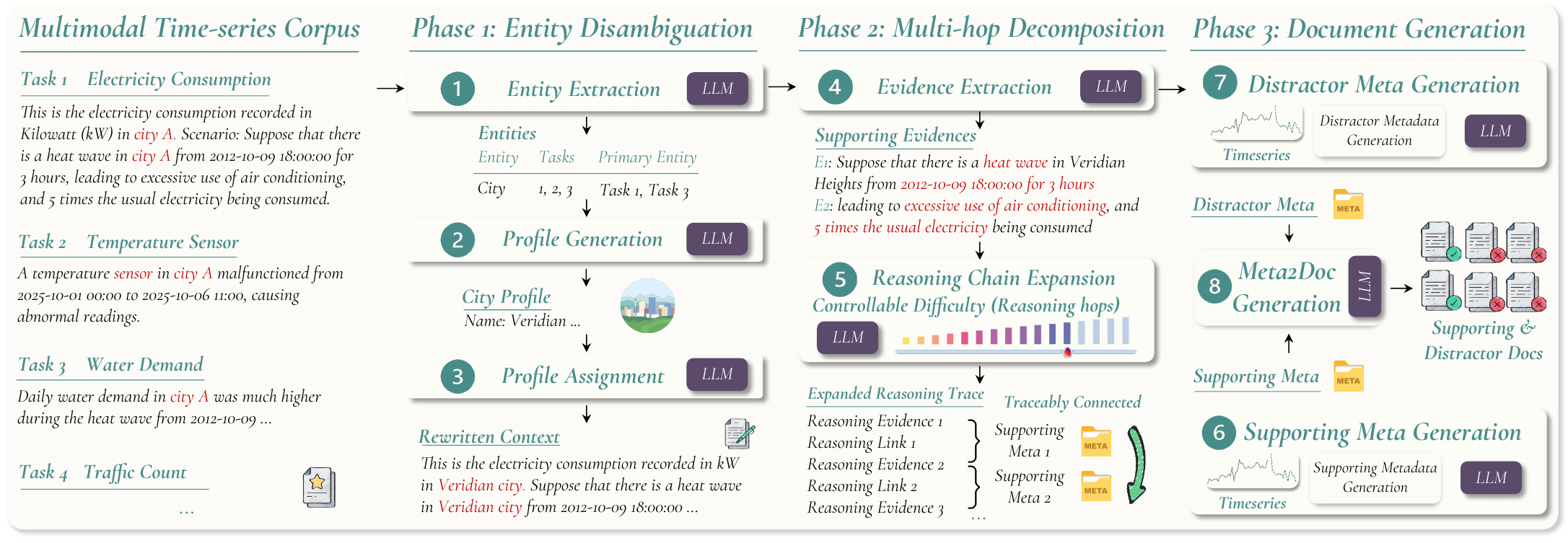}
    \vspace{-15pt}
    \caption{\textbf{The DR Environment Generation Pipeline.} This scalable pipeline transforms CAF instances into DR environments through three core phases. \textbf{(1) Entity Disambiguation (Steps 1-3):} Real-world entities are mapped to synthetic profiles to ensure identifiability and mitigate data contamination, \textbf{(2) Multi-hop Decomposition (Steps 4-5):} To prevent retrieval from collapsing into single-document lookup, supporting evidence is fragmented into a multi-hop reasoning chain whose hop count controls how broadly evidence must be gathered, \textbf{(3) Document Generation (Steps 6-8):} Supporting and distractor documents are synthesized from the expanded traces and our distractor taxonomy. Throughout all stages, an iterative, human-calibrated LLM-judge loop enforces factual and logical integrity. Finally, a coding agent judge audits the complete environment, with human experts executing targeted repairs on flagged inconsistencies. More details are shown in Figure~\ref{fig:workflow_example}.}
    \vspace{-7pt}
    \label{fig:workflow}
\end{figure}

A core contribution of Dr-CiK is its scalable task-generation pipeline (Figure~\ref{fig:workflow}), which transforms multimodal time-series instances into DR environments through three phases, each addressing a distinct evaluation requirement.
Specialized human-calibrated LLM-judges oversee the step outputs in a generate-until-correct loop, and a final Agent Judge audits the completed environment.

\paragraphtight{Phase 1: Entity Disambiguation (Steps 1-3).}
In our evaluation, documents from all tasks are pooled into a single shared corpus.
Entities may share generic names or types across tasks, so we generate task-specific synthetic profiles for supporting documents. This preserves realistic overlap while making each task's relevant entities distinguishable at DR time. We use the following procedure:
\stagecircle{1} \emph{Entity Extraction} identifies the type-level entities in each task, designates one primary entity aligned with the time-series variable, and extracts containment relations between entities.
\stagecircle{2} \emph{Profile Generation} computes the minimal number of synthetic profiles needed across the task pool, reusing profiles wherever possible, and generates realistic textual descriptions for each.
\stagecircle{3} \emph{Profile Assignment} maps each task's primary entity (and any ancestors) to its assigned profile and rewrites the original context accordingly.
This disambiguation pass also helps reduce the effect of memorization, as studied models may bypass retrieval by exploiting parametric memory of historical events.
We further shift the time window of real-world time series and adjust context timestamps accordingly. Together, these mechanisms mitigate both lexical and numerical memorization heuristics.

\paragraphtight{Phase 2: Multi-hop Decomposition (Steps 4-5).}
Placing the ground-truth supporting evidence into a single document would degenerate DR into a single-document lookup, far from real-world settings where the evidence a forecaster needs is spread across many sources. Phase 2 reproduces this distributed structure.
\stagecircle{4} \emph{Evidence Decomposition} decomposes the rewritten supporting context into a collectively exhaustive set of supporting evidence with their causal relationships and time periods preserved.
\stagecircle{5} \emph{Reasoning Chain Expansion} interleaves supporting evidence with generated and explicit Reasoning Link nodes that articulate each reasoning step, producing an alternating path constrained to have no branches, cycles, or shortcuts.
This path's hop count is the principal difficulty knob. In Phase 3, nodes are spread across supporting documents, so longer chains force an agent to retrieve more documents and reason over longer sequences.

\paragraphtight{Phase 3: Document Generation (Steps 6-8).}
Finally, \stagecircle{6} \emph{Supporting Document Generation} uses an \emph{overlapping assignment strategy} when grouping evidence nodes into document metadata: each supporting document covers two adjacent evidence nodes plus the connecting Reasoning Link, so consecutive documents share a causal anchor and the chain remains traceably connected for DR.
Then, \stagecircle{7} \emph{Distractor Document Generation} drafts distractor metadata according to our five-class taxonomy (see \S\ref{sec:distractor_taxonomy}), and finally, \stagecircle{8} \emph{Meta2Doc Generation} renders supporting and distractor metadata into natural-language documents of appropriate types (e.g., reports) to populate the corpus.

\paragraphtight{Quality Control: Human-Calibrated Local-Global Verification.}
Synthesizing environments at this scale requires automatically checking against hallucinations, logical shortcuts, leakage of evidence in distractors and incomplete coverage of evidence in supporting documents.
We achieve this through a strict human-in-the-loop verification protocol.
Authors first manually judge the quality of and annotate a subset of tasks for \emph{every individual generation step}. These human judgments serve as the ground truth to iteratively calibrate a suite of specialized LLM Judges (e.g., Evidence Judge, Reasoning Judge, Distractor Judge). We explicitly refine their prompts until their evaluation strictness aligns with or exceeds human baselines. During full-scale synthesis, we then enforce a \emph{generate-until-correct} strategy where intermediate outputs at every pipeline stage are repeatedly regenerated until they pass their respective calibrated judge.
We further validate the corpus with a cloze audit reconstructing masked evidence from supporting documents (initial mean score \rev{0.9372}, with all residual errors subsequently repaired) and a distractor distinguishability audit confirming all 6,975 distractors are explicitly rejectable (§\ref{app:corpus_validation}), indicating high-fidelity document generation with only minor, correctable gaps.
The completely synthesized environment finally undergoes a holistic audit by an overarching Agent Judge paired with human experts to execute targeted repairs on any residual macro-level inconsistencies. Calibration protocols and judge prompts are provided in \S\ref{appx:judge_details}.

\subsection{Forecast-dependent Distractor Design}
\label{sec:distractor_taxonomy}

\begin{figure}[t]
    \centering
    \vspace{-5pt}
    \includegraphics[width=\linewidth]{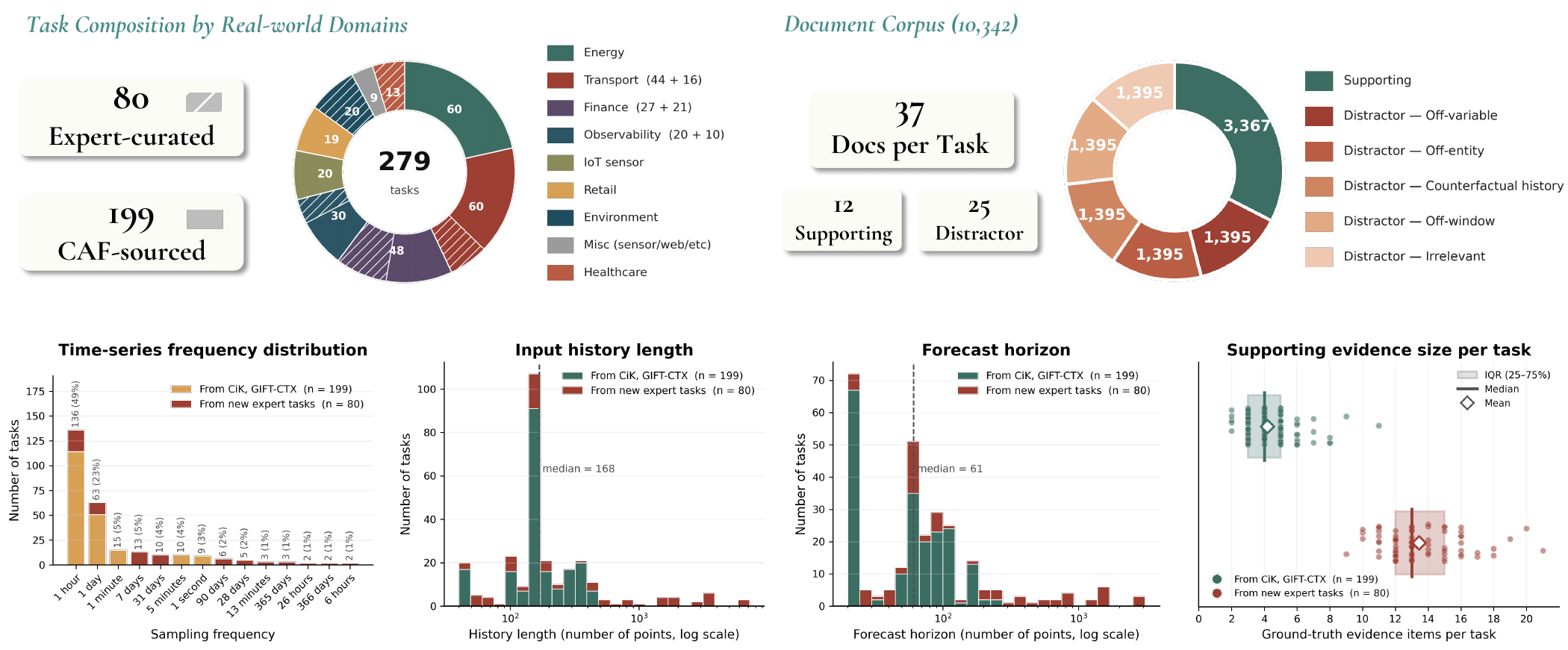}
    \vspace{-10pt}
    \caption{An overview of the released 279 tasks with supporting evidence, 10,342 total labeled documents, 5 categories of distractors, and diverse time-series spanning 9 domains. Dr-CiK is designed as a scalable pipeline with controllable difficulty and a fine-grained ground-truth, to evaluate context-aided forecasting via deep research (see \S\ref{sec:method}).}
    \vspace{-.7em}
    \label{fig:overview}
\end{figure}

Existing DR benchmarks rely on \emph{semantic hard negatives}, documents lexically similar to the query but irrelevant to the answer. We argue that this evaluation surface is misaligned with the demands of CAF.
In our setting, distractors are plausible contexts that can steer the forecast in the wrong direction.
To our knowledge, Dr-CiK is the first benchmark where distractors are explicitly \emph{forecast-dependent} (Figure~\ref{fig:overview}B). Each distractor is generated from the ground-truth supporting context and historical time series to drive the downstream forecaster toward a confidently incorrect prediction. Crucially, every distractor remains \emph{rejectable}: it can be logically ruled out by cross-referencing the task metadata or the observed historical time series.
This property is enforced during generation and human-LLM verification, ensuring intrinsic data ambiguity in the benchmark is low.
To systematically probe agents' failures, we organize these adversarial distractors into a five-class taxonomy:
\vspace{-0.3em}
\begin{itemize}[leftmargin=*, itemsep=0.1em, topsep=0.1em, parsep=0pt, partopsep=0pt]
    \item \textbf{Irrelevant.} Realistic, on-topic details about the correct entity, target variable, and time window, but with no forecast-relevant content. This tests whether agents distinguish true causal drivers from plausible fillers.
    \item \textbf{Off-variable.} A plausible causal mechanism that terminates on a different variable of the correct entity. This tests whether agents verify that retrieved context affects the target series.
    \item \textbf{Off-entity.} The ground-truth context, rewritten to apply to another entity of the same type and to imply a contradictory future. This tests whether agents enforce strict entity grounding under maximal semantic overlap.
    \item \textbf{Off-window.} The correct context for the correct entity, but pertaining to a period outside the task's time window. This tests whether agents anchor retrieved evidence to the prediction window.
    \item \textbf{Counterfactual history.} A pattern that contradicts
    the historical series (e.g., a substituted trend, periodicity, set of values, or
    windows) projected into a qualitatively wrong future.
    This tests whether agents validate narrative claims against the
    numerical series rather than trusting prose.
\end{itemize}

This taxonomy delivers two advantages over traditional keyword-style distractors.
First, it penalizes surface-level semantic similarity: because distractors share the linguistic structure of the supporting evidence, it is
indistinguishable from it on lexical or embedding similarity alone, and can be rejected only by verifying its grounding.
Second, it enables diagnostic attribution, allowing failures to be attributed directly to causal, temporal, entity-level, or numerical reasoning deficits.
Together, these properties make the benchmark a controlled stress test of the
multimodal discrimination that real-world forecasting agents must perform.
Distractor examples are detailed in \S~\ref{appx:distractor}.

\subsection{Dataset Statistics}

By applying the described generation pipeline, we release 199 tasks derived from CiK and GIFT-CTX and 80 derived from new expert-curated CAF tasks.
We add this second set because the first principally contain series that are synthetic or adjusted to match the supplied context, with contexts that lack the register and specificity a practitioner would produce. The expert-curated tasks instead impose deployment-realistic
constraints (real unedited series, context written by domain experts) and
prove the harder subset, requiring roughly three times the grounded evidence depth
of the generated tasks, as illustrated in Figure~\ref{fig:overview} along other key statistics.
Finally, supporting evidence in the released tasks is of two kinds. \emph{Retrospective} evidence characterizes the observed history, and can describe for instance anomalies to be ignored in the future. \emph{\rev{Prospective}} evidence reports conditions over the forecast horizon that are \emph{known in
advance} at prediction time, such as scheduled maintenance, announced policy changes,
planned interventions. We report results by evidence role in \S\ref{app:evidence_roles}: even \textsc{Codex}, the strongest DR agent, recovers only 25.98\% of retrospective evidence versus 15.85\% of prospective evidence, indicating that DR agents struggle disproportionately with future-oriented evidence that has no anchor in the observed history.

\section{Experiments}
\label{sec:experiments}

\vspace{-0.5em}
\label{sec:evaluation_protocol}
Evaluating CAF via DR systems is complex since failures can stem from multiple sources:
an inaccurate forecast could stem from failures to recover the supporting evidence, or to recognize distractors or to leverage the extracted evidence to make predictions.
Because Dr-CiK
supplies ground truth at both the intermediate stage (which documents and
evidence support the target) and the final one (the realized outcome), we can
attribute error to the stage that produced it through a three-level protocol:
\vspace{-0.4em}

\begin{itemize}[leftmargin=*, itemsep=0.2em, topsep=0.2em, parsep=0pt, partopsep=0pt]

\item \textbf{Level 1: End-to-End.} Measures downstream forecasting accuracy when a DR agent retrieves context and synthesizes evidence for a downstream forecaster, conflating all three failure sources.

\item \textbf{Level 2: Deep Research.} Isolates the DR agent's retrieval and synthesis quality (\emph{Evidence Recall}, \emph{Supporting Document Recall}, \emph{Distractor Avoidance}), independent of the forecaster.

\item \textbf{Level 3: Context-Aided Forecasting.} Studies forecasters' text-grounding capability and distractor resilience by evaluating the forecaster under controlled evidence conditions.

\end{itemize}

\subsection{Experimental Setup}

We evaluate five DR agents (\textsc{Codex}~\citep{openai_codex_2026}, \textsc{DrBench}~\citep{abaskohi2025drbench}, \textsc{Bench2Future}~\citep{wildman2025bench}, \textsc{OpenDR}~\citep{langchain_open_deep_research}, \textsc{Retrieval}) and four forecaster families: statistical baselines (ETS, SES, Naive, (S)ARIMA), pretrained time-series models (Chronos~\citep{ansari2024chronos}, TimesFM-3~\citep{jain2026timesfm3}, Toto-2.0~\citep{khwaja2026toto}, TiRex-2~\citep{auer2026tirex}, Moirai-2.0~\citep{woo2024unified}, Moirai-MoE~\citep{liu2024moirai}), multimodal models (Aurora~\citep{wu2026aurora}), zero-shot LLMs (Gemini-3.1-flash-lite~\citep{google_gemini_3_1_flash_lite_preview}, Mistral-medium-3.1~\citep{liu2026ministral}, Qwen-3.5 at 4B/9B/27B~\citep{qwen3.5}, Llama-3.2-3B~\citep{grattafiori2024llama}, Phi-4-mini~\citep{abdin2024phi}) with Direct Prompt (DP) forecasting strategy~\citep{williams2024context}, and a forecasting agent (MoiraiAgent~\citep{salesforce2026moiraiagent} backed by Gemini).
Each forecaster draws $S{=}25$ trajectories per Dr-CiK task, and we report the scaled Continuous Ranked Probability Score (sCRPS). For task-generation, we use Gemini-3-flash~\citep{google_gemini_3_flash_preview}. Evaluation metrics are defined in \S\ref{app:metrics}, and details are provided in \S~\ref{app:experimental-setup}.

\subsection{Level 1: End-to-End Results}

\vspace{-2pt}
\begin{table}[h]
\centering
\begin{minipage}[t]{0.54\textwidth}
    \centering
    \footnotesize
    \setlength{\tabcolsep}{4pt}
    \renewcommand{\arraystretch}{1.2}
    \captionof{table}{End-to-end forecasting performance (sCRPS$\downarrow$) across context conditions and forecasting methods on the full 279-task pool ($\pm$ indicates standard errors). Best results by forecaster over the DR contexts are bold, and best results (across forecaster and context) are underlined.}
    \label{tab:end_to_end}
    \vspace{-6pt}
    \resizebox{\linewidth}{!}{
\begin{tabular}{lccc}
\toprule
\textbf{\textsc{Context}}
& \textbf{Aurora}
& \textbf{DP-Gemini}
& \textbf{MoiraiAgent} \\
\midrule
\textsc{No Context}
& 0.484 $\pm$ 0.051
& 0.341 $\pm$ 0.031
& 0.357 $\pm$ 0.030 \\
\textsc{Original Context}
& 0.487 $\pm$ 0.052
& 0.253 $\pm$ 0.030
& \underline{0.228 $\pm$ 0.028} \\
\midrule
\textsc{Bench2Future}
& 0.483 $\pm$ 0.051
& 0.649 $\pm$ 0.062
& 0.539 $\pm$ 0.055 \\
\textsc{DrBench}
& 0.484 $\pm$ 0.051
& 0.599 $\pm$ 0.060
& 0.515 $\pm$ 0.054 \\
\textsc{Retrieval}
& \textbf{0.481 $\pm$ 0.051}
& 0.610 $\pm$ 0.061
& 0.548 $\pm$ 0.058 \\
\textsc{OpenDR}
& 0.483 $\pm$ 0.051
& 0.566 $\pm$ 0.054
& 0.421 $\pm$ 0.039 \\
\textsc{Codex}
& 0.484 $\pm$ 0.051
& \textbf{0.336 $\pm$ 0.030}
& \textbf{0.322 $\pm$ 0.032} \\
    \bottomrule
    \end{tabular}
    }
\end{minipage}\hfill
    \vspace{2pt}
\begin{minipage}[t]{0.44\textwidth}
    \centering
    \footnotesize
    \setlength{\tabcolsep}{2.8pt}
    \renewcommand{\arraystretch}{1.22}
    \captionof{table}{DR quality measured as the fractions of reported evidence (Evidence Recall), of retrieved supporting documents (Supporting Document Recall) and of distractors avoided (Distractor Avoidance) on the full 279-task pool.}
    \label{tab:agent_retrieval_metrics_transposed}
    \vspace{-6pt}
    \resizebox{\linewidth}{!}{
    \begin{tabular}{lccc}
        \toprule
        \textsc{DR Agents} & \makecell{\textsc{Evidence}\\\textsc{Recall}} & \makecell{\textsc{Sup. Doc.}\\\textsc{Recall}} & \makecell{\textsc{Distract.}\\\textsc{Avoid.}} \\
        \midrule
        \textsc{B2Future} & 3.3\% & 7.0\% & 22.7\% \\
        \textsc{DrBench} & 3.4\% & 9.7\% & 28.8\% \\
        \textsc{Retrieval} & 3.9\% & 9.9\% & 21.0\% \\
        \textsc{OpenDR} & 4.3\% & 9.7\% & 23.9\% \\
        \textsc{Codex} & \textbf{27.9\%} & \textbf{50.6\%} & \textbf{42.3\%} \\
        \bottomrule
    \end{tabular}
    }
\end{minipage}
    \vspace{-.8em}
\end{table}

Table~\ref{tab:end_to_end} shows that current deep research agents still fall well short of the performance achievable with supporting evidence.
With supporting evidence, DP-Gemini and MoiraiAgent substantially outperform their no-context counterparts, demonstrating their ability to leverage contextual information. However, they do not approach this ceiling with any DR-synthesized evidence: Bench2Future and DrBench systematically degrade performance below the no-context baseline across both forecasters, indicating that their DR-synthesized evidence is actively harmful. \textsc{Codex} is the only DR agent that significantly \rev{improves} over forecasting without context, and only when paired with MoiraiAgent, suggesting that a strong forecaster can mitigate some DR noise but cannot overcome missing supporting evidence.
Aurora's predictions are context-insensitive: forecast quality does not track context quality, suggesting its time-series modality dominates textual context, or it fails to propagate textual evidence into forecasts. The definitions of context conditions are provided in \S~\ref{appx:context_modes}.

\begin{figure*}[h]
\centering

\begin{minipage}[t]{0.55\textwidth}
    \vspace{-0pt}
    \centering

    \begin{minipage}[t]{0.49\linewidth}
        \centering
        \includegraphics[width=\linewidth]{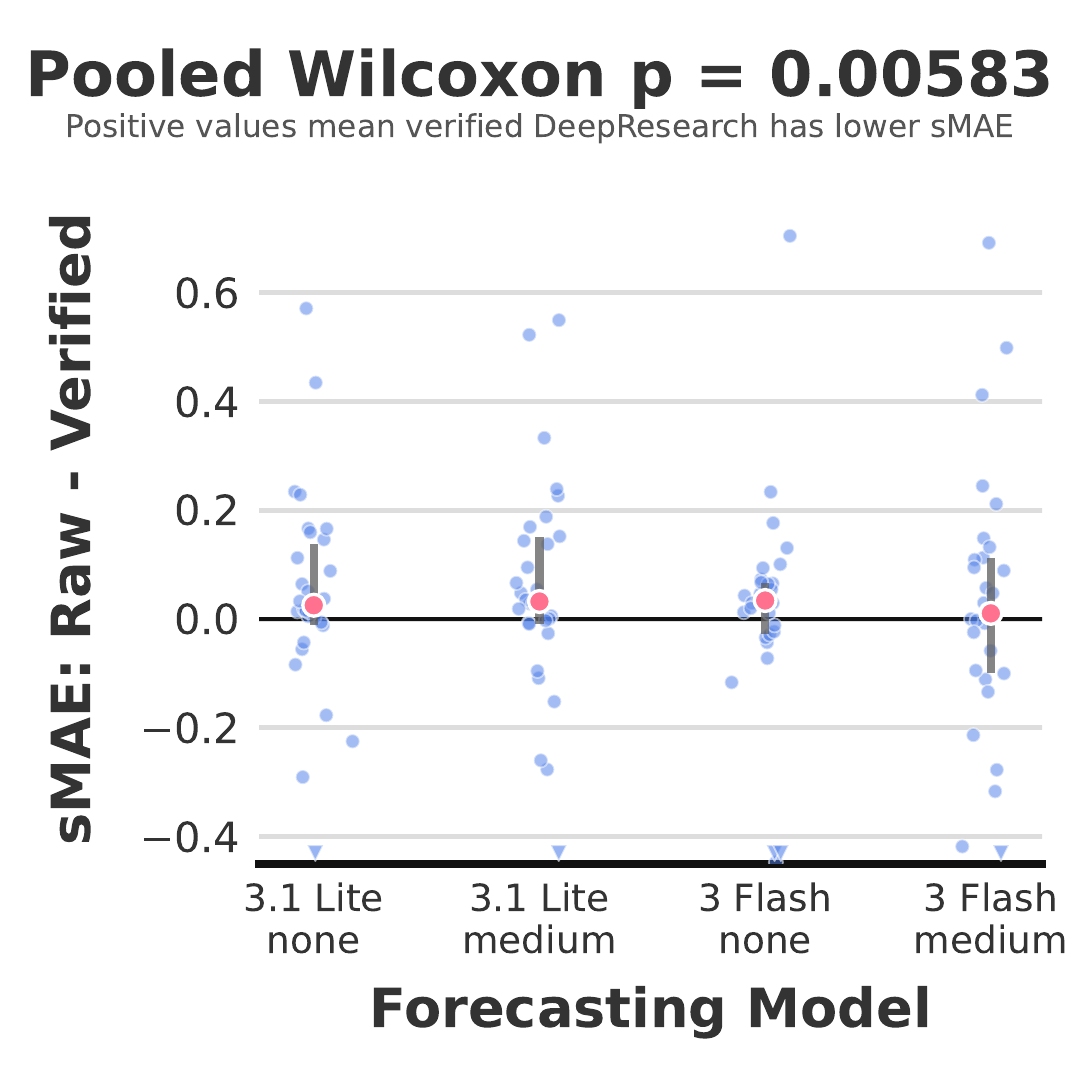}
        \label{fig:wilcoxon}
    \end{minipage}
    \hfill
    \begin{minipage}[t]{0.49\linewidth}
        \centering
        \includegraphics[width=\linewidth]{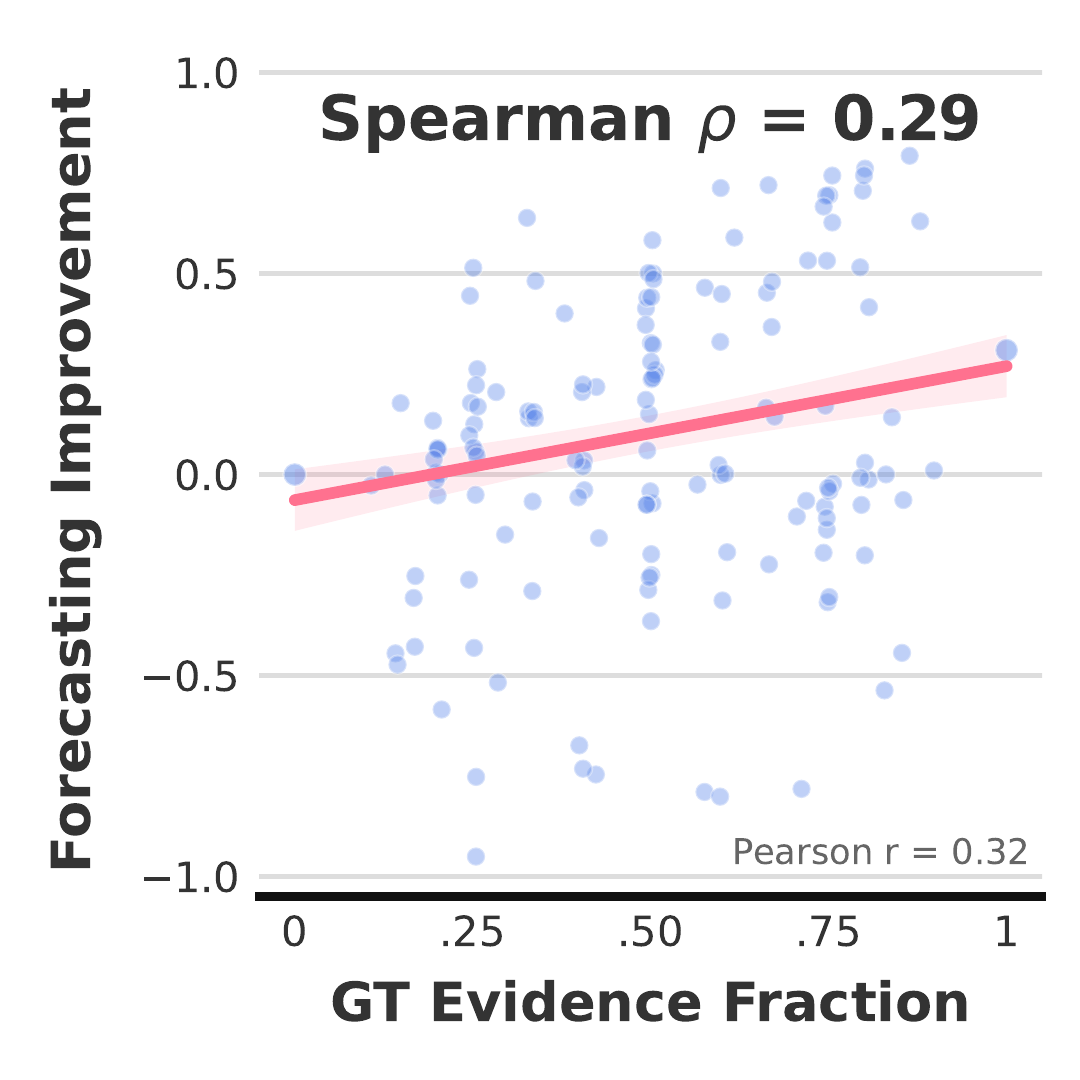}
        \label{fig:spearman}
    \end{minipage}

    \vspace{-13pt}
    \captionof{figure}{
Forecasting gains depend on context quality and evidence coverage.
Left: Filtering distractors improves sMAE.
Right: Higher supporting-evidence coverage correlates with larger forecasting gains.
}
    \label{fig:context_quality}
\end{minipage}
\hfill
\begin{minipage}[t]{0.42\textwidth}
    \vspace{8pt}
    \centering
    \includegraphics[width=\linewidth]{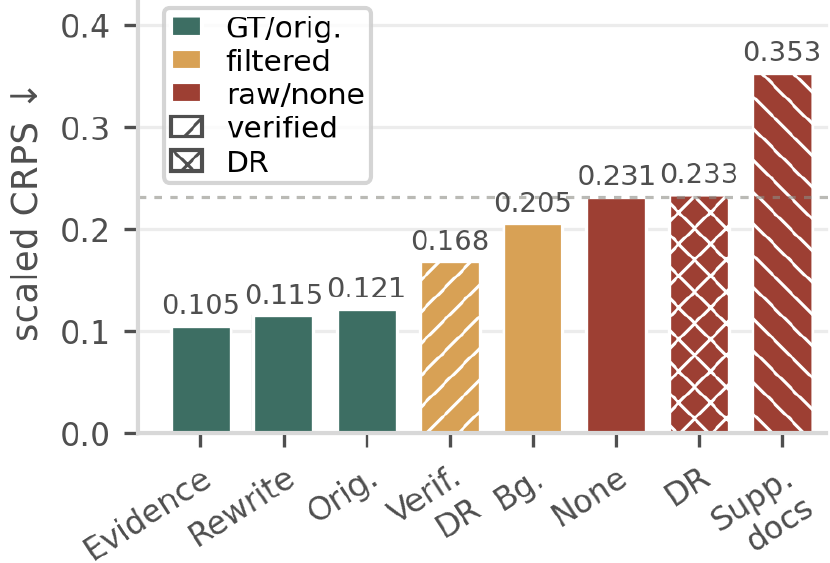}
    \vspace{-10pt}
    \captionof{figure}{Scaled CRPS across context-processing strategies. Lower is better.}
    \label{fig:context_mode}
\end{minipage}

\end{figure*}

To understand what limits DR-retrieved context quality, in Figure~\ref{fig:context_quality} we isolate two properties that plausibly drive the gap between DR-retrieved context and supporting evidence: the presence of distractors and the coverage of supporting evidence.
We first forecast from the agent's DR-synthesized evidence (Raw DR) and from a verified counterpart where distractors are filtered out (Verified DR), using DP with four backbone LLMs.
A one-sided Wilcoxon signed-rank test on the resulting paired sMAE values, pooled across methods, is significant, hence verification consistently improves forecasts.
Figure~\ref{fig:context_quality} studies the improvement from forecasting with retrieved context over forecasting without context, as a function of the proportion of supporting evidence an agent reports.
Supporting evidence coverage is positively rank-correlated with forecasting improvement over the no-context baseline.
These results suggest that both distractor avoidance and supporting evidence coverage substantially affect forecast accuracy.

\subsection{Level 2: Deep Research Analysis}

\begin{figure*}[h]
\centering
\newcommand{\leftpanelwidth}{0.29\textwidth}
\newcommand{\middlepanelwidth}{0.29\textwidth}
\newcommand{\rightpanelwidth}{0.4\textwidth}
\begin{minipage}[t]{\leftpanelwidth}
    \vspace{-5pt}
    \centering
    \includegraphics[width=\linewidth]{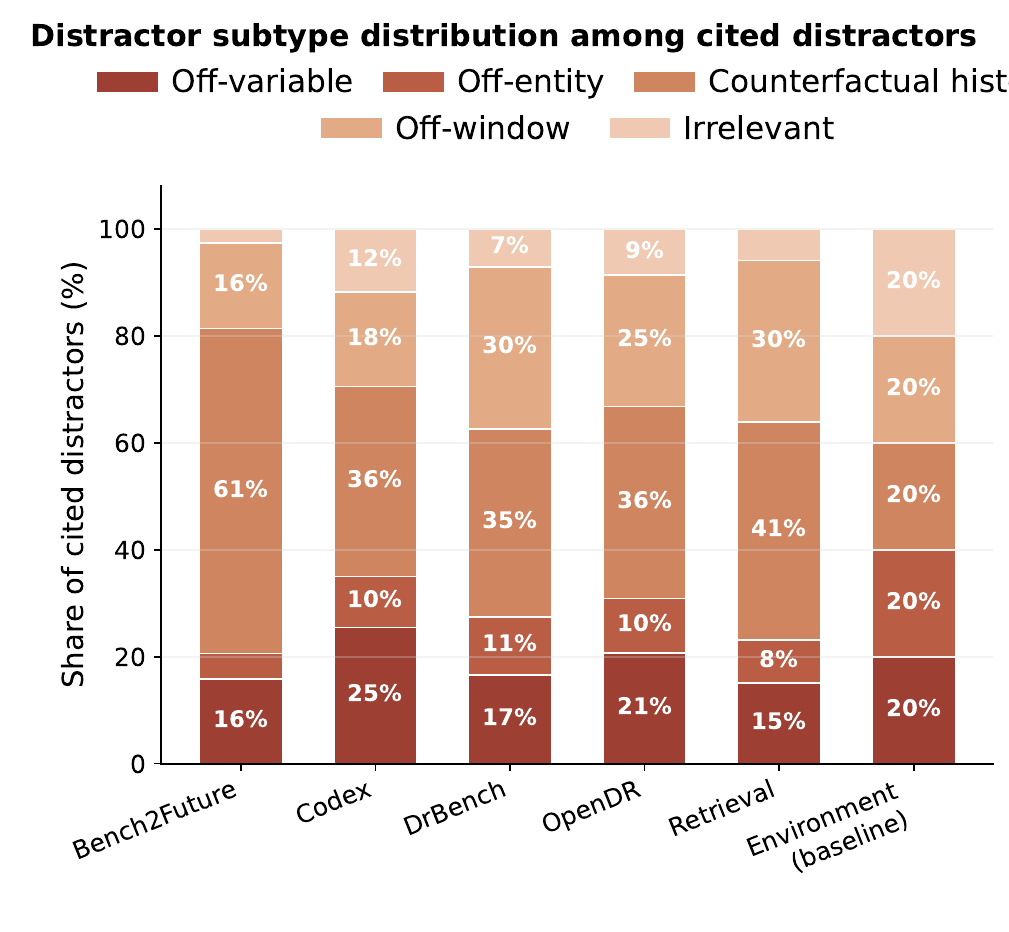}
    \vspace{-20pt}
    \captionof{figure}{Distractor patterns.}
    \label{fig:distractor_citation_patterns}
\end{minipage}
\hfill
\begin{minipage}[t]{\middlepanelwidth}
    \vspace{-5pt}
    \centering
    \includegraphics[width=\linewidth]{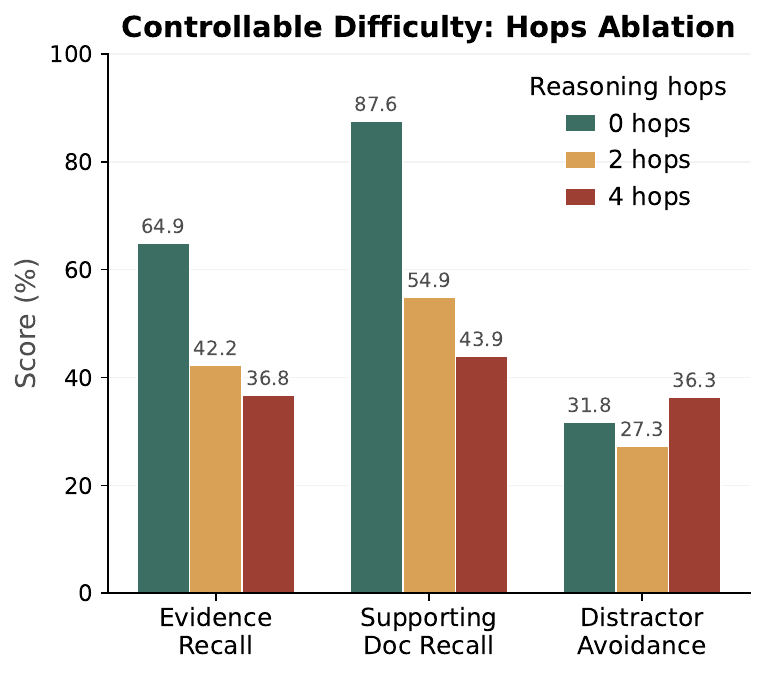}
    \vspace{-15pt}
    \captionof{figure}{Hops ablation.}
    \label{fig:dr_hops_ablation}
\end{minipage}
\hfill
\begin{minipage}[t]{\rightpanelwidth}
    \centering
    \vspace{-13pt}
    \includegraphics[width=0.8\linewidth]{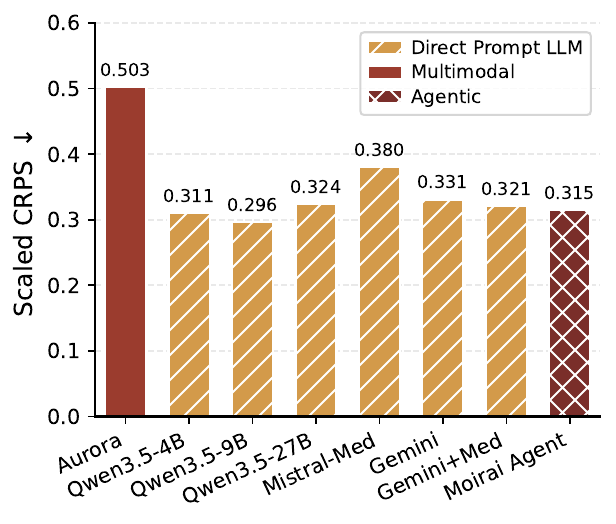}
    \vspace{-10pt}
    \captionof{figure}{Backbone comparison under \textsc{Codex}-synthesized evidence.}
    \label{fig:bar_backbone}
\end{minipage}
\end{figure*}

We evaluate each deep research strategy independently of the downstream forecaster.
For each task, we measure \emph{evidence recall}, \emph{supporting document recall}, and \emph{distractor avoidance} (see \S~\ref{app:metrics}).
Table~\ref{tab:agent_retrieval_metrics_transposed} shows large variation in evidence recall, with only \textsc{Codex} reaching 27.9\%, showing that recovering the ground-truth supporting evidence in Dr-CiK is non-trivial.
Supporting document recall consistently exceeds evidence recall, indicating that agents locate the right sources more easily than they extract forecast-relevant content.
\textsc{Codex} also leads in distractor avoidance, yet at 42.3\% still admits most distractors, making rejection of distracting context another major bottleneck.
Figure~\ref{fig:distractor_citation_patterns} shows that retrieved distractors are dominated by counterfactual-history distractors despite balanced dataset frequencies: agents often cite document-level plausible content inconsistent with the observed historical trajectory, suggesting limited reasoning over the time series to filter retrieved context (see \S~\ref{app:distractor_citation}).
Figure~\ref{fig:dr_hops_ablation} further shows that recall drops sharply with the number of reasoning hops needed to recover supporting evidence, confirming multi-hop retrieval as a key constraint.
Finally, our qualitative analysis in \S~\ref{appx:dr_fail} shows that even when agents retrieve the right documents and avoid distractors, synthesis can lose numerical anchors and modal qualifiers.

\subsection{Level 3: Context-Aided Forecasting}

We isolate context utilization by evaluating how forecasters combine numerical history with controlled evidence.
Figure~\ref{fig:bar_backbone} compares forecasters when supplied with the same \textsc{Codex}-synthesized evidence (the best DR method in the previous experiments), isolating differences in how each architecture exploits fixed, though imperfect, textual context alongside the numerical history.
The pattern from end-to-end results holds: MoiraiAgent leads, together with DP forecasters with Qwen3.5 as the backbone.
Within the Qwen3.5 family, scaling from 4B to 27B parameters does not improve performance, suggesting that model size is not the binding constraint on context utilization at this scale.
Figure~\ref{fig:context_mode} then varies the context (Definitions in \S~\ref{appx:context_modes}) provided to a single text-consuming forecaster (DP-Gemini) \rev{on a subset }across eight conditions, isolating which properties of the provided context translate into forecast gains.
Three observations follow. First, ground-truth supporting context, whether supplied as the supporting evidence, original context (from CAF source), or the rewritten context (task generation pipeline step 3's output), establishes a clear ceiling at roughly half the no-context sCRPS. The small gap among these three conditions confirms that our Entity Disambiguation Phase preserves the forecast-relevant content.
Second, raw \textsc{Codex}-synthesized evidence is statistically indistinguishable from No Context: even the strongest current DR agent provides no measurable benefit without verification or filtering.
Filtering its output to the verified subset (Verif. DR) recovers a substantial fraction of the headroom, closing roughly half the gap to the oracle condition.
Third, feeding the model the raw supporting documents (Supp. docs) degrades performance well below No Context, suggesting that an upstream synthesizer is necessary: forecasters do not extract and assemble supporting evidence from raw documents on their own.

Beyond these eight conditions, expert-curated supporting evidence carries an additional structural distinction: each causal chain contains \emph{retrospective} evidence, explaining patterns already visible in the historical window, and \emph{prospective} evidence, describing the future window as if it had not yet been observed. Both roles occur naturally in context-aided forecasting; we surface this distinction wherever the input source's evidence is authored as an explicit causal chain, which is currently the 80 expert-curated tasks. \S~\ref{app:evidence_roles} reports DR Evidence Recall and forecasting accuracy broken down by evidence role, including \rev{a} retrospective-only condition that approximates a more deployment-realistic setting in which no information about the future is available in advance.

\section{Discussion}
\label{sec:discussion}

\paragraphtight{Key takeaways.}
Dr-CiK reveals that the main bottleneck in CAF via DR is not whether forecasters can use context, but whether agents can find the right context. Ground-truth supporting evidence reduces sCRPS by more than 2$\times$ than other context conditions \rev{(Figure~\ref{fig:context_mode})}, showing that correctly identified evidence can substantially improve forecasting.
Second, current DR agents often fail to recover this evidence and instead retrieve plausible distractors, causing DR-synthesized context to perform no better, and sometimes worse, than no context. Thus, retrieval quality cannot be judged only by topical relevance or citation plausibility; it must be evaluated by downstream forecasting utility. Third, filtering distractors improves performance but leaves a large gap to the ground-truth context ceiling, indicating that the dominant remaining failure is incomplete supporting-evidence recovery. Overall, Dr-CiK shows that CAF via DR requires not merely good retrievers, but systems that search, filter, and synthesize evidence according to its expected utility for prediction.

\paragraphtight{Limitations.}
We identify four primary limitations framing our results.
First, Dr-CiK evaluates DR on a local synthetic corpus rather than the open web. Lacking real-world source conflicts or provenance challenges, our benchmark likely represents an upper bound on deployment performance.
Second, residual errors in expert-curated source contexts may propagate into the ground-truth supporting evidence, bounding achievable accuracy.
Third, despite robust contamination mitigation strategies (entity anonymization, time-shifting), models may still exploit memorized series-level patterns from pretraining data when temporal alignment is preserved.
Fourth, our reliance on LLM judges for evidence recall and document validation introduces potential systematic biases that small-scale human validation might not fully capture, bounding the certainty of our aggregate metrics.
Future work could expand the benchmark with tasks from broader datasets (e.g., CAMEF~\citep{zhang2025camef} and TIME-IMM~\citep{chang2025time}).
A promising extension for Dr-CiK is incorporating purely numerical entries, such as tables and raw time series, alongside text to evaluate agents' ability to extract and synthesize supporting evidence from these sources.

\section*{AI Use Statement}

Generative AI tools were used for benchmark construction, including document and evidence generation, LLM-based judging and quality control, and agentic auditing. They were also used to assist with manuscript editing. All AI-assisted outputs were reviewed and validated by the authors, who take full responsibility for the final content, claims, results, and artifacts.

\section*{Ethics Statement}

Dr-CiK is a benchmark and evaluation framework; it does not release a deployable forecasting or retrieval system, so we do not foresee it introducing new societal harms beyond those already attached to the publicly available LLM and retrieval components it evaluates. All time-series sources used to build the benchmark are publicly available and permit redistribution under their respective licenses (Table~\ref{tab:licenses}, \S\ref{app:benchmark-composition}), and we provide the required attribution for each source in the benchmark metadata. None of the underlying time series contains personally identifiable information: all sources publish aggregate statistics or pre-anonymized measurements, and we additionally apply entity anonymization when constructing supporting evidence and distractors. The expert-curated split (\S\ref{app:expert-motivation}) involved domain experts who authored context for real, unedited time series; these experts were compensated at or above the minimum hourly rate in their region of work. We do not foresee this work raising other ethical concerns regarding human subjects, dual use, or conflicts of interest.

\section*{Reproducibility Statement}

Dr-CiK is available at the anonymized link provided in the introduction, together with the generated corpus, task annotations, and evaluation code. The task-generation pipeline is described in \S\ref{sec:method} and \S\ref{app:env-gen-pipeline}, including corpus and evidence validation (\S\ref{app:corpus_validation}) and the human-calibrated LLM-judge protocol (\S\ref{appx:judge_details}). The forecast-dependent distractor taxonomy is defined in \S\ref{sec:method} and further exemplified in Appendix~\ref{appx:distractor}. Dataset composition, sources, and licenses are detailed in \S\ref{app:benchmark-composition}, and fine-grained difficulty annotations are described in \S\ref{appendix:difficulty} and \S\ref{appx:difficulty_annotations}. Full experimental setup, including evaluated DR agents, forecasters, and context conditions, is given in \S\ref{app:experimental-setup} and \S\ref{appx:context_modes}, and the evidence-recovery and forecasting metrics are defined in \S\ref{app:metrics}. Prompts used throughout the pipeline are provided in Appendix~\ref{appendix:prompts}.

\bibliographystyle{servicenow}
\bibliography{ref}

\clearpage

\appendix
\begin{minipage}{\textwidth}
\setstretch{1}
\setlength{\parskip}{5pt}
\setlength{\parindent}{2pt}
\textbf{\huge Appendix}\par
\vspace{6pt}
\textbf{Table of Contents}\par
\noindent\rule{\linewidth}{0.4pt}

\textbf{\hyperref[app:benchmark-composition]{A Benchmark Composition}}
\dotfill \hyperref[app:benchmark-composition]{\pageref{app:benchmark-composition}}

\hspace{0.5cm}\hyperref[app:expert-motivation]{A.1 Expert-Crafted Context-Aided Forecasting Tasks}
\dotfill \hyperref[app:expert-motivation]{\pageref{app:expert-motivation}}

\hspace{1.0cm}\hyperref[app:expert-task-creation]{A.1.1 Task creation}
\dotfill \hyperref[app:expert-task-creation]{\pageref{app:expert-task-creation}}

\hspace{1.0cm}\hyperref[app:expert-ts-stats]{A.1.2 Time series statistics and pre-processing}
\dotfill \hyperref[app:expert-ts-stats]{\pageref{app:expert-ts-stats}}

\hspace{0.5cm}\hyperref[app:scalable-subset]{A.2 Scalable Synthetic Subset}
\dotfill \hyperref[app:scalable-subset]{\pageref{app:scalable-subset}}

\textbf{\hyperref[appendix:difficulty]{B Fine-Grained Diagnostic Dimensions}}
\dotfill \hyperref[appendix:difficulty]{\pageref{appendix:difficulty}}

\textbf{\hyperref[appx:difficulty_annotations]{C Difficulty Annotations}}
\dotfill \hyperref[appx:difficulty_annotations]{\pageref{appx:difficulty_annotations}}

\textbf{\hyperref[app:env-gen-pipeline]{D Environment Generation Pipeline}}
\dotfill \hyperref[app:env-gen-pipeline]{\pageref{app:env-gen-pipeline}}

\hspace{0.5cm}\hyperref[app:corpus_validation]{D.1 Corpus and Evidence Validation}
\dotfill \hyperref[app:corpus_validation]{\pageref{app:corpus_validation}}

\hspace{0.5cm}\hyperref[appx:judge_details]{D.2 Human-Calibrated LLM Judges}
\dotfill \hyperref[appx:judge_details]{\pageref{appx:judge_details}}

\textbf{\hyperref[appx:distractor]{E Distractor Examples}}
\dotfill \hyperref[appx:distractor]{\pageref{appx:distractor}}

\hspace{0.5cm}\hyperref[appx:distractor-task-a]{E.1 Task A: 4-Week Moving Average of Initial Claims}
\dotfill \hyperref[appx:distractor-task-a]{\pageref{appx:distractor-task-a}}

\hspace{0.5cm}\hyperref[appx:distractor-task-b]{E.2 Task B: Global Price of a Standard Industrial Raw Material}
\dotfill \hyperref[appx:distractor-task-b]{\pageref{appx:distractor-task-b}}

\textbf{\hyperref[appx:context_modes]{F Context Conditions}}
\dotfill \hyperref[appx:context_modes]{\pageref{appx:context_modes}}

\textbf{\hyperref[app:experimental-setup]{G Implementation Details}}
\dotfill \hyperref[app:experimental-setup]{\pageref{app:experimental-setup}}

\textbf{\hyperref[app:metrics]{H Evaluation Metrics}}
\dotfill \hyperref[app:metrics]{\pageref{app:metrics}}

\hspace{0.5cm}\hyperref[app:metrics-dr]{H.1 Deep Research Metrics}
\dotfill \hyperref[app:metrics-dr]{\pageref{app:metrics-dr}}

\hspace{0.5cm}\hyperref[app:metrics-forecast]{H.2 Forecasting Metrics}
\dotfill \hyperref[app:metrics-forecast]{\pageref{app:metrics-forecast}}

\textbf{\hyperref[app:distractor_citation]{I DR Agents Distractor Patterns}}
\dotfill \hyperref[app:distractor_citation]{\pageref{app:distractor_citation}}

\hspace{0.5cm}\hyperref[subsec:dr-in-vs-cross]{I.1 In-Task vs Cross-Task Distractor Citation}
\dotfill \hyperref[subsec:dr-in-vs-cross]{\pageref{subsec:dr-in-vs-cross}}

\hspace{0.5cm}\hyperref[subsec:dr-subtypes]{I.2 Distractor Citation by Taxonomy}
\dotfill \hyperref[subsec:dr-subtypes]{\pageref{subsec:dr-subtypes}}

\textbf{\hyperref[appx:dr_fail]{J DR Agent Failure-Mode Analysis}}
\dotfill \hyperref[appx:dr_fail]{\pageref{appx:dr_fail}}

\hspace{0.5cm}\hyperref[appx:task216:synth]{J.1 \textsc{Codex}-Synthesized Evidence}
\dotfill \hyperref[appx:task216:synth]{\pageref{appx:task216:synth}}

\hspace{0.5cm}\hyperref[appx:task216:per_item]{J.2 Item-by-Item Failure Analysis}
\dotfill \hyperref[appx:task216:per_item]{\pageref{appx:task216:per_item}}

\hspace{0.5cm}\hyperref[appx:task216:distractors]{J.3 Distractor Docs Were Not Avoided}
\dotfill \hyperref[appx:task216:distractors]{\pageref{appx:task216:distractors}}

\hspace{0.5cm}\hyperref[appx:task216:summary]{J.4 What the Failure Reveals}
\dotfill \hyperref[appx:task216:summary]{\pageref{appx:task216:summary}}

\textbf{\hyperref[appx:additional_caf_results]{K Additional Context-Aided Forecasting Results}}
\dotfill \hyperref[appx:additional_caf_results]{\pageref{appx:additional_caf_results}}

\hspace{0.5cm}\hyperref[app:by-source-results]{K.1 Deep Research and Forecasting Quality by Input Source}
\dotfill \hyperref[app:by-source-results]{\pageref{app:by-source-results}}

\hspace{0.5cm}\hyperref[app:evidence_roles]{K.2 Evidence Role: Retrospective vs. Prospective}
\dotfill \hyperref[app:evidence_roles]{\pageref{app:evidence_roles}}

\textbf{\hyperref[appendix:prompts]{L Prompts}}
\dotfill \hyperref[appendix:prompts]{\pageref{appendix:prompts}}

\hspace{0.5cm}\hyperref[app:prompts-generation]{L.1 Generation Prompts}
\dotfill \hyperref[app:prompts-generation]{\pageref{app:prompts-generation}}

\hspace{0.5cm}\hyperref[app:prompts-judges]{L.2 LLM Judge Prompts}
\dotfill \hyperref[app:prompts-judges]{\pageref{app:prompts-judges}}

\hspace{0.5cm}\hyperref[app:prompts-difficulty]{L.3 Task-Difficulty Labeling Prompts}
\dotfill \hyperref[app:prompts-difficulty]{\pageref{app:prompts-difficulty}}

\hspace{0.5cm}\hyperref[app:prompts-audit]{L.4 Final Agent Audit Prompt}
\dotfill \hyperref[app:prompts-audit]{\pageref{app:prompts-audit}}
\par\noindent\rule{\linewidth}{0.4pt}
\end{minipage}

\clearpage

\section{Benchmark Composition}
\label{app:benchmark-composition}

Dr-CiK's 279 tasks are produced by a single scalable pipeline (\S\ref{sec:method}) applied to two evidence sources: CAF-sourced (\S\ref{app:scalable-subset}) and expert-curated series (\S\ref{app:expert-motivation}). Table~\ref{tab:dataset-card} summarizes both sources and their combination.

\begin{table}[h]
\centering
\small
\setlength{\tabcolsep}{4pt}
\caption{Dataset card (median [range]). Agents search a shared corpus of 10,342 documents across 279 tasks: 3,367 supporting documents and 6,975 distractors, with exactly 25 distractors per task.}
\label{tab:dataset-card}
\resizebox{\linewidth}{!}{
\begin{tabular}{lrrlrrrr}
\toprule
\textbf{Source} & \textbf{Tasks} & \textbf{Docs} & \textbf{Frequencies} & \makecell{\textbf{History}\\\textbf{length}} & \makecell{\textbf{Forecast}\\\textbf{horizon}} & \makecell{\textbf{Supp. docs}\\\textbf{/task}} & \makecell{\textbf{Distractors}\\\textbf{/task}} \\
\midrule
CAF-sourced & 199 & 7,302 & 1s; 1/5min; 1h; 1d & 168 [48--462] & 56 [24--240] & 11 [5--49] & 25 \\
Expert-curated & 80 & 3,040 & 13min; 1/6/26h; 1/7/28/31/90/365/366d & 160 [33--6,528] & 69.5 [15--2,798] & 13 [9--20] & 25 \\
\midrule
\textbf{Overall} & \textbf{279} & \textbf{10,342} & --- & 168 [33--6,528] & 61 [15--2,798] & 12 [5--49] & 25 \\
\bottomrule
\end{tabular}
}
\end{table}

\subsection{Expert-curated Tasks}
\label{app:expert-motivation}

The Context is Key benchmark~\citep{williams2024context} established context-aided forecasting as a measurable task, but several of its design choices limit its suitability as a target for retrieval.
CiK tasks deviate from data an agent retrieving from real sources would encounter in at least one of the following ways:
(1) they rely on synthetic time series or (2) on real series modified to reflect the supplied context, resulting in patterns that lack the richness of series produced by real-world processes, (3) contexts are not written by domain experts, hence do not necessarily reflect the register, terminology, or level of detail that a practitioner would write.
We therefore apply the same scalable generation pipeline to a complementary evidence source: real, unedited series, context written by domain experts in their own register.
This makes the expert-curated portion of Dr-CiK's tasks suitable for evaluating whether agents can discover context in settings where the relationship between context and series is closer to real-world deployment.

\subsubsection{Task creation}
\label{app:expert-task-creation}

\begin{table}[h]
\centering
\small
\caption{Per-domain summary of the collected expert-curated tasks. We report average history and forecast lengths in time steps.}
\label{tab:domain-summary}
\vspace{5pt}
\begin{tabular}{lclcc}
\toprule
\textbf{Domain} & \textbf{Tasks} & \textbf{Frequencies} & \textbf{History} & \textbf{Forecast} \\
\midrule
Finance \& economics       & 21 & B, D, W, ME, MS                 & 1{,}747 & 749 \\
Environment \& oceanography & 20 & 1h, D                          & 1{,}503 & 645 \\
Transport                  & 16 & D, W, MS, QS                    & 107   & 47  \\
Healthcare                 & 13 & W, YS                           & 105   & 45  \\
Observability              & 10 & 13min, 1h, 6h, 26h              & 190   & 82  \\
\midrule
All                        & 80 & ---                             & 897   & 385 \\
\bottomrule
\end{tabular}
\end{table}

The expert-curated task pool comprises $80$ forecasting tasks built from unique time series drawn from \rev{six} publicly available sources, spanning five domains: environmental and oceanographic monitoring, finance and economics, healthcare, infrastructure observability, and transportation.
(see Table~\ref{tab:domain-summary} for a summary of characteristics).
We then contracted SuperAnnotate for crafting the corresponding contexts.
A domain expert was commissioned to author a forecasting task by identifying meaningful patterns in the series and explaining their causes through structured causal chains.
For that, the context creator was provided with a time-series plot, with both historical and future windows, and background information, a brief description of the quantity of interest.
For each task, they were instructed to identify meaningful temporal patterns and compose two causal chains of events that could plausibly explain them:
one explaining a pattern in the historical data window and one predicting a pattern in the future window (hence using the future tense).
Each causal chain contains one or multiple sequences of factors, from root causes to observable impacts, passing through intermediate effects.

\paragraphtight{Pattern examples.}
To anchor annotators on the kinds of patterns worth explaining, we provided a non-exhaustive list of nine pattern types: sustained regime change (a persistent shift to a new level or behavior), mean shift, slope or trend change, variance shift, seasonality change (in amplitude, phase, or shape), cycle-length change, spikes and drops (transient deviations), anomalies (isolated values incompatible with the surrounding behavior), and missing data segments.
Annotators were free to identify patterns outside the list.

\paragraphtight{From causal chains to benchmark tasks.}
Each completed task consists of a time series, a description of the quantity of interest, and two causal chains: one historical and one forecast-oriented.
The construction pipeline (Section~\ref{app:env-gen-pipeline}) transforms this material into the benchmark instance presented to agents: it derives supporting-evidence units from the causal chains, generates supporting documents that instantiate them, creates distractor documents that share surface features with supporting documents but are rejectable, and records difficulty annotations indicating which steps require temporal reasoning, time-series reasoning, or topical rejection.

\subsubsection{Time series statistics and pre-processing}
\label{app:expert-ts-stats}

We source the expert-curated time series from the datasets summarized in Table~\ref{tab:source-summary}, with licenses and redistribution permissions listed in Table~\ref{tab:licenses}.
A central design choice of the expert-curated tasks is to select series whose forecast windows contain changes, such as anomalies, regime shifts, or other events that cannot be predicted from history alone.
This ensures that each task has a non-trivial forecast that requires supporting context beyond the historical series.
We also seek diversity across domain, sampling frequency, and sequence length, so that the expert-curated tasks cover a broad range of real-world dynamics.

We preprocess series by resampling with mean aggregation to the target frequency and applying a 70/30 historical/forecast split.
Two sources receive date adjustment as a contamination mitigation: Melbourne pedestrian counts and Eurostat rail are shifted forward, projecting their date ranges into a window not covered by current model training cutoffs.

\begin{table}[t]
\centering
\footnotesize
\setlength{\tabcolsep}{5pt}
\renewcommand{\arraystretch}{1.12}
\caption{Per-source summary of the collected expert-curated tasks.}
\vspace{-0.5em}
\label{tab:source-summary}
\begin{tabularx}{\linewidth}{@{}l c >{\raggedright\arraybackslash}X@{}}
\toprule
\textbf{Source} & \textbf{Tasks} & \textbf{Datasets} \\
\midrule
It's TIME~\citep{qiao2026s}
& 54
& Coastal\_T\_S, Crypto, Global\_Influenza, Global\_Price, JOLTS, Job\_Claims, Oil\_Price, Port\_Activity, SG\_Carpark, SG\_PM25, US\_Term\_Structure, Uncertainty\_1M, Water\_Quality\_Darwin, azure2019, current\_velocity, Finland\_Traffic \\

BOOM~\citep{cohen2025timedifferentobservabilityperspective}
& 9
& ds-141, ds-1082, ds-1324, ds-1396, ds-1424, ds-1440, ds-1501, ds-1761, ds-1849 \\

Eurostat
& \rev{6}
& rail\_pa\_quartal, rail\_go\_quartal \\

OECD Health Statistics
& 5
& DISCHARGE, STAY, BED\_DAY \\

BTS TranStats
& 4
& ATL, MSP, OH, SEA \\

Monash~\citep{godahewa2021monash}
& 2
& Melbourne\_Pedestrian \\
\bottomrule
\end{tabularx}
\vspace{-1em}
\end{table}

\begin{table}[t]
\centering
\footnotesize
\caption{Data sources, licenses, and redistribution permissions for the benchmark sources. All sources permit redistribution under their respective terms; we include the required attribution for each source in the benchmark metadata.}
\label{tab:licenses}
\vspace{5pt}
\begin{tabularx}{\linewidth}{
    p{2.8cm}
    p{2.8cm}
    p{1.2cm}
    >{\raggedright\arraybackslash}X
}
\toprule
\textbf{Source} & \textbf{License} & \textbf{Redist.} & \textbf{License reference} \\
\midrule
CiK & Apache 2.0 & Permitted &
\url{huggingface.co/datasets/ServiceNow/context-is-key} \\
\addlinespace

It's TIME & Apache 2.0 & Permitted &
\url{huggingface.co/datasets/Real-TSF/TIME} \\
\addlinespace

Datadog BOOM & Apache 2.0 & Permitted &
\url{huggingface.co/datasets/Datadog/BOOM} \\
\addlinespace

Eurostat & CC BY 4.0 & Permitted &
\url{ec.europa.eu/eurostat/help/copyright-notice} \\
\addlinespace

OECD Health Statistics & OECD Terms and Conditions & Permitted &
\url{oecd.org/en/about/terms-conditions.html} \\
\addlinespace

BTS TranStats & US Public Domain (17 U.S.C.~\S105) & Permitted &
17 U.S.C.~\S105 \\
\addlinespace

Monash Repository & CC BY 4.0 & Permitted &
\url{huggingface.co/datasets/Monash-University/monash_tsf} \\
\bottomrule
\end{tabularx}

\end{table}

\subsection{CAF-Sourced Tasks}
\label{app:scalable-subset}

Applying the same scalable pipeline to prior CAF benchmarks yields 199 tasks derived from CiK~\citep{williams2024context} and GIFT-CTX\citep{salesforce2026giftctx}.
We construct this source by prioritizing task families that reflect the core reasoning demands of Dr-CiK: forecasting under context-driven distribution shifts such as event-induced demand changes, operational outages and maintenance, anomaly correction, holiday-conditioned traffic variation, and weather-state transitions.
We exclude constraint-only items, verbose near-duplicate variants, and distributionally narrow reserve-style families that are less representative of the target task.
To avoid overlap between sources, we also remove GIFT-CTX examples whose raw source field indicates CiK provenance, so that the GIFT-CTX portion provides complementary rather than duplicated coverage.

\section{Fine-Grained Diagnostic Dimensions}
\label{appendix:difficulty}

Beyond the number of supporting documents and distractors, Dr-CiK annotates four additional sources of task difficulty.
These dimensions capture distinct reasoning challenges in converting retrieved text into forecast-useful evidence (Figure~\ref{fig:difficulty_annotations}), enabling aggregate performance to be decomposed into interpretable failure profiles.

\paragraphtight{Domain Knowledge.}
This dimension measures whether a causal step can be resolved with common-sense knowledge or requires specialized expertise.
We label each causal edge as \emph{General} (e.g., \emph{heat causes increased air conditioning use}) or \emph{Specialist} (e.g., \emph{freshwater-driven water-column stratification traps fine particles near the surface and amplifies turbidity readings}).
Specialist edges test whether an agent can go beyond identifying a trigger and correctly infer its domain-specific effect on the target series.

\paragraphtight{Temporal Complexity.}
This dimension measures the temporal reasoning needed to map a causal step onto the forecast horizon.
We use three levels.
\emph{Straightforward} effects are immediate and bounded, such as a storm causing turbidity to spike within hours.
\emph{Predictable} effects involve delayed onset, moderate duration, or recurrence, such as a wet season shifting turbidity into an elevated regime over several weeks.
\emph{Variable} effects are long-lagged, cumulative, or irregularly recurring, such as sediment accumulation amplifying storm-driven turbidity peaks.
Higher levels require agents to align triggers with the forecast window and reason about accumulated effects across events.

\paragraphtight{Explicitness.}
This dimension measures how directly the relevant information is stated in a document chunk.
\emph{Explicit} chunks name the target quantity, trigger, and effect (e.g., \emph{turbidity at the Darwin buoy spiked to 162 NTU following Tropical Cyclone Marcus}).
\emph{Implied} chunks mention the trigger but leave the target effect qualitative or indirect (e.g., \emph{Cyclone Marcus generated 4--6m wave heights across the Timor Sea}).
\emph{Implicit} chunks describe relevant conditions without naming the target, requiring cross-document or background reasoning (e.g., \emph{the 2017--18 wet season was one of the most active cyclone seasons in the northern Australian basin in a decade}).
Lower explicitness penalizes agents that rely on surface overlap rather than causal inference.

\paragraphtight{Certainty.}
This dimension captures the evidence's epistemic status, independent of how explicitly it is stated.
\emph{Certain} chunks report observations with negligible ambiguity (e.g., \emph{measured 162 NTU at 14:15 on December 7th}).
\emph{Likely} chunks use probabilistic language or modeled estimates (e.g., \emph{wave heights estimated at 3-5m, suggesting turbidity likely exceeded 100 NTU}).
\emph{Uncertain} chunks compound uncertainties or describe projections (e.g., \emph{projected increases in cyclone intensity could push peak turbidity beyond historical maxima next season}).
This dimension tests whether agents and forecasters propagate uncertainty rather than treating retrieved evidence as equally reliable.

\paragraphtight{Coverage Across Difficulty Levels.}
These dimensions motivate Dr-CiK's two-source design.
The CAF-sourced tasks provide broad coverage across difficulty levels, while the expert-curated tasks concentrate on harder cases: Nearly all expert-curated tasks contain at least one Specialist step, and most exhibit Variable temporal effects.
Together, the two sources evaluate both routine context use and harder cases requiring domain knowledge, temporal reasoning, and uncertainty handling.

\paragraphtight{Preliminary Observations.}
\rev{Conditioning results on these diagnostic dimensions reveals patterns that validate the labels themselves: DR Evidence Recall declines across most dimensions, confirming that harder-labeled tasks are indeed harder for agents to research. Table~\ref{tab:difficulty-stratified} (\S\ref{appx:additional_caf_results}) reports the complete per-level breakdown across all four dimensions. The Domain Knowledge axis shows the cleanest joint effect: Specialist tasks yield both lower Evidence Recall (22.9\% vs.\ 39.6\% for General) and higher end-to-end sCRPS (0.340 vs.\ 0.308), indicating that domain-specific causal knowledge is a genuine bottleneck for both retrieval and downstream forecasting. End-to-end sCRPS itself does not track difficulty as cleanly on the other axes, since incomplete evidence recovery interacts with each forecaster's own sensitivity to the retrieved context.}

\section{Difficulty Annotations}
\label{appx:difficulty_annotations}

\begin{figure}[h]
    \centering
    \includegraphics[width=\linewidth]{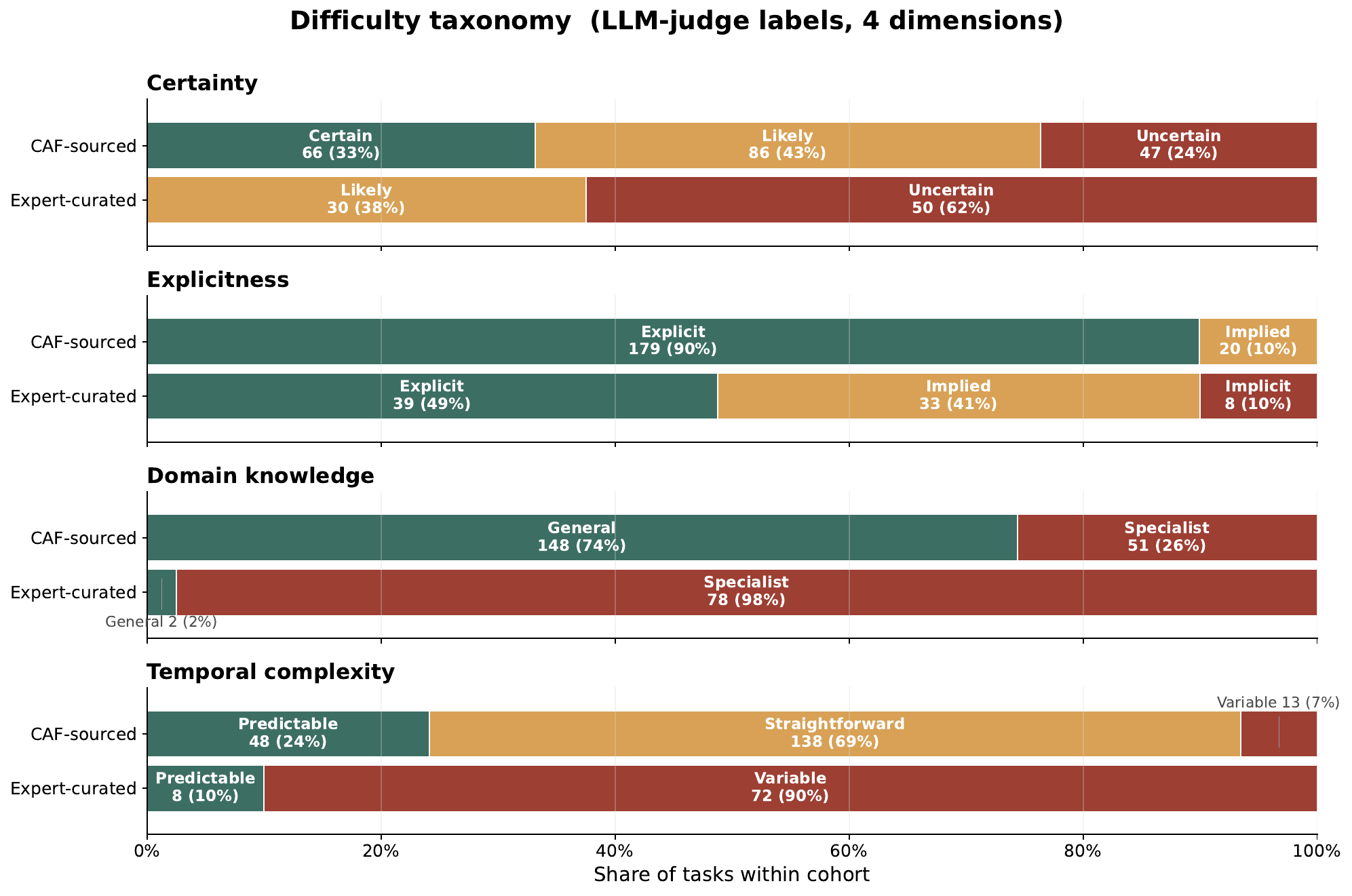}
    \caption{\textbf{Difficulty annotations for CAF-sourced and expert-curated tasks.}
    Each row compares the distribution of a difficulty dimension across the CAF-sourced tasks and the expert-curated tasks.
    Expert-curated tasks are consistently harder: they contain a larger share of uncertain evidence, implied rather than explicit context, specialist domain knowledge, and variable temporal structure.
    In contrast, the CAF-sourced tasks provide broader coverage but are more often certain, explicit, general-domain, and temporally straightforward.}
    \label{fig:difficulty_annotations}
\end{figure}

To characterize the reasoning demands of Dr-CiK, we annotate each task along four difficulty dimensions: evidence certainty, context explicitness, domain knowledge, and temporal complexity.
These annotations are used for diagnostic analysis rather than filtering: they allow us to relate DR and forecasting failures to the kinds of reasoning required by each task.
Figure~\ref{fig:difficulty_annotations} shows expert-curated tasks are substantially harder than the CAF-sourced tasks across multiple dimensions. In CAF-sourced tasks, supporting evidence is more often certain and explicit, tasks more often rely on general rather than specialist knowledge, and temporal effects are usually straightforward. By contrast, expert-curated tasks contain more uncertainty, more implied evidence, more specialist knowledge, and more variable temporal structure. This confirms that the two evidence sources play complementary roles within the same scalable pipeline: CAF-sourced tasks provide broad coverage and statistical power, while expert-curated tasks stress deployment-realistic reasoning, where relevant context is less direct, less certain, and harder to operationalize for forecasting.

\section{Environment Generation Pipeline}
\label{app:env-gen-pipeline}

\begin{figure}[h!]
    \centering
    \includegraphics[width=\linewidth]{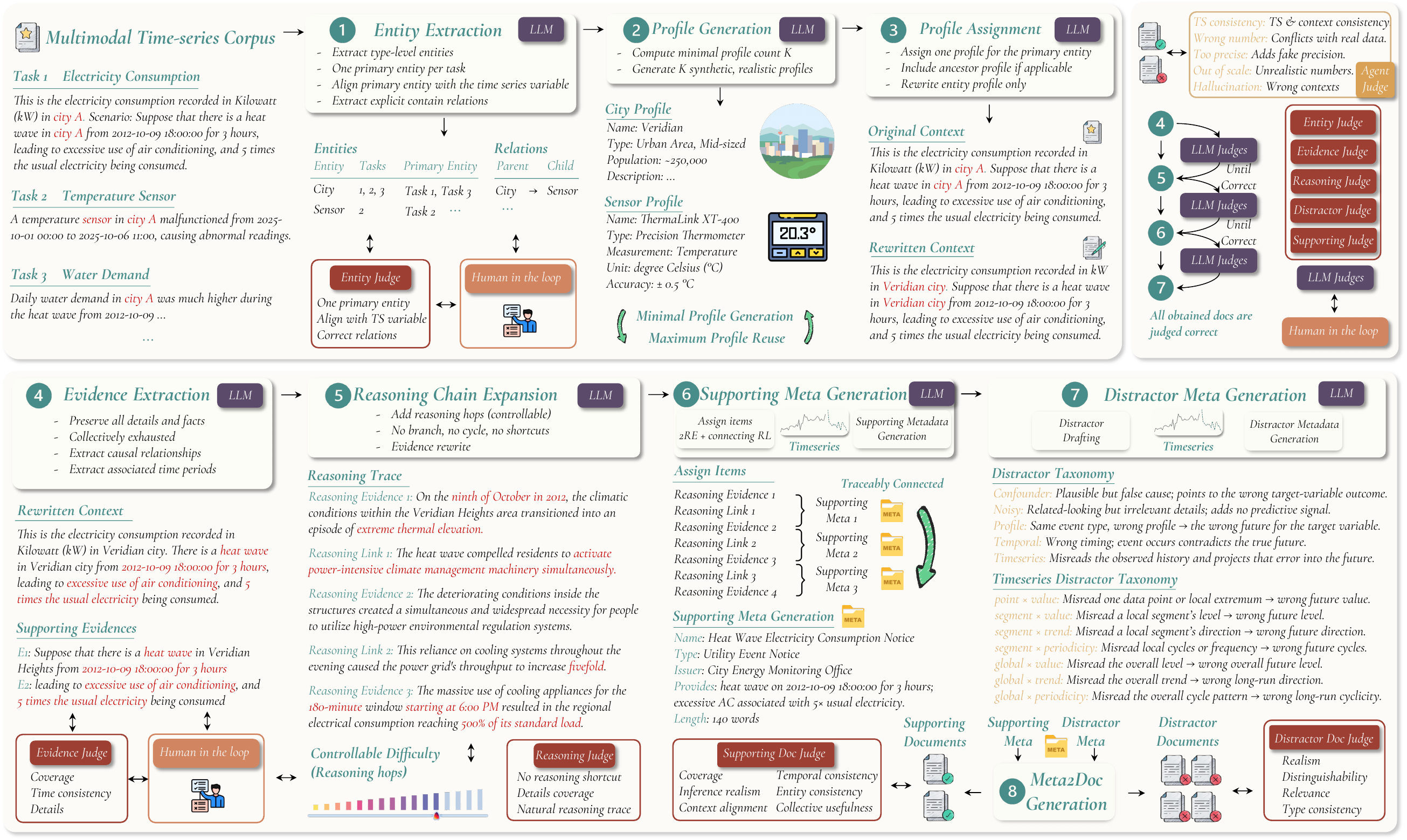}
    \caption{\textbf{Worked Example of the DR Environment Generation Pipeline.} This expanded view instantiates the pipeline in Figure~\ref{fig:workflow} on a concrete CAF task. Starting from a multimodal time-series instance, the pipeline \textbf{(1)} extracts entities, assigns synthetic profiles, and rewrites the original context to reduce entity ambiguity and memorization shortcuts, \textbf{(2)} decomposes the rewritten supporting context into supporting-evidence units and expands them into a controllable multi-hop reasoning chain, and \textbf{(3)} converts the resulting reasoning trace into supporting documents while generating forecast-dependent distractors from the distractor taxonomy. The right-hand audit loop shows the local LLM judges and final Agent Judge used to verify intermediate outputs and repair residual inconsistencies with human oversight.}
    \label{fig:workflow_example}
\end{figure}

\subsection{Corpus and Evidence Validation}
\label{app:corpus_validation}

Two programmatic audits and one ablation validate that the generated corpus faithfully carries the inherited evidence rather than fabricating or losing it.

\paragraph{Cloze audit on supporting documents.} We task an LLM with reconstructing masked ground-truth evidence values and entities from the generated supporting documents alone, to confirm no evidence is lost during document generation. Of 908 blanks across 96 tasks, 769 were immediately correct, 131 partial, and 8 wrong, \rev{for an initial overall mean score of 0.9372}; all 139 incorrect blanks were repaired via minimal edits to 148 documents and reconfirmed correct (139/139)\rev{, yielding a final score of 908/908}.

\begin{table}[h]
\centering
\small
\caption{Supporting-document cloze audit.}
\label{tab:cloze-audit}
\begin{tabular}{lrrr}
\toprule
\textbf{Category} & \textbf{Blanks} & \textbf{Mean score} & \textbf{Correct / Partial / Wrong} \\
\midrule
Forecast driver & 280 & 0.9188 & 221 / 58 / 1 \\
Driver timing & 164 & 0.9177 & 135 / 27 / 2 \\
Driver effect & 283 & 0.9611 & 252 / 30 / 1 \\
Persistence/counterfactual & 105 & 0.9619 & 98 / 4 / 3 \\
Quantitative anchor & 76 & 0.9243 & 63 / 12 / 1 \\
\midrule
\textbf{Overall} & \textbf{908} & \textbf{\rev{0.9372}} & 769 / 131 / 8 \\
\bottomrule
\end{tabular}
\end{table}

\paragraph{Distractor distinguishability audit.} For each of the 6,975 released distractors, we check whether an agent seeing only its test-time inputs (history series, forecasting objective, and the candidate document) can find an explicit cue to reject it as non-evidence, combining deterministic rule passes (domain/entity mismatch, out-of-window dates, disqualifier language, subtype-specific markers) with agentic and human-in-the-loop checks on ambiguous cases.

\begin{table}[h]
\centering
\small
\caption{Distractor distinguishability audit (6,975 distractors, 279 tasks).}
\label{tab:distractor-distinguishability}
\begin{tabular}{lr}
\toprule
\textbf{Outcome} & \textbf{Documents} \\
\midrule
Cleared by rule check (explicit cue present) & 6,583 \\
Flagged as ambiguous $\to$ cleared on stricter re-check & 386 \\
Manually inspected $\to$ cleared & 6 \\
\midrule
\textbf{Distinguishable from supporting evidence} & \textbf{6,975 / 6,975 (100\%)} \\
\bottomrule
\end{tabular}
\end{table}

\paragraph{Raw source context vs. Dr-CiK corpus.} A possible concern is that retrieval in Dr-CiK could collapse to trivial entity/topic matching if source contexts were used directly, since CiK's near-duplicate seed variants per base task are easy to disambiguate by entity alone. We run the \textsc{Retrieval} and \textsc{Codex} agents over 15 stratified CiK-origin targets under two corpora: (A) the 355 raw CiK task-variant contexts before Dr-CiK augmentation, and (B) our released 7,302-document Dr-CiK corpus. This is the ablation referenced in \S\ref{sec:method}, confirming that existing CAF resources cannot directly supply a non-trivial DR environment.

\begin{table}[h]
\centering
\small
\caption{Deep-research ablation on raw CiK vs. Dr-CiK corpus.}
\label{tab:corpus-ablation}
\begin{tabular}{lrrrr}
\toprule
\textbf{Metric} & \textbf{Retrieval A} & \textbf{Retrieval B} & \textbf{\textsc{Codex} A} & \textbf{\textsc{Codex} B} \\
\midrule
Evidence Recall $\uparrow$ & \textbf{0.513} & 0.035 & \textbf{0.731} & 0.329 \\
Support Doc Recall (strict) $\uparrow$ & 0.867 & 0.071 & \textbf{1.000} & 0.605 \\
Distractor Avoidance $\uparrow$ & 0.087 & \textbf{0.727} & 0.094 & \textbf{0.503} \\
\bottomrule
\end{tabular}
\end{table}

On corpus A, the target's own document is retrieved trivially with no headroom to test DR capability, and Distractor Avoidance collapses because CiK's near-duplicate seed variants cannot be disambiguated. On corpus B, all three metrics separate the agents, confirming that Dr-CiK's document generation, not the raw source context, is what makes retrieval non-trivial.

\subsection{Human-Calibrated LLM Judges}
\label{appx:judge_details}

Dr-CiK uses LLM judges to verify outputs during task generation, but we do not rely on them out of the box. Instead, we calibrate each judge against human review before using it in the generation pipeline. The goal of calibration is conservative filtering: a selected LLM judge should agree with the human majority when possible, and otherwise be stricter rather than more permissive.

\begin{figure}[h]
    \centering
    \includegraphics[width=.9\linewidth]{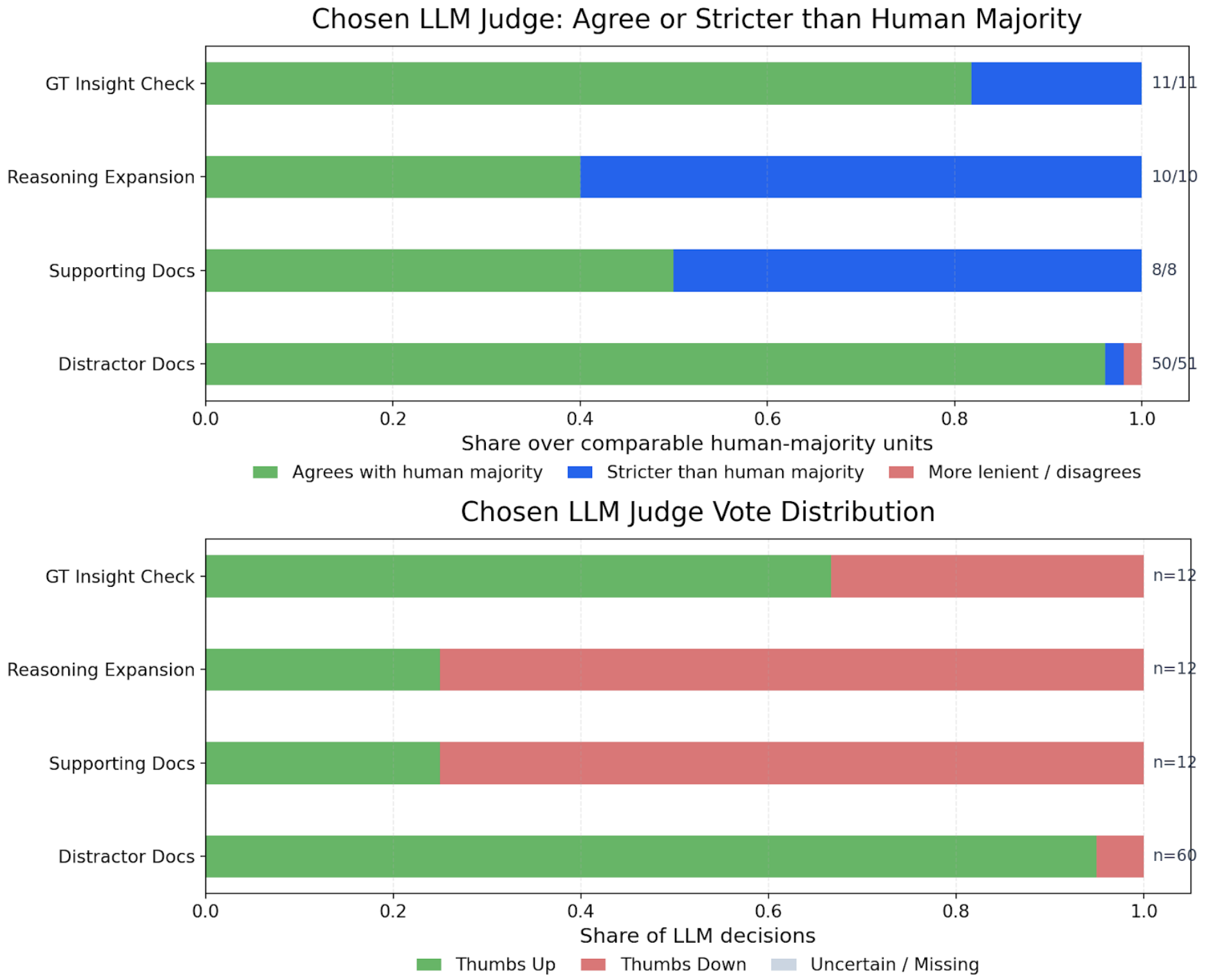}
    \caption{\textbf{Calibration of selected LLM judges against human majority labels.}
    For each generation stage, we compare the selected LLM judge against human majority decisions on a held-out calibration set.
    The top panel reports the share of comparable units where the chosen LLM judge agrees with the human majority, is stricter than the human majority, or is more lenient/disagrees.
    The bottom panel shows the final vote distribution of the chosen LLM judge.
    The selected judges are calibrated to be conservative: when they do not agree with the human majority, they are intended to reject rather than accept borderline outputs.}
    \label{fig:judge_calibration}
\end{figure}

\begin{figure}[t]
    \centering
    \includegraphics[width=\linewidth]{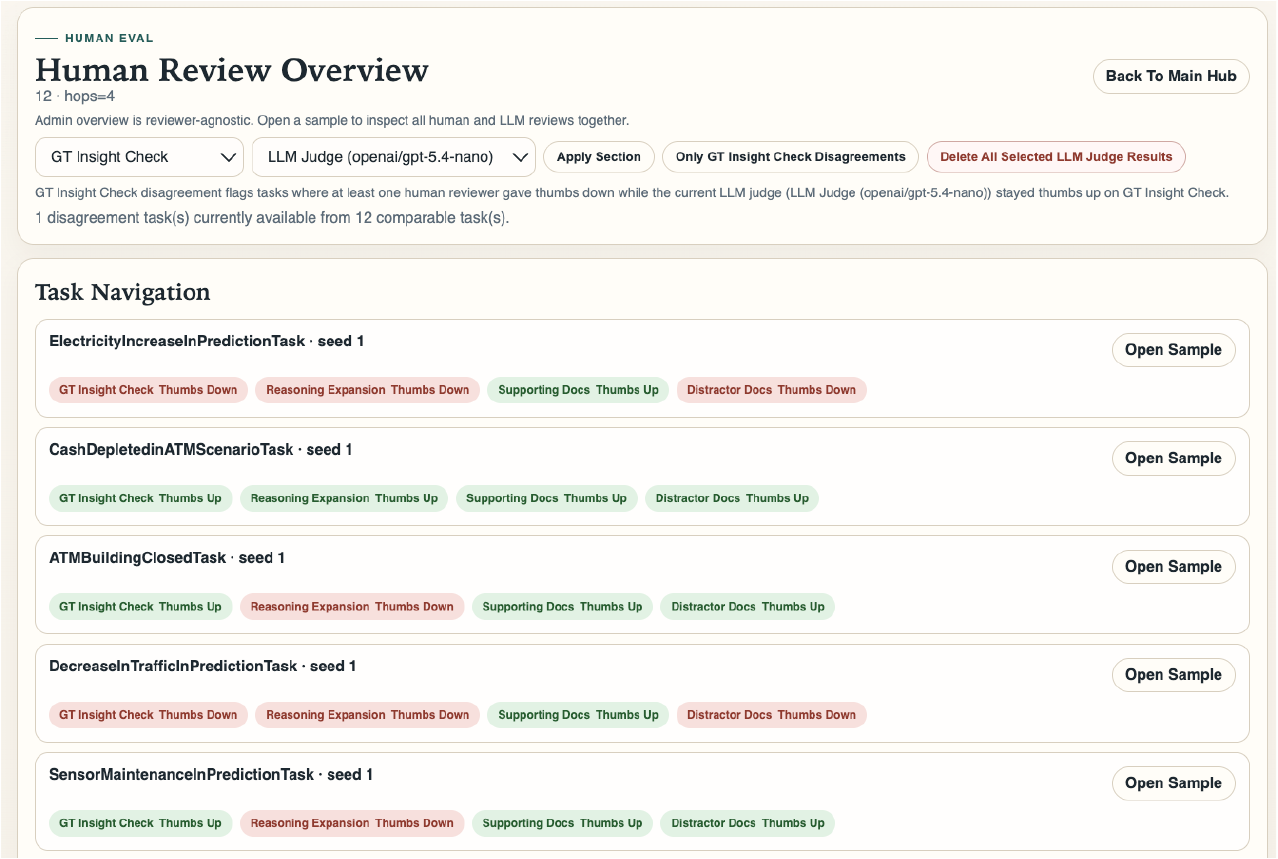}
    \caption{\textbf{Human review dashboard.}
    The dashboard summarizes human and LLM-judge decisions across sampled tasks and generation stages.
    Reviewers can filter by stage, inspect disagreement cases, and open individual samples for detailed comparison.
    This interface is used to collect human majority labels for calibrating the LLM judges before full-scale benchmark generation.}
    \label{fig:human_label_dashboard}
\end{figure}

\begin{figure}[t]
    \centering
    \includegraphics[width=\linewidth]{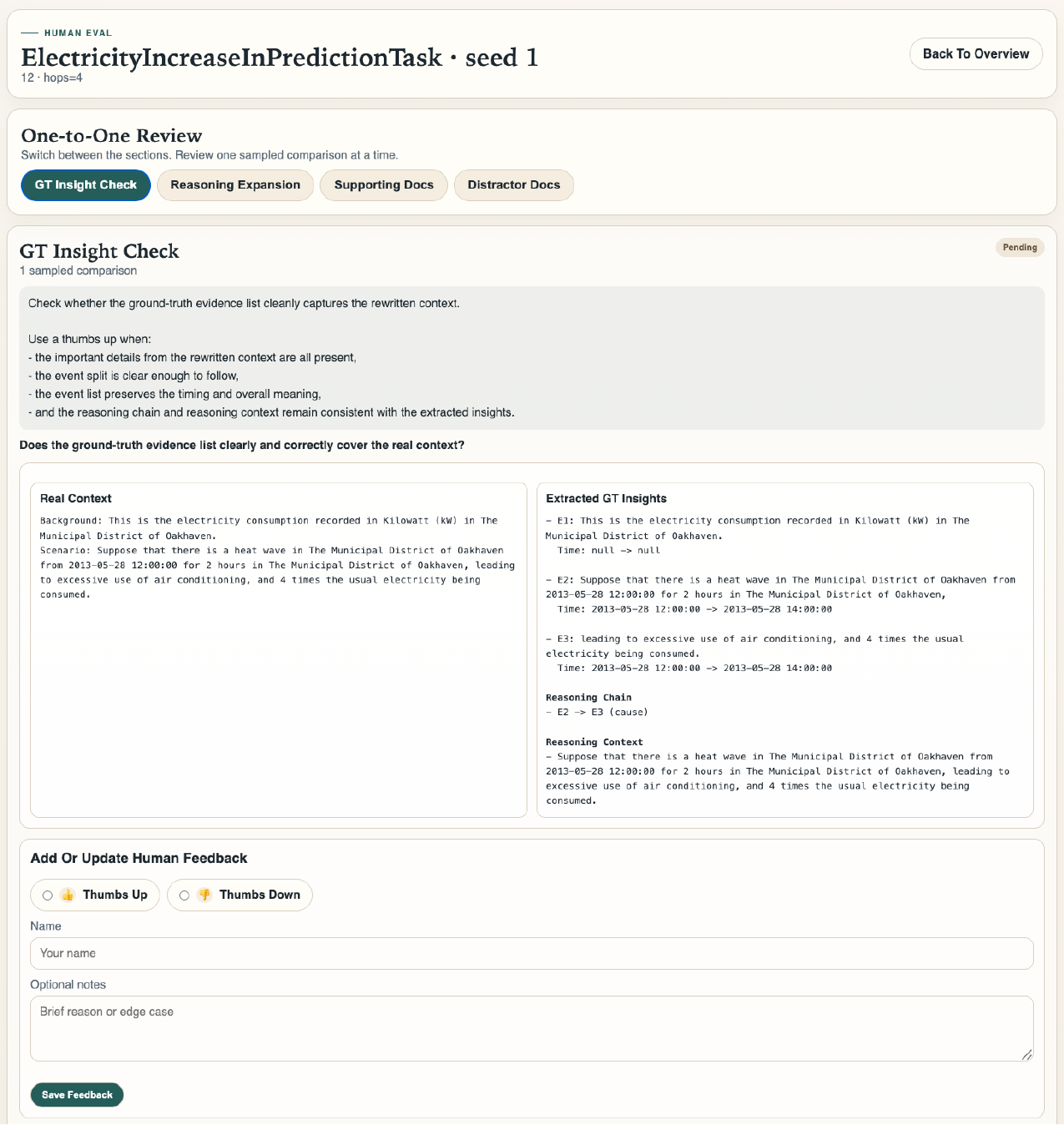}
    \caption{\textbf{Detailed human review interface.}
    For each sampled comparison, reviewers see the relevant source input and generated output side by side, along with the stage-specific review instructions.
    Reviewers mark the output as acceptable or unacceptable and may add notes explaining the failure mode.
    These annotations provide the human reference labels used to tune and select conservative LLM judges.}
    \label{fig:human_label_detail}
\end{figure}

\paragraph{Human calibration set.}
We first generate a representative set of preliminary tasks and ask human reviewers to label the major generation stages: ground-truth supporting-evidence extraction, reasoning-chain expansion, supporting-document generation, and distractor-document generation.
Across 12 calibration tasks and four generation stages, seven reviewers provided 117 human review records, comprising 79 stage-level judgments and 38 fine-grained per-entry judgments.
Figure~\ref{fig:human_label_dashboard} shows the reviewer dashboard, and Figure~\ref{fig:human_label_detail} shows the detailed one-to-one review interface. For each stage, reviewers inspect the relevant inputs and outputs, then assign a binary judgment indicating whether the output is acceptable. When an output is marked incorrect, reviewers can provide free-form notes describing the failure mode.

\paragraph{Judge calibration.}
For each generation stage, we design a specialized LLM judge with stage-specific acceptance criteria. These include an Evidence Judge for checking whether supporting-evidence units cover the rewritten context, a Reasoning Judge for validating reasoning-chain structure and completeness, a Supporting Document Judge for verifying that generated supporting documents preserve the intended evidence, and a Distractor Judge for checking realism, relevance, and rejectability of distractor documents. We iteratively refine each judge prompt until the selected LLM judge is at least as strict as the human majority on the calibration set. As shown in Figure~\ref{fig:judge_calibration}, the final selected judges either agree with the human majority or reject more outputs than the human majority in nearly all comparable cases. Full judge prompts and rubrics are provided in \S~\ref{appendix:prompts}.

\paragraph{Regeneration after calibration.}
The preliminary tasks used for calibration are not included in the final benchmark. After calibration, we discard those tasks and regenerate the benchmark from scratch. During this final synthesis run, each pipeline stage is guarded by its corresponding calibrated LLM judge. We use a \emph{generate-until-correct} procedure: if an intermediate output fails its judge, that output is regenerated and rechecked until it passes. This procedure is applied separately to evidence extraction, reasoning expansion, supporting-document generation, and distractor-document generation. Each stage retries for up to 5 rounds. In practice, most units pass on the first attempt (e.g., 98.6\% of ground-truth-evidence extractions and 94.0\% of supporting-document generations), and any remaining failures are flagged for human repair. The prompts used for each generator and judge are listed in \S~\ref{appendix:prompts}.

\paragraph{Judge models.}
After calibration, we use a fixed judge assignment for full-scale generation. Entity validation is performed by \texttt{google/gemini-3-flash-preview}\footnote{\url{https://openrouter.ai/google/gemini-3-flash-preview}}. Ground-truth evidence extraction and supporting-document validation use \texttt{openai/gpt-5.4-nano}\footnote{\url{https://openrouter.ai/openai/gpt-5.4-nano}}, while reasoning-chain validation and distractor-document validation use \texttt{openai/gpt-5.4-mini}\footnote{\url{https://openrouter.ai/openai/gpt-5.4-mini}}. For downstream evaluation, evidence-recall matching is judged by \texttt{google/gemini-3-flash-preview}. The final global audit is not part of the automatic generation pipeline: we use Claude Code\footnote{\url{https://www.anthropic.com/claude-code}} as an agentic auditor to inspect the completed environment, flag cross-document inconsistencies or leakage, and guide targeted human repairs.

\paragraph{Global agent audit.}
Local judges verify individual stages, but some errors only become visible at the completed-environment level, such as inconsistencies across documents, leakage between supporting and distractor documents, or conflicts between the generated corpus and the time-series metadata. We therefore apply a final Agent Judge to audit the complete DR environment after all documents are generated. The Agent Judge checks global consistency, source-document alignment, distractor rejectability, and whether the final corpus supports the intended CAF via DR task. Flagged issues are repaired with human oversight. The Agent Judge prompt is also provided in \S~\ref{appendix:prompts}.

\paragraph{Judge robustness to judge-model choice.} To assess robustness to judge-model choice, we repeat the Evidence Recall evaluation using independent judge models (same instructions and input; only semantic matching is judge-dependent), on a stratified 40-task multi-judge sample (1,518 ground-truth-evidence-item judgments per judge, pooled over all 5 DR methods, one independent judge per major lab). New judges agree with our default judge 88--92\% of the time, with the exception of \texttt{gpt-5.6-luna}, which is stricter and requires near-literal coverage of evidence. Importantly, the DR-method ranking is identical under every judge. Notably, judges with different reasoning configurations (e.g., \texttt{claude-sonnet-5}) show agreement rates comparable to non-reasoning judges, indicating that judge reliability is not sensitive to this factor either.

\begin{table}[h]
\centering
\small
\caption{Multi-judge sensitivity.}
\label{tab:multi-judge}
\begin{tabular}{lcc}
\toprule
\textbf{Judge (organization)} & \textbf{Pooled ER $\uparrow$} & \makecell{\textbf{Raw item-level agreement}\\\textbf{w/ default (agreeing/total)}} \\
\midrule
gemini-3-flash-preview (Google, default) & 14.4\% & --- \\
claude-sonnet-5 (Anthropic) & 16.3\% & 91.4\% (1,388/1,518) \\
gpt-5.6-luna (OpenAI) & 11.1\% & 92.0\% (1,396/1,518) \\
kimi-k2-0905 (Moonshot) & 17.5\% & 88.4\% (1,342/1,518) \\
\bottomrule
\end{tabular}
\end{table}

\clearpage

\section{Distractor Examples}
\label{appx:distractor}

For each example below, we display the time series and task background separately and reproduce here only the distractor document content, its type, and an analysis of why the document fits that type. The two source tasks (\texttt{task\_A}: 4-week moving average of Initial Claims, \texttt{task\_B}: Global Price of a Standard Industrial Raw Material) together cover all five distractor classes and every $\mathit{scope}\!\times\!\mathit{feature}$ counterfactual-history sub-combination that appears in the released bundle.

\subsection{Task A: 4-Week Moving Average of Initial Claims}
\label{appx:distractor-task-a}

\begin{figure}[h]
    \centering
    \includegraphics[width=\linewidth]{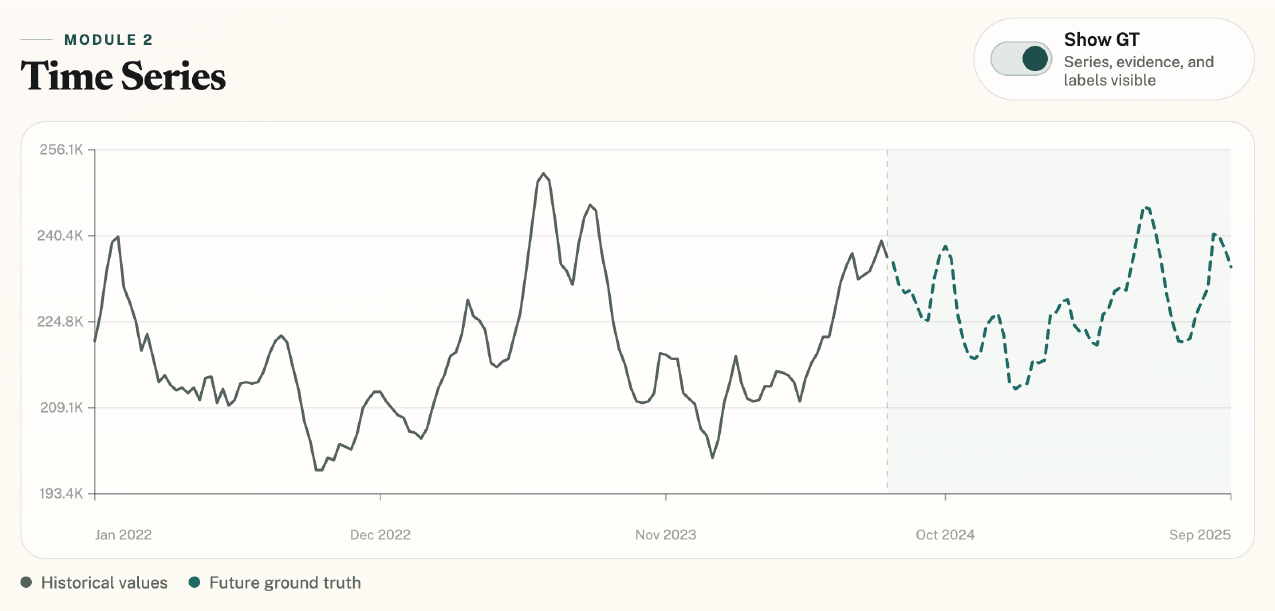}
    \caption{Task A: 4-week moving average of Initial Claims, weekly from Jan. 2022 to Sept. 2025.}
    \label{fig:taskA}
\end{figure}

\paragraph{Background.}
The Urban-Industrial Employment Ledger tracks the 4-week moving average of Initial Claims, sampled weekly from January 2022 to September 2025.

\paragraph{Ground-truth Supporting Evidence.}
The task carries two human-labeled causal chains, one explaining the history (\textsc{HC1}) and one specifying the projected future (\textsc{HC2}), both anchored to a shared background event (\textsc{E1}). Distractor analyses below should be read against this ground truth.

\vspace{5pt}

\begin{tcolorbox}[title=Task A: Initial Claims, breakable]
\begin{minted}{markdown}
History mechanism

1. The central regulatory body's aggressive monetary tightening cycle, raising the base lending rate from approximately 0.25% in early 2022 to approximately 5.3% by mid-2023, created competing pressures on the Urban-Industrial Employment Ledger.
2. Post-pandemic retention strategies, where firms kept staff despite slowing demand to avoid rehiring costs in a tight environment, suppressed initial claims filing rates below structural equilibrium levels. This phenomenon, combined with historically low resignation rates in 2022, drove the 4-week moving average from ~240,000 toward the ~200,000 trough observed in early 2023, well below the pre-pandemic structural baseline of ~220,000.
3. As these tight conditions in the Urban-Industrial Employment Ledger began to reverse under monetary tightening, the rate hikes began transmitting into tighter financial conditions by mid-2023. Rate-sensitive sectors, particularly technology, real estate, and financial services, initiated significant layoff cycles. This sector-specific shock drove the sharp spike from ~200,000 to ~250,000 between January and July 2023, while more resilient sectors like healthcare and government employment partially offset the broader deterioration.
4. As these layoffs spread and financial conditions remained restrictive, the early-2024 reabsorption of workers from tech and financial sectors into healthcare, logistics, and services only partly offset the broader pass-through of restrictive credit conditions. As tighter financing and slower hiring spread beyond the first wave of rate-sensitive layoffs, initial claims rose again toward the mid-230,000s by mid-2024 rather than stabilizing at the early-2024 trough.
5. The 4-week moving average of initial claims within the Urban-Industrial Employment Ledger fell from roughly 240,000 in early 2022 to near 200,000 by early 2023, rebounded to about 250,000 by mid-2023 as rate-sensitive layoffs intensified, eased back toward roughly 200,000-205,000 around early 2024, and then climbed again toward roughly 235,000-240,000 by mid-2024 as restrictive financial conditions continued to broaden across employers.

Forecast mechanism

1. Restrictive monetary policy continues to maintain elevated borrowing costs, with the lagged transmission of high interest rates into business investment and hiring decisions still working through the economy.
2. The sustained inversion of the treasury yield curve through 2024 is projected to continue compressing net interest margins for regional banks, tightening credit availability for small and medium enterprises.
3. Since these enterprises account for a significant part of the Urban-Industrial Employment Ledger, credit-constrained hiring and episodic layoffs in this segment are expected to generate recurring upward spikes.
4. Given this tightening in credit conditions, sectors that over-hired during the 2021-2022 expansion, particularly technology, e-commerce, and professional services, are projected to continue rightsizing their workforces, contributing to episodic layoff waves that can drive the claims above 240,000.
5. However, the rightsizing-driven layoffs remain sector-specific rather than economy-wide. Continued strength in healthcare, government, and infrastructure hiring partially offsets these losses, keeping the oscillations in the Urban-Industrial Employment Ledger from developing into a sustained uptrend, so that spikes toward 240,000-246,000 remain temporary rather than establishing a new structural baseline.
6. The 4-week moving average of initial claims in the Urban-Industrial Employment Ledger is projected to fluctuate within an elevated range through September 2025, with most weekly readings expected to fall between 220,000-235,000, reaching a low near 213,000 and a high near 246,000 only at the extremes of the oscillation, with a periodicity of approximately 6-8 weeks and no sustained directional trend.
\end{minted}
\end{tcolorbox}

\paragraph{Off-variable.}
\begin{quote}\small
``Beginning in early 2022, the Central Regulatory Body transitioned its primary regulatory focus toward the integration and enforcement of carbon-neutral manufacturing standards. \ldots\ current increases in worker participation in mandatory retraining programs are primarily driven by the implementation of requirements associated with the green energy transition.''

\end{quote}
The document preserves the same entity and time window and presents a plausible causal mechanism. However, the mechanism terminates on a different variable, namely worker participation in mandatory retraining programs, rather than the 4-week moving average of Initial Claims. It therefore provides no causal support for the target series.

\paragraph{Irrelevant.}
\begin{quote}\small
``This manual establishes the protocols for the administration, data integrity, and systematic auditing of the Urban-Industrial Employment Ledger. \ldots\ Reporting must encompass all active personnel changes, including new hires, terminations, and shifts in industrial labor hours.''
\end{quote}
The document is realistic, on-topic at the surface, refers to the same ledger and same agency, and is full of detail, but its content is exclusively administrative: classification tiers, submission windows, and digital-transition protocols. None of it carries forecast-relevant information about claims behaviour. The document tests whether the agent anchors on realistic peripheral filler instead of separating supporting evidence from noise.

\paragraph{Off-entity.}
\begin{quote}\small
``\textbf{Issuer:} Metropolitan Workforce Registry. \ldots\ Labor volatility \ldots\ is being fundamentally reshaped by a dual transition \ldots\ green energy sectors \ldots\ industrial automation. \ldots\ the 4-week moving average of initial claims remains atypically depressed, consistently tracking within a narrow range between 180,000 and 195,000.''
\end{quote}
The scenario template, labor ledger, 4-week moving average of initial claims, and contemporary timeframe, is preserved, but the entity is swapped to a different same-type registry, \textit{Metropolitan Workforce Registry}, with different drivers, automation and skill-matching efficiency. The implied future is a tight 180k to 195k band, contradicting the realized future, which fluctuates in the 213k to 246k range. This is the canonical off-entity distractor: maximal scenario-level overlap, contradicting the forecast.

\paragraph{Off-window.}
\begin{quote}\small
``This analysis examines the specific interactions between monetary policy adjustments \ldots\ during the critical \textbf{2015 to 2016} window \ldots\ This tightening cycle moved the benchmark interest rates from a baseline of 0 percent to a peak of 0.75 percent.''
\end{quote}
The cause type, monetary tightening, and entity, the same ledger, are correct, but the entire event is anchored to 2015 to 2016, which lies before the historical observation window from 2022 to 2025. The rate path described, 0\% to 0.75\%, is also inconsistent with the 0.25\% to 5.3\% trajectory governing the actual forecast window. Same mechanism, wrong temporal anchor: the document has no predictive bearing on the prediction window.

\paragraph{Counterfactual history, $\mathit{global}\!\times\!\mathit{value}$.}
\begin{quote}\small
``A longitudinal review of the Urban-Industrial Employment Ledger reveals a distinct pattern of \textbf{diminishing peaks} \ldots\ future oscillations in the 4-week moving average of initial claims are expected to dampen significantly.''
\end{quote}
The document misreads the global level and extremum profile of the history. Empirically, peaks taken over rolling 8-week windows are roughly stable around 240k to 252k, not diminishing, with \texttt{240k} early, \texttt{251k} mid-history, and \texttt{236k} late. Propagating the false ``dampening'' assertion forward produces a forecast that is too tightly bracketed near the mean, missing the realized peaks up to 246k.

\paragraph{Counterfactual history, $\mathit{segment}\!\times\!\mathit{periodicity}$.}
\begin{quote}\small
``\ldots\ a persistent departure from symmetric oscillation, favoring a distinct asymmetric profile \ldots\ a gradual build and a slow, incremental climb in filing rates \ldots\ filing rates do not gradually recede, instead, they are observed collapsing back to the floor immediately after a peak is reached.''
\end{quote}
The document misstates the periodic structure of local segments, replacing the actual symmetric multi-week oscillation, weekly autocorrelation $\approx 0.40$ at lag~7 and $0.46$ at lag~6, with an asymmetric ``slow rise / vertical drop'' sawtooth. An agent that internalises this segment-level periodicity description will project long upward runs ending in abrupt collapses, which does not match the realized mostly-symmetric fluctuations.

\paragraph{Counterfactual history, $\mathit{global}\!\times\!\mathit{trend}$.}
\begin{quote}\small
``\ldots\ have formally entered a phase of \textbf{secular expansion} \ldots\ Each successive market trough \ldots\ is now being established at a progressively higher level than the previous one. \ldots\ projected to consistently break above the 250,000 threshold.''
\end{quote}
The document misreads the global trend, asserting a sustained ascending trajectory and a breakout above 250k. The actual history is range-bound: troughs over rolling 8-week windows do not monotonically rise, they vary between $\sim$197k and $\sim$236k, and the realized future stays within 213k to 246k, never breaking 250k.

\paragraph{Counterfactual history, $\mathit{global}\!\times\!\mathit{periodicity}$.}
\begin{quote}\small
``\ldots\ characterized by high-momentum phases where directional shifts \ldots\ persist once established. \ldots\ the prevailing time series lacks short-term cyclicality or mid-year reversals \ldots\ projected to continue their ascent and break past the 260,000 level.''
\end{quote}
The document misreads the global periodicity by asserting that the series is non-oscillatory and lacks short-term cyclicality, and uses this to justify a monotonic forecast above 260k. This contradicts the observed periodicity, autocorrelation $\approx 0.63$ at lag~4 and $0.40$ at lag~7, and the realized future never approaches 260k. The error is global because the claim is made about the overall regime, not a local segment.

\subsection{Task B: Global Price of a Standard Industrial Raw Material}
\label{appx:distractor-task-b}

\begin{figure}[h]
    \centering
    \includegraphics[width=\linewidth]{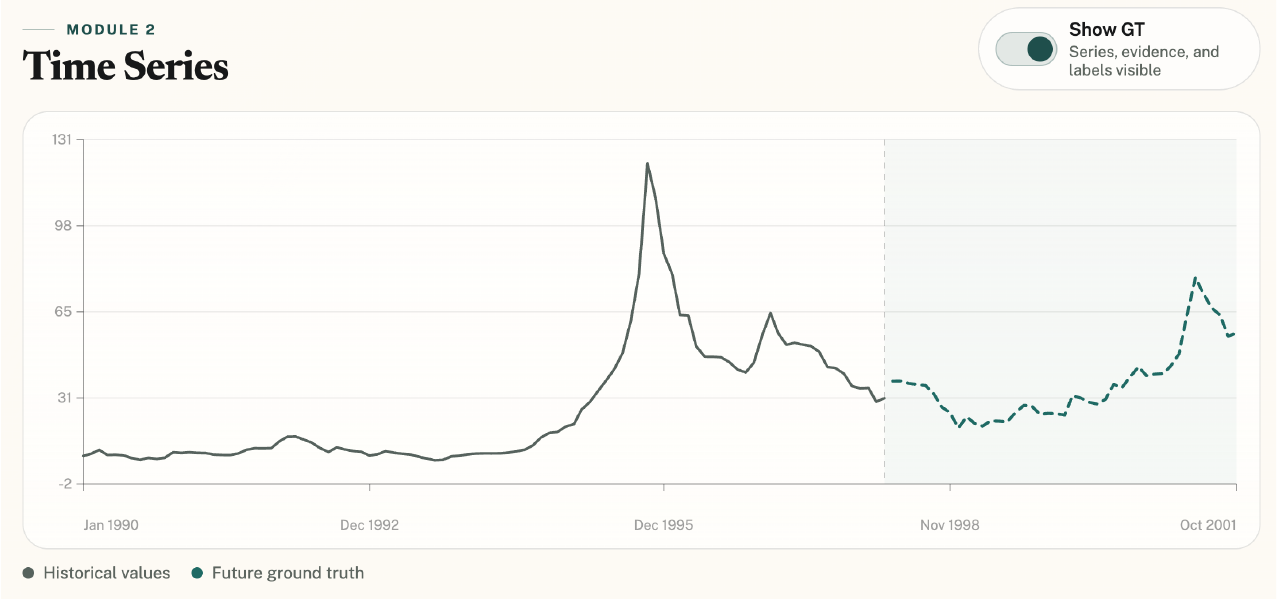}
    \caption{Task B: quarterly global price of a Standard Industrial Raw Material from January 1990 to October 2001.}
    \label{fig:taskB}
\end{figure}

\paragraph{Background.}
This series tracks the quarterly global market price of a Standard Industrial Raw Material, sampled from January 1990 to October 2001.

\paragraph{Ground-truth Supporting Evidence.}
The task carries two human-labeled causal chains, one explaining the history (\textsc{HC1}) and one specifying the projected future (\textsc{HC2}), both are anchored to a shared background event (\textsc{E1}). Distractor analyses below should be read against this ground truth.

\begin{tcolorbox}[title=Task B: Standard Industrial Raw Material price, breakable]
\begin{minted}{markdown}
History mechanism

1. In the early 1990s, the market for a Standard Industrial Raw Material faced a severe primary production deficit as strategic inventories were depleted faster than new extraction operations could be commissioned. Primary global production remained consistently below annual industrial requirements, forcing the sector to rely on finite discretionary inventories that eventually reached critical lows.
2. The exhaustion of these stockpiles triggered a phase of panic-buying by industrial utilities, creating a speculative environment where buyers competed for limited uncommitted supply.
3. This led to a realized speculative price bubble peaking at approximately $120/lb in late 1995.
4. The peak was broken when a large-scale conversion program, launched in 1993, began successfully repurposing decommissioned materials into commercial fuel from around 1995, flooding the market with secondary supply.
5. Followed by a rapid market correction back toward $30/lb by March, 1998.

Forecast mechanism

1. The Standard Industrial Raw Material sector in late 1998 entered a phase of severe supply-side rationalization, as market prices dropped below the marginal cost of production for global Tier-1 extraction assets.
2. Persistent low prices are expected to force major operators to suspend high-cost production facilities, significantly tightening the available primary supply in the global market.
3. A freeze in exploration capital expenditure during the market slump is projected to prevent the development of a new project pipeline; these extraction sites typically require 7-10 years from discovery to first production due to the complexity of regulatory licensing, geological confirmation drilling, and facility construction, meaning that even a future price recovery cannot quickly unlock new supply to meet growing industrial demand.
4. As secondary supplies from conversion programs are projected to stabilize, the market is expected to recognize an impending structural deficit triggered by steady growth in reactor requirements.
5. That tightening forward balance, built by facility suspensions and the absence of new project development, is expected to draw utilities back into aggressive long-term contracting, lifting prices sharply toward approximately $80/lb by late 2001.
6. We anticipate a projected non-linear recovery, with prices finding a cyclical floor at $20/lb in 1999 before trending upward toward a secondary peak of approximately $80/lb by late 2001.
\end{minted}
\end{tcolorbox}

\paragraph{Counterfactual history, $\mathit{segment}\!\times\!\mathit{trend}$.}
\begin{quote}\small
``Following a phase of sharp price corrections, the sector has transitioned into a definitive \textbf{rounding-base behavior} \ldots\ the sector is projected to bypass previously anticipated lower cyclical floors \ldots\ a direct and sustained climb toward previous price highs \ldots\ recovery will be \textbf{linear rather than erratic}.''
\end{quote}
The document misreads the directional behavior of the late-history segment: it claims the most recent prices have already turned upward in a smooth rounding-base shape, justifying a linear recovery that bypasses any further dip. Empirically, the last 20 historical observations slope downward, with terminal value 31.20 below the window mean, providing no evidence of the claimed upward rounding-base behavior. The projected linear recovery is therefore derived from a misread of the observed recent trend. The error is local to a segment: it is a misstatement of the recent direction, not of the global regime.

\paragraph{Counterfactual history, $\mathit{segment}\!\times\!\mathit{value}$.}
\begin{quote}\small
``\ldots\ the Standard Industrial Raw Material sector has reached a \textbf{definitive structural floor} \ldots\ Recent price action has consistently validated the marginal cost level \ldots\ prices are expected to enter an immediate and sustained recovery phase \ldots\ \textbf{without further downside volatility}.''
\end{quote}
The document misstates the value level of the late-history segment by claiming that price observations have validated a structural floor at the marginal cost level. The observed series provides no such local support region: the late-history values do not cluster around or return to a stable floor before the forecast cutoff. The claimed support level is inconsistent with the historical data available at prediction time. The resulting projection of an immediate and sustained recovery is a wrong future derived from this misread historical value pattern. The misinterpretation is segment-level because it concerns the late-history portion of the series, and value-based because the error lies in asserting a local price level that is not present in the observed data.

\section{Context Conditions}
\label{appx:context_modes}

We evaluate forecasters under eight context conditions to separate the effect of background information, ground-truth supporting evidence, raw supporting documents, and DR-generated outputs.
These conditions differ only in the textual input supplied to the forecaster, the historical time series and forecast horizon are held fixed.

\begin{itemize}[leftmargin=*, itemsep=0.2em, topsep=0.2em, parsep=0pt, partopsep=0pt]
    \item \textbf{No Context (None).} The forecaster receives only the historical time series and forecast request, with no textual context.

    \item \textbf{Background (Bg.).} The forecaster receives a short generic description of the time-series variable, such as the measured quantity and unit, but no task-specific future-relevant information.

    \item \textbf{Original Context (Orig.).} The forecaster receives the original context from the source CAF instance before our task-generation transformations.

    \item \textbf{Rewritten Context (Rewrite).} The forecaster receives the context after entity disambiguation and time-shifting in the task-generation pipeline. This condition tests whether these transformations preserve the forecast-relevant content of the original context.

    \item \textbf{Supporting Evidence (Evidence).} The forecaster receives the ground-truth supporting evidence used to evaluate the DR stage. This contains the specific forecast-relevant content without the surrounding documents.

    \item \textbf{Supporting Documents (Supp. docs).} The forecaster receives the concatenation of all supporting documents for the task. This condition tests whether a forecaster can extract and assemble supporting evidence directly from raw documents.

    \item \textbf{DR-synthesized Evidence (DR).} The forecaster receives the evidence synthesized by a DR agent from the corpus. This condition reflects the raw output of the DR stage before any additional verification or filtering.

    \item \textbf{Verified DR (Verif. DR).} The forecaster receives the subset of DR-synthesized evidence that passes verification against the supporting evidence and distractor labels. This condition isolates the effect of filtering distractors and unsupported claims from the DR output.
\end{itemize}

The following shows an example of these eight context conditions for the task we used for DR failure mode analysis in \S\ref{appx:dr_fail}.

\begin{tcolorbox}[title=Example context conditions, breakable]
\begin{minted}{markdown}
## No Context (None)

```
```


---

## Background (Bg.)

```
This time series records the electricity usage (in Watt, W) of a household.
```

---

## Original Context (Orig.)

```
This time series records the electricity usage (in Watt, W) of a household. The readings were recorded from 2025-10-06 00:00:00 to 2025-10-19 19:00:00, and the recording frequency is hourly (H). Initially, the owner started renovation from 2025-10-12 18:00:00 to 2025-10-13 00:00:00 and did not stay at the unit, leading to low power usage of the first 169 readings. After the renovation finished on 2025-10-13 01:00:00, the house is fully occupied, leading to a permanent increase in electricity usage.
Scenario: Forecast the future electricity usage of this household in the next 180 hours, from 2025-10-19 20:00:00 to 2025-10-27 07:00:00.
```

---

## Rewritten Context (Rewrite)


```
This time series records the electricity usage (in Watt, W) of Riverside Cottage 12. The readings were recorded from 2025-10-06 00:00:00 to 2025-10-19 19:00:00, and the recording frequency is hourly (H). Initially, the owner started renovation from 2025-10-12 18:00:00 to 2025-10-13 00:00:00 and did not stay at the unit, leading to low power usage of the first 169 readings. After the renovation finished on 2025-10-13 01:00:00, this household is fully occupied, leading to a permanent increase in electricity usage.
Scenario: Forecast the future electricity usage of Riverside Cottage 12 in the next 180 hours, from 2025-10-19 20:00:00 to 2025-10-27 07:00:00.
```

---

## Supporting Evidence (Evidence)

```
1. This time series records the electricity usage (in Watt, W) of a household. The readings were recorded from 2025-10-06 00:00:00 to 2025-10-19 19:00:00, and the recording frequency is hourly (H).
2. the owner started rennovation from 2025-10-12 18:00:00 to 2025-10-13 00:00:00 and did not stay at the unit
3. leading to low power usage of the first 169 readings
4. After the renovation finished on 2025-10-13 01:00:00, this household is fully occupied
5. leading to a permanent increase in electricity usage
```

---

## Supporting Documents (Supp. docs)

The full mode concatenates all 14 supporting documents (about 25k characters total). 


---

## DR-synthesized evidence (DR)

```
This time series records the electricity usage (in Watt, W) of the household Riverside Cottage 12.

- Riverside Cottage 12 electricity usage is documented as Watt-based household power consumption, with specific focus on October 2025 load behavior and occupancy-driven baseline changes. (doc_7245, doc_7271)
- During the renovation and heavy-labor phase, the owner was not occupying the premises and major appliances and climate-control systems were deactivated, resulting in Wattage reductions due to a lack of human activity. (doc_7236, doc_7238)
- Construction concluded on October 13, at which point living quarters became available and the household reached full occupancy; this elevated post-handover baseline persisted through at least October 19. (doc_7241, doc_7242, doc_7243, doc_7246)
- Full occupancy is the primary driver of ongoing electricity use, linked to continuous environmental support, active residential services, and the simultaneous operation of essential systems that were formerly inactive. (doc_7244, doc_7245)
- Post-occupancy electricity use in Cottage 12 is rising, characterized by a drifting higher daily baseline and usage scaling through steady upward momentum. (doc_7259, doc_7260, doc_7261, doc_7274, doc_7275)
- A smart billing meter was installed on October 9, coinciding with a change in utility provider, followed by a safety inspection on October 15 after equipment updates. (doc_7250)
- A smart-meter calibration window from October 10 to October 15 caused temporary reporting-metric shifts and recorded spikes attributed to automated diagnostics rather than occupancy. (doc_7262)
- A localized storm and regional grid fluctuations between October 8 and October 14 caused abnormal high readings while surge protectors remained active. (doc_7263)
- Security and lighting installation from October 7 to October 12 introduced transient peak consumption due to high-frequency motion sensors and testing-related loads. (doc_7264)
- A strict energy quota system was implemented months before October 2025 to regulate peak demand and localized grid capacity, which may constrain extreme usage even under full occupancy. (doc_7258)
- Localized environmental monitoring from October 10 to October 14 recorded humidity variation, changing cloud cover, low nighttime light, and the presence of a reflective roof system designed to reduce heat gain. (doc_7251)
- The cottage features smart-home controls for temperature, lighting, security, media, and whole-home audio, which allows occupied behavior to affect both standby and active domestic electrical loads. (doc_7267)
- Logistics records for furniture and staging are inconsistent, with one document scheduling a curated furniture package for October 18 and another listing deliveries for late October. (doc_7252, doc_7265)
- There is conflicting evidence regarding usage trends: one report suggests usage is gradually declining after a post-renovation surge, while others describe sustained upward momentum after full occupancy. (doc_7276, doc_7259, doc_7260, doc_7275)
- Some documents describe a stable low-energy or flat profile with no October occupancy changes, contradicting reports of a post-October-13 elevated baseline following handover. (doc_7271, doc_7272, doc_7273, doc_7241, doc_7246)
- Evidence supports modeling a structural break in electricity usage from renovation/vacancy conditions to a fully occupied residential baseline following the October 13 handover. (doc_7238, doc_7241, doc_7242, doc_7246)
- Within the occupied regime, the directional signal indicates a higher and potentially rising baseline driven by residents' domestic routines and concurrent essential-system operation. (doc_7244, doc_7245, doc_7259, doc_7275)
- Short-lived drivers such as meter diagnostics, storms, security testing, and logistics introduce uncertainty into the operational records. (doc_7262, doc_7263, doc_7264, doc_7252, doc_7265)
```

---

## Verified DR (Verif. DR)

```
(After verification, no Codex-synthesized evidence remains.)
```
\end{minted}
\end{tcolorbox}

\section{Implementation Details}
\label{app:experimental-setup}

\paragraph{Common interface.}
All five deep-research (DR) agents are evaluated through the same per-task interface.
Each agent receives the historical time series, the forecast task instructions, and the task document corpus, which contains both supporting documents and distractor documents but does not expose their labels.
Each agent produces two outputs: a markdown research report and a structured list of synthesized evidence items.
Each synthesized evidence item contains a free-form claim and the identifiers of the source documents it cites.
Downstream forecasters consume the structured evidence list under the \texttt{deepresearch\_insights} context mode and the full report under the \texttt{deepresearch} context mode.
The agents differ only in how they search, read, and synthesize information from the shared corpus.

\paragraph{Deep-research agents.}
All DR agents are prompted with the same DP forecasting prompt plus the following research instruction: ``Search the local environment and gather the most relevant documents and evidence snippets that would help forecast the target time-series values, while avoiding irrelevant or distracting information.''
\textsc{Codex}~\citep{openai_codex_2026} wraps the OpenAI \textsc{Codex} command-line agent backed by GPT-5.5 at high reasoning effort.
It runs an iterative tool-use loop in which the model autonomously reads documents, drafts partial summaries, revises its reasoning, and commits a final report with cited synthesized evidence.
\textsc{DrBench}~\citep{abaskohi2025drbench} implements a search-summarize-synthesize cascade: it retrieves relevant passages, compresses retrieved documents into per-document briefs, and synthesizes evidence from those briefs.
\textsc{Bench2Future}~\citep{wildman2025bench} follows a ReAct-style trajectory, alternating between search actions and observations over the corpus while maintaining a running outline of the forecast-relevant context.
\textsc{OpenDR}~\citep{langchain_open_deep_research} decomposes research into planning, retrieval through ReAct-style tool calls, and report writing.
\textsc{Retrieval} is a minimal non-agentic baseline that performs a single embedding-based retrieval pass followed by a one-shot synthesis call to extract evidence items, without iterative tool use.
\textsc{Codex} uses GPT-5.5 as its underlying model with High reasoning efforts, the other four DR agents use Gemini-3 Flash.

\paragraph{Forecasters.}
We evaluate four families of forecasters.
Statistical baselines include ARIMA, SARIMA, ETS, SES, and Naive forecasting.
All statistical baselines are univariate and use only the observed history, without textual context. ARIMA is implemented with \texttt{statsmodels} by selecting the lowest-AIC model over a grid with $p \leq 3$, $d \leq 2$, and $q \leq 3$. SARIMA fixes the seasonal order at $(1,0,1,m)$, with the seasonal period $m$ set from each task's sampling frequency (e.g., $m{=}24$ for hourly, $m{=}7$ for daily, $m{=}12$ for monthly-scale data, deferring to explicit seasonal-period metadata when available), while the non-seasonal order $(p,d,q)$ is selected by the lowest-AIC model over $p,d,q\in\{0,1\}$. ETS uses \texttt{statsmodels} exponential smoothing and selects by AIC between an additive-trend model and a level-only model, both without seasonality. SES is \texttt{statsmodels} simple exponential smoothing with optimized smoothing level. The Naive baseline samples future paths by bootstrapping historical first differences from the last observed value.
Pretrained time-series models include Chronos~\citep{ansari2024chronos}, TimesFM-3~\citep{jain2026timesfm3}, Toto-2.0~\citep{khwaja2026toto}, TiRex-2~\citep{auer2026tirex}, Moirai-2.0~\citep{woo2024unified}, and Moirai-MoE~\citep{liu2024moirai}.
Multimodal and time-series models include Aurora.
Zero-shot LLM forecasters use the Direct Prompt (DP) forecasting strategy~\citep{williams2024context} with Gemini-3.1-flash-lite, Mistral-medium-3.1, Qwen-3.5 at 4B/9B/27B~\citep{qwen3.5}, Llama-3.2-3B~\citep{grattafiori2024llama}, and Phi-4-mini~\citep{abdin2024phi}.
We also evaluate MoiraiAgent~\citep{salesforce2026moiraiagent}, a forecasting agent that combines a Moirai forecasting backbone with a Gemini LLM component.

\paragraph{Context conditions.}
For end-to-end evaluation, the forecaster receives the evidence synthesized by a DR agent.
For controlled CAF analysis, we additionally evaluate forecasters under the context conditions defined in \S~\ref{appx:context_modes}, including No Context, Background, Original Context, Rewritten Context, Supporting Evidence, Supporting Documents, DR-synthesized Evidence, and Verified DR.
These conditions allow us to separate failures due to missing supporting evidence, distractor contamination, and forecasters' inability to use textual context.

\paragraph{Infrastructure.}
Closed-source language models, including GPT-5.5 for \textsc{Codex} and Gemini-3 Flash, Gemini-3.1-flash-lite, and Mistral-medium-3.1 for DR agents and zero-shot LLM forecasters, are accessed through OpenRouter\footnote{\url{https://openrouter.ai}}.
Open-weight language models, including the Qwen-3.5 family, Llama-3.2-3B, and Phi-4-mini, are served locally through vLLM\footnote{\url{https://github.com/vllm-project/vllm}}.
Time-series and multimodal forecasters, including Chronos and Aurora, are also served locally.
MoiraiAgent runs with its forecasting backbone served locally and its language-model component routed through OpenRouter.
Unless otherwise stated, each forecaster draws $S{=}25$ forecast trajectories per task.

\section{Evaluation Metrics}
\label{app:metrics}

\subsection{Deep Research Metrics}
\label{app:metrics-dr}

\paragraphtight{Notation.}
For each task $t$, let $\mathcal{E}_t$ denote the set of ground-truth supporting-evidence items evaluated for that task, excluding the initial background item.
For each supporting-evidence item $e \in \mathcal{E}_t$, let $R_e$ denote the set of supporting documents required to substantiate it.
Each DR agent produces an ordered list of synthesized evidence items.
We evaluate the top $K_t = |\mathcal{E}_t| + 5$ synthesized evidence items for task $t$.
Let $M_e \in \{0,1\}$ indicate whether the judge finds a semantic match between ground-truth supporting-evidence item $e$ and at least one synthesized evidence item in the top $K_t$.
Let $C_e$ denote the set of source documents cited by the matched synthesized evidence item or items.
For distractor metrics, let $Q_t$ denote the set of all resolved documents cited by the DR report, $D_t$ the set of distractor documents available for task $t$, and $P_t$ the full document pool available for task $t$.

\paragraphtight{Evidence Recall.}
Evidence Recall measures whether a DR agent recovers the ground-truth supporting evidence and cites the documents needed to support it:
\[
\mathrm{EvidenceRecall}
= \frac{1}{\sum_t |\mathcal{E}_t|}
  \sum_t \sum_{e \in \mathcal{E}_t}
  M_e
  \frac{|R_e \cap C_e|}{|R_e|}.
\]
This metric gives credit only when the synthesized evidence semantically matches a ground-truth supporting-evidence item, with partial credit for citing the required supporting documents.

\paragraphtight{Supporting Document Recall.}
Supporting Document Recall measures whether the agent locates the source documents that contain the ground-truth supporting evidence, independent of whether the synthesized evidence text is semantically correct:
\[
\mathrm{SuppDocRecall}
= \frac{1}{\sum_t |\mathcal{E}_t|}
  \sum_t \sum_{e \in \mathcal{E}_t}
  \frac{|R_e \cap Q_t|}{|R_e|}.
\]
This separates document-level retrieval from evidence-level synthesis.

\paragraphtight{Distractor Avoidance.}
Distractor Avoidance measures the fraction of cited documents that are not distractors:
\[
\mathrm{DistractorAvoidance}
= 1 -
\frac{|Q_t \cap D_t|}{|Q_t|}
\]
for each task with at least one resolved citation, and is averaged across tasks.
Higher values indicate that the agent cites fewer distractor documents. This metric is intentionally not normalized by the base rate of distractors in the corpus, since our goal is to measure the quality of the agent's actually cited evidence sources. Accordingly, a cited distractor counts against the score even if the agent identifies it as conflicting or irrelevant, since it remains part of the evidence presented to the downstream forecaster and may still adversely affect the forecast. Such cases are rare in our audit (11/4,002 Codex cited distractor docs), supporting citation purity as a direct operational measure of distractor avoidance.

\subsection{Forecasting Metrics}
\label{app:metrics-forecast}

For each task, let $T$ be the forecast horizon, $S$ be the number of forecast samples, $\hat{y}_{s,t}$ the $s$-th forecast sample at horizon step $t$, and $y_t$ the ground-truth value.
We normalize all errors by the mean absolute value of the target series over the forecast horizon:
\[
a \;=\; \Bigl(\,\frac{1}{T}\sum_{t=1}^{T} |y_t|\Bigr)^{-1}.
\]

\paragraphtight{Point forecast metrics.}
The point forecast is the sample mean,
\[
\bar{y}_t \;=\; \frac{1}{S}\sum_{s=1}^{S}\hat{y}_{s,t}.
\]
We report scaled mean absolute error and scaled root mean squared error:
\[
\mathrm{sMAE}
\;=\; a \cdot \frac{1}{T}\sum_{t=1}^{T}\left|\bar{y}_t - y_t\right|,
\qquad
\mathrm{sRMSE}
\;=\; a \cdot
\sqrt{\frac{1}{T}\sum_{t=1}^{T}\left(\bar{y}_t - y_t\right)^2}.
\]

\paragraphtight{Distributional forecast metric.}
For probabilistic accuracy, we compute the empirical Continuous Ranked Probability Score (CRPS) at each horizon step from the forecast samples:
\[
\mathrm{CRPS}_t
\;=\;
\frac{1}{S}\sum_{s=1}^{S}\left|\hat{y}_{s,t} - y_t\right|
\;-\;
\frac{1}{2S^2}\sum_{s=1}^{S}\sum_{s'=1}^{S}
\left|\hat{y}_{s,t} - \hat{y}_{s',t}\right|.
\]
We report the scaled CRPS averaged across the forecast horizon:
\[
\mathrm{sCRPS}
\;=\;
a \cdot \frac{1}{T}\sum_{t=1}^{T}\mathrm{CRPS}_t.
\]
Lower sCRPS indicates better distributional forecasts.

\paragraphtight{Aggregation across tasks.}
For each forecaster-context condition, we report the mean metric value over tasks together with the standard error of the mean across tasks.
Following standard winsorization for heavy-tailed benchmark distributions, per-task values of $\mathrm{sMAE}$, $\mathrm{sRMSE}$, and $\mathrm{sCRPS}$ exceeding $5$ are clipped to $5$ independently for each metric.
We also report an \textsc{Average Rank} metric computed per task across all evaluated cells: ranking by $\mathrm{sRMSE}$ gives the point-estimate rank, and ranking by $\mathrm{sCRPS}$ gives the distributional rank.

\section{DR Agents Distractor Patterns}
\label{app:distractor_citation}

\begin{figure}[h]
    \centering
    \includegraphics[width=\linewidth]{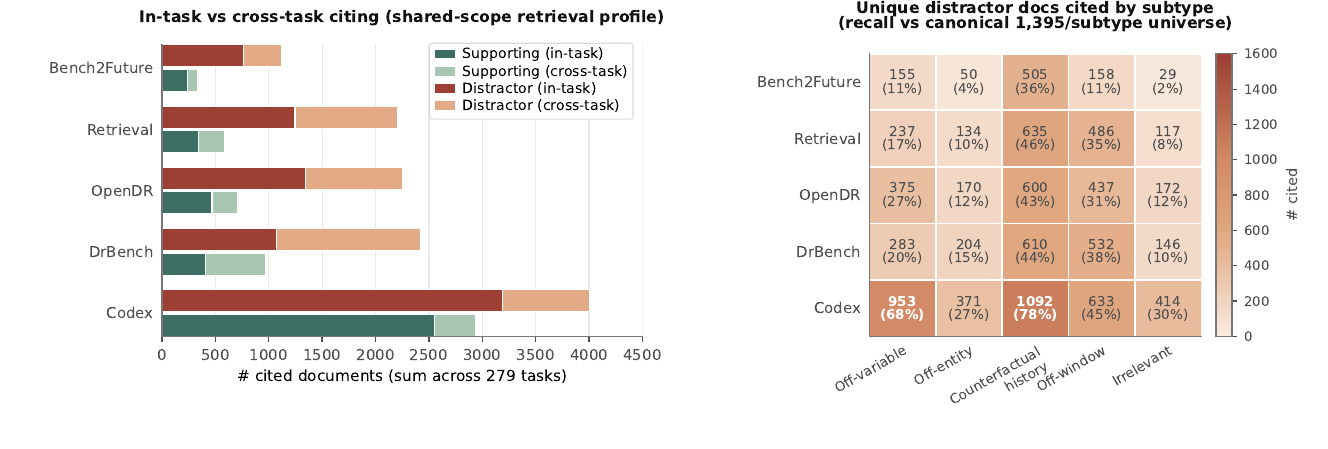}
\caption{Distractor citation patterns across DR agents. Left: cited documents decomposed by whether they are supporting or distracting, and whether they come from the target task or another task in the shared document corpus. Right: unique cited distractor documents by taxonomy, with percentages computed against the $1{,}395$ canonical documents per subtype (five per task across 279 tasks). The results show two main failure modes: agents often cite off-task distractors, and counterfactual-history distractors are consistently the hardest type to avoid.}
    \label{fig:distractor_appx}
\end{figure}

\subsection{In-Task vs Cross-Task Distractor Citation}
\label{subsec:dr-in-vs-cross}

For DrBench, we include provenance-resolved citations from two saved outputs that failed final citation validation; their two unresolved document IDs are excluded as hallucinated.
The left panel of Figure~\ref{fig:distractor_appx} shows that distractor failures arise not only from misleading documents within the target task, but also from off-task documents retrieved from the shared corpus. \textsc{DrBench}, \textsc{OpenDR}, and the retrieval baseline cite a large number of cross-task distractors, indicating that they rely on semantic similarity without sufficiently grounding documents to the correct entity, horizon, and forecasting target. \textsc{Codex} is more strongly in-task, which helps explain its higher evidence recall in Table~\ref{tab:agent_retrieval_metrics_transposed}, but it still cites many in-task distractors, showing that task grounding alone is insufficient. Agents must also reject documents that are temporally misaligned, causally irrelevant, or inconsistent with the observed series. This validates Dr-CiK's shared-corpus design: it separates coarse task-grounding failures from fine-grained distractor-rejection failures, making retrieval errors more diagnostic than in isolated per-task retrieval settings.

\subsection{Distractor Citation by Taxonomy}
\label{subsec:dr-subtypes}

The right panel of Figure~\ref{fig:distractor_appx} shows a clear hardness hierarchy across distractor types. \textit{Counterfactual-history} distractors are the dominant failure case for every agent, despite all subtypes having the same canonical universe size: they account for the largest share of citations across the board, and \textsc{Codex} cites $\rev{78\%}$ of the unique counterfactual-history distractor universe. In contrast, \textit{off-entity} and \textit{irrelevant} distractors are consistently avoided, with substantially lower recall for every agent, showing that they function as relatively easy negatives. This separation validates the design of our distractor taxonomy. Simple topical or entity-level distractors can be filtered by current agents, but forecast-dependent counterfactual-history distractors remain hard because they are both semantically plausible and numerically misleading: rejecting them requires checking whether the document's interpretation of the observed trajectory is consistent with the actual series and forecast horizon. This is precisely the capability that standard DR benchmarks do not test, and it explains why Dr-CiK exposes failures that would be hidden by generic hard-negative retrieval evaluation.

\section{DR Agent Failure-Mode Analysis}
\label{appx:dr_fail}

\begin{figure}[h]
\centering
\includegraphics[width=\linewidth]{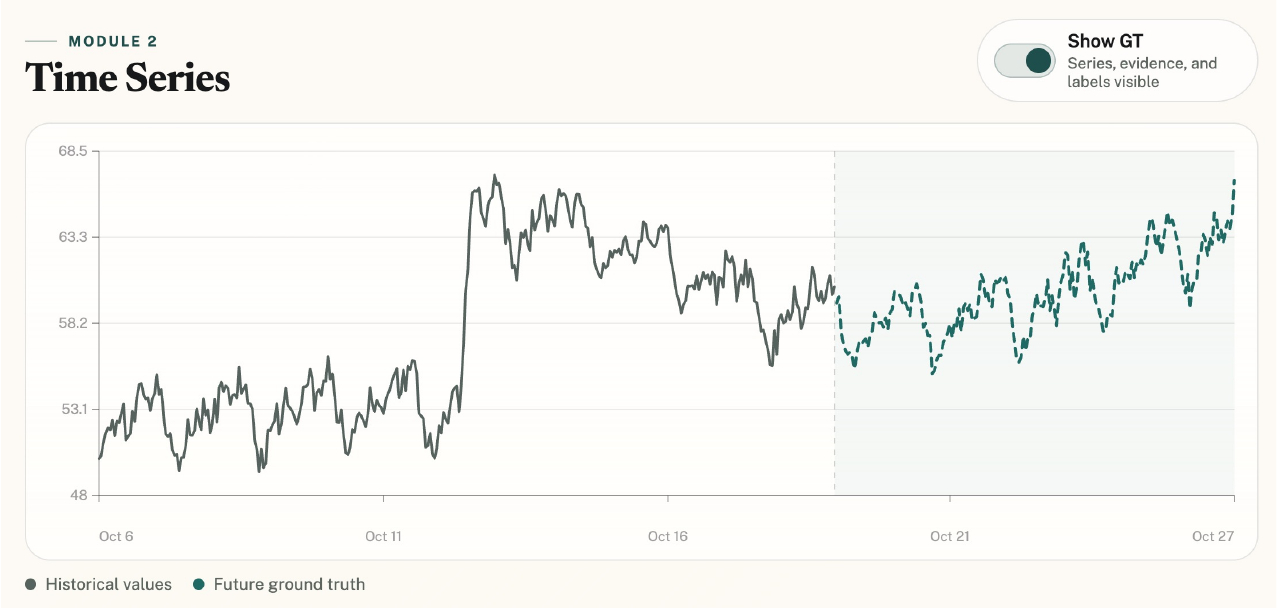}
\caption{Hourly electricity usage of Riverside Cottage 12 over the historical window 2025-10-06 to 2025-10-19.
The series exhibits two regimes.
From 2025-10-06 00:00 to 2025-10-13 00:00, the series sits at a low baseline because the unit was vacant and major appliances were deactivated; the renovation work itself occupies only the final 6 hours of this window, beginning at 2025-10-12 18:00.
From 2025-10-13 01:00 onward, the household is fully occupied and the baseline jumps to a permanently elevated level that persists for the remainder of the historical window.
The forecast horizon extends 180 hours past the right edge of the series.}
\label{fig:task216_timeseries}
\end{figure}

\paragraph{Setup.}
This task asks the agent to forecast the electricity usage of Riverside Cottage 12 over a 180-hour horizon (2025-10-19 20:00 to 2025-10-27 07:00). The historical window covers two weeks (2025-10-06 to 2025-10-19), during which the household underwent a brief renovation between the evening of October 12 and the early morning of October 13. The renovation forced the resident off-site, all major appliances were deactivated, and once construction concluded, the household became fully occupied with a permanently elevated baseline. The agent is given 14 supporting documents and 25 distractor documents, plus the time series itself. \textsc{Codex} reports 18 synthesized evidence items and cites 31 documents. None of the four non-trivial supporting evidence items (E2 through E5) is matched, so the supporting evidence recall on this task is zero.

\paragraph{Supporting evidence.}
The benchmark provides five items of supporting evidence for this task. E1 fixes the dataset metadata, E2 through E5 form the causal chain the agent must reconstruct.

\begin{tcolorbox}[title=Supporting evidence (Task 216), breakable]
\begin{minted}{markdown}
E1. This time series records the electricity usage (in Watt, W) of Riverside Cottage 12. The readings were recorded from 2025-10-06 00:00:00 to 2025-10-19 19:00:00, and the recording frequency is hourly (H).

E2. the owner started rennovation from 2025-10-12 18:00:00 to 2025-10-13 00:00:00 and did not stay at the unit

E3. leading to low power usage of the first 169 readings

E4. After the renovation finished on 2025-10-13 01:00:00, this household is fully occupied

E5. leading to a permanent increase in electricity usage
\end{minted}
\end{tcolorbox}

\subsection{Codex-Synthesized Evidence}
\label{appx:task216:synth}

For reference, the 18 synthesized evidence items Codex produced are listed below.

\begin{tcolorbox}[title=Codex-synthesized evidence (Task 216), breakable]
\begin{minted}{markdown}
1. Riverside Cottage 12 electricity usage is documented as Watt-based household power consumption, with specific focus on October 2025 load behavior and occupancy-driven baseline changes.

2. During the renovation and heavy-labor phase, the owner was not occupying the premises and major appliances and climate-control systems were deactivated, resulting in Wattage reductions due to a lack of human activity.

3. Construction concluded on October 13, at which point living quarters became available and the household reached full occupancy; this elevated post-handover baseline persisted through at least October 19.

4. Full occupancy is the primary driver of ongoing electricity use, linked to continuous environmental support, active residential services, and the simultaneous operation of essential systems that were formerly inactive.

5. Post-occupancy electricity use in Cottage 12 is rising, characterized by a drifting higher daily baseline and usage scaling through steady upward momentum.

6. A smart billing meter was installed on October 9, coinciding with a change in utility provider, followed by a safety inspection on October 15 after equipment updates.

7. A smart-meter calibration window from October 10 to October 15 caused temporary reporting-metric shifts and recorded spikes attributed to automated diagnostics rather than occupancy.

8. A localized storm and regional grid fluctuations between October 8 and October 14 caused abnormal high readings while surge protectors remained active.

9. Security and lighting installation from October 7 to October 12 introduced transient peak consumption due to high-frequency motion sensors and testing-related loads.

10. A strict energy quota system was implemented months before October 2025 to regulate peak demand and localized grid capacity, which may constrain extreme usage even under full occupancy.

11. Localized environmental monitoring from October 10 to October 14 recorded humidity variation, changing cloud cover, low nighttime light, and the presence of a reflective roof system designed to reduce heat gain.

12. The cottage features smart-home controls for temperature, lighting, security, media, and whole-home audio, which allows occupied behavior to affect both standby and active domestic electrical loads.

13. Logistics records for furniture and staging are inconsistent, with one document scheduling a curated furniture package for October 18 and another listing deliveries for late October.

14. There is conflicting evidence regarding usage trends: one report suggests usage is gradually declining after a post-renovation surge, while others describe sustained upward momentum after full occupancy.

15. Some documents describe a stable low-energy or flat profile with no October occupancy changes, contradicting reports of a post-October-13 elevated baseline following handover.

16. Evidence supports modeling a structural break in electricity usage from renovation/vacancy conditions to a fully occupied residential baseline following the October 13 handover.

17. Within the occupied regime, the directional signal indicates a higher and potentially rising baseline driven by residents' domestic routines and concurrent essential-system operation.

18. Short-lived drivers such as meter diagnostics, storms, security testing, and logistics introduce uncertainty into the operational records.
\end{minted}
\end{tcolorbox}

\subsection{Item-by-Item Failure Analysis}
\label{appx:task216:per_item}

\subsubsection*{Supporting evidence E2: renovation window}

Two supporting docs make the renovation window explicit. The project schedule memorandum states that ``the construction timeframe is set between 18:00 on October 12, 2025, and midnight the following day. Following the conclusion of these works on October 13, the residence will remain fully occupied with the new modernized baseline persisting continuously through the 19th of October and into the following week.'' The property-management notice to the resident is even more specific, scheduling the work ``between 18:00 on October 12, 2025, and midnight the following day \dots\ the resident is required to find alternative overnight accommodation for the duration of this window.'' The closest codex-synthesized evidence is item~2, which states that during the ``renovation and heavy-labor phase'' the owner was not occupying the premises and major appliances were deactivated. The qualitative content is correct, but the explicit calendar-and-clock interval (18:00 on October 12 to midnight) has been replaced by an aspectual phrase (``renovation and heavy-labor phase''). The judge's verdict was that the synthesis ``lacks the specific dates and times.'' The string \texttt{18:00} appears zero times in the entire \textsc{Codex} report, despite being printed verbatim in two supporting docs.

\subsubsection*{Supporting evidence E3: 169 low-usage readings}

The count is not a derived quantity\rev{;} it is stated outright in two supporting docs. The energy consumption analysis reports that ``the software analysis aggregated the initial 169 logged entries for the period ending October 13th. \dots\ the facility experienced a significant decline in its load profile.'' The hardware operations summary is even more emphatic, repeating the count three times within four sentences: ``\dots\ an isolated data set consisting of 169 logged entries. \dots\ the initial 169 logged entries associated with this hardware operations unit. \dots\ the resulting tally of these 169 specific readings reflected a prolonged low-use state of the unit throughout the monitored period.'' Codex's nearest synthesis is again item~2, which abstracts the entire phenomenon into ``Wattage reductions due to a lack of human activity.'' The specific count was discarded during synthesis. The judge concluded that ``none of the predicted insights mention the specific quantity of `169 readings' or the specific low power usage associated with that exact count.'' The token \texttt{169} appears zero times in the \textsc{Codex} report.

\subsubsection*{Supporting evidence E4: precise handover timestamp}

The on-site project completion report supplies this detail with the same phrasing the \textsc{Codex} evidence later borrows: ``Structural upgrades and interior remodeling were officially concluded at 1:00 AM on Monday, October 13, 2025. All construction activities have ceased, and all protective barriers have been removed from the premises. \dots\ The household is now fully occupied, and this residential status, along with the associated elevated power baseline, will persist continuously and indefinitely after October 13, 2025.'' Codex-synthesized evidence item~3 reads ``Construction concluded on October 13, at which point living quarters became available and the household reached full occupancy, this elevated post-handover baseline persisted through at least October 19.'' The lexical alignment with the source is close enough to confirm that the agent did read this document and used it to construct this evidence item. Yet the \texttt{1:00 AM} timestamp present in the source has been collapsed to a calendar-day grain (``October 13''). The judge's verdict was that the synthesis ``fails to include the specific timestamp (01:00:00).'' Neither \texttt{01:00} nor \texttt{1:00 AM} appears in the \textsc{Codex} report.

\subsubsection*{Supporting evidence E5: permanent versus rising}

The facility-engineering assessment labors the point with three near-synonyms in a single paragraph: ``Engineering observations confirm that the change in the residential population is enduring in nature. Consequently, the building is experiencing a permanent increase in total electricity usage due to this population change. These factors, stable occupancy and higher demand, have resulted in a lasting change to the overall power profile.'' \textsc{Codex} describes the same regime in two synthesized evidence items. Item~5 characterizes post-occupancy usage as ``rising, characterized by a drifting higher daily baseline and usage scaling through steady upward momentum,'' and item~17 says the ``directional signal indicates a higher and potentially rising baseline.'' Unlike E2 through E4 this is not a calendar-precision issue, it is a more subtle modal shift. The supporting evidence asserts a steady-state outcome (the baseline has stepped up and is expected to remain there). The synthesized evidence asserts a dynamic process (the baseline is still moving, with no commitment to whether it stabilizes or continues to drift). The two descriptions imply different forecast trajectories: a step function followed by a flat run versus a positive-slope continuous function. The judge's verdict was that the predicted insights ``do not characterize this change as `permanent'.'' The token \texttt{permanent} does not appear in the \textsc{Codex} report.

\subsection{Distractor Docs Were Not Avoided}
\label{appx:task216:distractors}

The supporting evidence is supported by 14 supporting docs and is challenged by 25 distractor docs. Several distractor docs offer alternative explanations for the structural break observed in the time series. The utility maintenance log attributes the shift to firmware: ``this variation is directly caused by a modification in the data logging logic integrated into the new firmware. This change represents a technical adjustment to reporting protocols rather than a shift in actual usage habits.'' The estate maintenance log introduces a billing-meter swap and a safety inspection in the relevant window: ``On 2025-10-09, a smart billing meter was officially installed at Riverside Cottage 12. \dots\ The local power company conducted a routine safety inspection at Riverside Cottage 12 on 2025-10-15 to ensure all systems meet current regulatory standards following the recent equipment updates.'' Several of these distractor-derived explanations nevertheless appear in the synthesized evidence.

\subsection{What the Failure Reveals}
\label{appx:task216:summary}

The four unmatched supporting evidence items share a common failure pattern, which is best described as \emph{specificity collapse during synthesis}.
In each case the agent retrieved the relevant supporting docs, identified the correct claim, and even preserved phrase-level lexical similarity in the synthesized evidence\rev{; what} it discarded was the quantitative anchor printed in the supporting doc.
The renovation interval 18:00 to midnight became ``renovation and heavy-labor phase.'' The 169-reading low-usage segment, repeated three times in the hardware operations summary, became ``Wattage reductions.'' The 1:00 AM handover moment, used verbatim in the completion report, became ``October 13.'' The adjective \emph{permanent}, used three times in the facility-engineering assessment, became \emph{rising}, \emph{drifting higher}, and \emph{steady upward momentum}.

The pattern is consistent with a generation-time preference for fluent analyst-style narrative over verbatim transcription of source specifics. Quoting the precise hour, the precise count, and the precise modal status of a baseline change reads as datasheet output rather than analytical synthesis, and the agent appears to default toward the latter style. For benchmarks and downstream tasks that consume the report as input to a forecasting model, this default is costly: the discarded specifics are precisely the anchors a forecaster needs to place a regime change at the right index, partition the training window correctly, and decide between a step-function extrapolation and a trend extrapolation. The specificity failure is therefore distinct from the distractor-avoidance failure described above: the relevant supporting documents were available to the agent, but their numeric anchors and modal qualifiers were lost during synthesis.

\section{Additional Context-Aided Forecasting Results}
\label{appx:additional_caf_results}

\begin{table*}[h]
\centering
\scriptsize
\setlength{\tabcolsep}{5pt}
\renewcommand{\arraystretch}{1.08}
\caption{
Context-aided forecasting results on the 225-task subset of Dr-CiK where every listed model produced valid metrics.
Values are mean $\pm$ sample standard deviation.
Rows are grouped by context family: \textsc{No Context} uses only the time series, \textsc{DeepResearch Evidence} uses Codex-synthesized evidence, and \textsc{Original Context} uses the source CAF context.
Scaled error metrics are winsorized at $5.0$ per task before aggregation.
Models marked with $^\ast$ are direct-prompt LLM forecasters.
\texttt{Gemini} denotes \texttt{gemini-3.1-flash-lite}, all \textit{MoiraiAgent} variants use \texttt{Gemini}, and ``+ Med.'' indicates medium reasoning effort.
}
\label{tab:caf_results}
\resizebox{\textwidth}{!}{\begin{tabular}{lccccc}
\toprule
\textsc{Model}
& \shortstack{\textsc{Avg. Rank}\\\textsc{Point}}
& \shortstack{\textsc{Avg. Rank}\\\textsc{Distribution}}
& \shortstack{\textsc{Scaled}\\\textsc{MAE}}
& \shortstack{\textsc{Scaled}\\\textsc{RMSE}}
& \shortstack{\textsc{Scaled}\\\textsc{CRPS}} \\
\midrule

\multicolumn{6}{l}{\textsc{No Context}} \\
\addlinespace[2pt]
\hspace{1.2em}\texttt{ARIMA} & 18.63 $\pm$ 6.67 & 18.52 $\pm$ 6.52 & 0.718 $\pm$ 1.162 & 0.867 $\pm$ 1.174 & 0.508 $\pm$ 0.882 \\
\hspace{1.2em}\texttt{SARIMA} & 16.69 $\pm$ 6.20 & 16.54 $\pm$ 6.21 & 0.597 $\pm$ 0.985 & 0.754 $\pm$ 1.026 & 0.422 $\pm$ 0.709 \\
\hspace{1.2em}\texttt{ETS} & 18.58 $\pm$ 8.35 & 18.44 $\pm$ 8.53 & 0.524 $\pm$ 0.532 & 0.704 $\pm$ 0.656 & 0.431 $\pm$ 0.431 \\
\hspace{1.2em}\texttt{Naive} & 20.84 $\pm$ 7.20 & 21.54 $\pm$ 6.94 & 0.793 $\pm$ 1.045 & 0.941 $\pm$ 1.075 & 0.515 $\pm$ 0.679 \\
\hspace{1.2em}\texttt{SES} & 17.65 $\pm$ 7.68 & 17.53 $\pm$ 7.91 & 0.520 $\pm$ 0.629 & 0.700 $\pm$ 0.723 & 0.388 $\pm$ 0.387 \\
\hspace{1.2em}\texttt{Chronos} & 13.49 $\pm$ 7.26 & 12.30 $\pm$ 7.38 & 0.428 $\pm$ 0.681 & 0.632 $\pm$ 0.839 & 0.327 $\pm$ 0.463 \\
\hspace{1.2em}\texttt{Aurora} & 16.13 $\pm$ 6.34 & 17.45 $\pm$ 6.74 & 0.577 $\pm$ 1.000 & 0.763 $\pm$ 1.092 & 0.503 $\pm$ 0.921 \\
\hspace{1.2em}\texttt{Gemini}$^\ast$ & 13.95 $\pm$ 6.81 & 14.79 $\pm$ 7.42 & 0.377 $\pm$ 0.564 & 0.582 $\pm$ 0.734 & 0.327 $\pm$ 0.545 \\
\hspace{1.2em}\texttt{MoiraiAgent} & 12.77 $\pm$ 6.72 & 16.33 $\pm$ 7.55 & 0.366 $\pm$ 0.540 & 0.560 $\pm$ 0.722 & 0.347 $\pm$ 0.516 \\
\hspace{1.2em}\texttt{TimesFM-3} & 12.40 $\pm$ 7.10 & 11.38 $\pm$ 6.60 & 0.382 $\pm$ 0.554 & 0.589 $\pm$ 0.779 & 0.357 $\pm$ 0.613 \\
\hspace{1.2em}\texttt{Toto-2.0-313M} & 11.00 $\pm$ 7.43 & 10.30 $\pm$ 6.88 & 0.322 $\pm$ 0.465 & 0.508 $\pm$ 0.645 & 0.308 $\pm$ 0.532 \\
\hspace{1.2em}\texttt{TiRex-2} & 12.31 $\pm$ 6.84 & 10.98 $\pm$ 5.81 & 0.357 $\pm$ 0.492 & 0.551 $\pm$ 0.677 & 0.296 $\pm$ 0.428 \\
\hspace{1.2em}\texttt{Moirai-2.0-R-small} & 14.20 $\pm$ 6.28 & 12.98 $\pm$ 5.93 & 0.386 $\pm$ 0.514 & 0.584 $\pm$ 0.724 & 0.312 $\pm$ 0.416 \\
\hspace{1.2em}\texttt{Moirai-MoE-Base} & 16.04 $\pm$ 6.22 & 13.84 $\pm$ 6.22 & 0.382 $\pm$ 0.455 & 0.598 $\pm$ 0.691 & 0.304 $\pm$ 0.360 \\
\midrule
\multicolumn{6}{l}{\textsc{DeepResearch Evidence}} \\
\addlinespace[2pt]
\hspace{1.2em}\texttt{Aurora} & 16.24 $\pm$ 6.27 & 17.50 $\pm$ 6.80 & 0.579 $\pm$ 1.002 & 0.763 $\pm$ 1.090 & 0.503 $\pm$ 0.918 \\
\hspace{1.2em}\texttt{Qwen3.5-4B}$^\ast$ & 15.08 $\pm$ 7.17 & 13.46 $\pm$ 6.89 & 0.453 $\pm$ 0.638 & 0.640 $\pm$ 0.834 & 0.311 $\pm$ 0.444 \\
\hspace{1.2em}\texttt{Qwen3.5-9B}$^\ast$ & 14.80 $\pm$ 7.63 & 12.79 $\pm$ 7.05 & 0.432 $\pm$ 0.604 & 0.608 $\pm$ 0.762 & 0.296 $\pm$ 0.412 \\
\hspace{1.2em}\texttt{Qwen3.5-27B}$^\ast$ & 14.02 $\pm$ 8.07 & 13.11 $\pm$ 8.09 & 0.415 $\pm$ 0.616 & 0.572 $\pm$ 0.778 & 0.324 $\pm$ 0.589 \\
\hspace{1.2em}\texttt{Mistral-Medium-3.1}$^\ast$ & 15.45 $\pm$ 8.31 & 16.00 $\pm$ 8.44 & 0.447 $\pm$ 0.640 & 0.616 $\pm$ 0.831 & 0.380 $\pm$ 0.622 \\
\hspace{1.2em}\texttt{Gemini}$^\ast$ & 13.14 $\pm$ 8.56 & 13.78 $\pm$ 8.82 & 0.408 $\pm$ 0.629 & 0.567 $\pm$ 0.796 & 0.331 $\pm$ 0.528 \\
\hspace{1.2em}\texttt{Gemini + Med.}$^\ast$ & 13.03 $\pm$ 8.16 & 13.33 $\pm$ 8.44 & 0.377 $\pm$ 0.554 & 0.529 $\pm$ 0.738 & 0.321 $\pm$ 0.532 \\
\hspace{1.2em}\texttt{MoiraiAgent} & 11.26 $\pm$ 7.28 & 12.64 $\pm$ 7.51 & 0.384 $\pm$ 0.652 & 0.528 $\pm$ 0.798 & 0.315 $\pm$ 0.548 \\
\hspace{1.2em}\texttt{Llama-3.2-3B}$^\ast$ & 21.33 $\pm$ 7.50 & 20.15 $\pm$ 7.79 & 1.370 $\pm$ 1.651 & 1.556 $\pm$ 1.676 & 0.637 $\pm$ 0.899 \\
\hspace{1.2em}\texttt{Phi-4-mini}$^\ast$ & 20.17 $\pm$ 7.37 & 19.34 $\pm$ 7.38 & 0.816 $\pm$ 1.239 & 1.031 $\pm$ 1.288 & 0.448 $\pm$ 0.586 \\
\midrule
\multicolumn{6}{l}{\textsc{Original Context}} \\
\addlinespace[2pt]
\hspace{1.2em}\texttt{Aurora} & 16.38 $\pm$ 6.29 & 17.48 $\pm$ 6.49 & 0.580 $\pm$ 1.007 & 0.762 $\pm$ 1.093 & 0.507 $\pm$ 0.932 \\
\hspace{1.2em}\texttt{Gemini}$^\ast$ & 8.26 $\pm$ 7.40 & 8.81 $\pm$ 7.87 & 0.274 $\pm$ 0.519 & 0.393 $\pm$ 0.680 & 0.240 $\pm$ 0.506 \\
\hspace{1.2em}\texttt{MoiraiAgent} & 6.00 $\pm$ 5.31 & 7.25 $\pm$ 6.18 & 0.248 $\pm$ 0.498 & 0.350 $\pm$ 0.663 & 0.212 $\pm$ 0.472 \\
\hspace{1.2em}\texttt{MoiraiAgent + Med.} & 6.16 $\pm$ 5.78 & 7.42 $\pm$ 6.27 & 0.252 $\pm$ 0.497 & 0.355 $\pm$ 0.664 & 0.216 $\pm$ 0.467 \\
\bottomrule
\end{tabular}}
\end{table*}

Table~\ref{tab:caf_results} compares forecasters on the 225-task intersection on which every listed model produced a valid metric, so per-model task drift cannot affect the comparison.
The table groups forecasters into three context families: \textsc{No Context}, where the forecaster receives only the time series, \textsc{DeepResearch}, where the forecaster receives Codex-synthesized evidence, and \textsc{Original Context}, where the forecaster receives the source CAF context.
The same qualitative pattern holds throughout: high-quality supporting context improves DP-Gemini and MoiraiAgent substantially over no context, whereas Codex-synthesized evidence closes only part of this gap. Aurora remains largely context-insensitive, suggesting that its forecasts are dominated by the time-series modality rather than the textual context.

\subsection{Deep Research and Forecasting Quality by Input Source}
\label{app:by-source-results}

Table~\ref{tab:e2e-by-source} breaks down end-to-end forecasting by input source.
For the two context-sensitive forecasters, DP-Gemini and MoiraiAgent,
expert-curated tasks yield slightly higher no-context forecasting error than
CAF-sourced tasks and remain more challenging when the original context is provided.
Original context improves forecasting on both sources, but does not eliminate this gap.
Aurora shows a different cross-source pattern and remains largely insensitive to context.

\begin{table}[h]
\centering
\small
\caption{End-to-end forecasting by input source (sCRPS $\downarrow$).}
\label{tab:e2e-by-source}
\resizebox{\linewidth}{!}{
\begin{tabular}{lcccccc}
\toprule
\textbf{Context} & \makecell{\textbf{Aurora:}\\\textbf{CAF-Sourced}} & \makecell{\textbf{Aurora:}\\\textbf{Expert}} & \makecell{\textbf{DP-Gemini:}\\\textbf{CAF-Sourced}} & \makecell{\textbf{DP-Gemini:}\\\textbf{Expert}} & \makecell{\textbf{MoiraiAgent:}\\\textbf{CAF-Sourced}} & \makecell{\textbf{MoiraiAgent:}\\\textbf{Expert}} \\
\midrule
\textsc{No Context} & 0.537 & 0.341 & 0.339 & 0.344 & 0.356 & 0.362 \\
\textsc{Original Context} & 0.541 & 0.344 & 0.243 & 0.282 & 0.220 & 0.252 \\
\textsc{Bench2Future} & 0.534 & 0.346 & 0.691 & 0.534 & 0.565 & 0.466 \\
\textsc{DrBench} & 0.535 & 0.345 & 0.593 & 0.616 & 0.530 & 0.471 \\
\textsc{Codex} & 0.536 & 0.346 & 0.340 & 0.325 & 0.325 & 0.314 \\
\bottomrule
\end{tabular}
}
\end{table}

Table~\ref{tab:er-by-source} breaks down DR Evidence Recall by input source: expert-curated tasks are also more challenging to research for every agent (\textsc{Codex} 38.5\% on CAF-sourced vs.\ 21.0\% on expert-curated).

\begin{table}[h]
\centering
\small
\caption{Deep-research Evidence Recall by input source.}
\label{tab:er-by-source}
\begin{tabular}{lcc}
\toprule
\textbf{DR method} & \textbf{CAF-Sourced ER $\uparrow$} & \textbf{Expert ER $\uparrow$} \\
\midrule
\textsc{Bench2Future} & 4.00\% & 2.83\% \\
\textsc{DrBench} & 3.80\% & 3.20\% \\
\textsc{Retrieval} & 4.72\% & 3.37\% \\
\textsc{OpenDR} & 4.69\% & 3.98\% \\
\textsc{Codex} & 38.54\% & 21.04\% \\
\bottomrule
\end{tabular}
\end{table}

Table~\ref{tab:difficulty-stratified} reports forecasting error and DR Evidence Recall stratified by difficulty level (annotated from each task's causal context) across the full 279-task pool. DR Evidence Recall declines monotonically with difficulty on \rev{most dimensions}, suggesting that these labels track real challenges; forecast error does not show the same trend on every dimension, consistent with not all evidence being retrieved and the forecasting module not always being able to apply what is retrieved.

\begin{table}[h]
\centering
\small
\caption{Difficulty-stratified results across the full 279-task pool (end-to-end sCRPS uses Codex+MoiraiAgent; DR Evidence Recall uses Codex).}
\label{tab:difficulty-stratified}
\begin{tabular}{llrrr}
\toprule
\textbf{Dimension} & \textbf{Level} & \textbf{Tasks} & \textbf{End-to-end sCRPS $\downarrow$} & \textbf{DR Evidence Recall $\uparrow$} \\
\midrule
Certainty & Certain & 66 & 0.318 & 36.9\% \\
 & Likely & 116 & 0.375 & 27.7\% \\
 & Uncertain & 97 & 0.257 & 25.8\% \\
Explicitness & Explicit & 218 & 0.314 & 31.2\% \\
 & Implied & 53 & 0.360 & 20.6\% \\
 & Implicit & 8 & 0.301 & 25.6\% \\
Domain knowledge & General & 150 & 0.308 & 39.6\% \\
 & Specialist & 129 & 0.340 & 22.9\% \\
Temporal complexity & Straightforward & 138 & 0.297 & 39.8\% \\
 & Predictable & 56 & 0.435 & 29.4\% \\
 & Variable & 85 & 0.283 & 21.5\% \\
\bottomrule
\end{tabular}
\end{table}

\subsection{Evidence Role: Retrospective vs.\ Prospective}
\label{app:evidence_roles}

Expert-curated supporting evidence spans two roles: \emph{retrospective} evidence explains patterns already visible in the historical (pre-forecast) window, while \emph{prospective} evidence describes the future window as if it had not yet been observed. Both roles occur naturally in context-aided forecasting, and the source context underlying CiK and GIFT-CTX already blends them; our pipeline surfaces this structure explicitly wherever the input source's evidence is authored as an explicit multi-step causal chain, which is currently the case for the 80 expert-curated tasks. This role tag is recorded per supporting-evidence item, so DR retrieval and forecasting can be evaluated separately for each role.

Table~\ref{tab:er-by-role} shows that DR methods struggle to retrieve both roles, even though retrospective evidence should be easier to identify since the historical window is already given to the agent.

\begin{table}[h]
\centering
\small
\caption{DR Evidence Recall by evidence role (80-task expert split).}
\label{tab:er-by-role}
\begin{tabular}{lccc}
\toprule
\textbf{DR method} & \textbf{Retrospective ER $\uparrow$} & \textbf{Prospective ER $\uparrow$} & \textbf{Combined ER $\uparrow$} \\
\midrule
\textsc{Bench2Future} & 2.16\% & 3.52\% & 2.83\% \\
\textsc{DrBench} & 3.94\% & 2.43\% & 3.20\% \\
\textsc{Retrieval} & 4.26\% & 2.43\% & 3.37\% \\
\textsc{OpenDR} & 3.43\% & 4.55\% & 3.98\% \\
\textsc{Codex} & 25.98\% & 15.85\% & 21.04\% \\
\bottomrule
\end{tabular}
\end{table}

Table~\ref{tab:forecast-by-role} isolates the forecasting-side contribution of each role by supplying the forecaster with only the retrospective evidence or only the \rev{prospective} evidence for each task (in addition to the numerical history). Prospective evidence carries the dominant signal a forecaster can act on, but the gap to the full original-context ceiling closes only when retrospective evidence is included as well. The \textsc{Retrospective only} condition also approximates a more deployment-realistic setting in which no information about the future is available in advance.

\begin{table}[h]
\centering
\small
\caption{Forecasting accuracy by evidence role (80-task expert split, sCRPS $\downarrow$).}
\label{tab:forecast-by-role}
\begin{tabular}{lccc}
\toprule
\textbf{Context} & \textbf{Aurora} & \textbf{DP-Gemini} & \textbf{MoiraiAgent} \\
\midrule
No context        & 0.341 & 0.344 & 0.362 \\
Retrospective only  & 0.343 & 0.322 & 0.318 \\
Prospective only    & 0.344 & 0.293 & 0.260 \\
Original context  & 0.344 & 0.282 & 0.252 \\
\bottomrule
\end{tabular}
\end{table}

\section{Prompts}
\label{appendix:prompts}

This appendix reproduces the prompts used by every LLM-driven step of
Dr-CiK environment generation, the LLM judges, and the
task-difficulty probes. Placeholders in curly braces (e.g.\
\texttt{\{context\}}, \texttt{\{task\_id\}}) denote runtime fields that are filled in for each invocation.

\subsection{Generation Prompts}
\label{app:prompts-generation}

\subsubsection{Entity Pipeline}

\begin{tcolorbox}[title=Entity Extraction (Stage 1), breakable]
\begin{minted}{markdown}
Here is a collection of background text from multiple forecasting tasks:

<background_corpus>
{background_corpus}
</background_corpus>

---

Can you help me extract the entities, the associated time-series variables, and the 'contain' relations between them (only between extracted entities)? 
For each task, there must be exactly one primary entity (the object directly associated with the time series variable).

Please return JSON with this schema:
{
  "entities": [
    {
      "entity_name": "...",
      "time_series_variable": ["..."],
      "source_tasks": ["..."],
      "source_tasks_as_primary_entity": ["..."]
    }
  ],
  "contain_relations": [
    {
      "parent_entity": "...",
      "child_entity": "...",
      "evidence": "..."
    }
  ]
}

Requirements:
1. Entity list must be NON-REPEATED (deduplicated by real-world entity). The order in time_series_variable should be the same as the order in source_tasks_as_primary_entity for the same entity.
2. The source_tasks_as_primary_entity should be the source_tasks that entities are directly associated with the time series variable.
3. Entity names must refer to GENERAL real-world categories (type-level concepts). If an entity contains descriptive modifiers (e.g., location, ownership, time, temporary context), REMOVE these modifiers and keep only the base category name, and modifiers should not be extracted as entities.
4. Additional entities should only be extracted if they serve as parent entities that contain the primary entity in a clear physical, geographical, or organizational sense. No additional entities should be extracted otherwise.
5. Only include containment relations that are explicitly stated or universally true by definition. Do not include contain_relations that require assumptions, background knowledge expansion, or multi-hop inference.
6. The parent_entity and the child_entity must be from the entity list. If no contain relation is clear, return an empty list for "contain_relations".
7. Do NOT output transitive/ancestor containment edges. Only output the nearest parent (most specific parent) for each child entity when multiple levels exist.
8. Only return a JSON object with the specified schema, and make sure it satisfies all the requirements above. Do not include any additional text in the response.
\end{minted}
\end{tcolorbox}
\vspace{5pt}

\begin{tcolorbox}[title=Entity Profile Generation (Stage 2), breakable]
\begin{minted}{markdown}
Now generate realistic but purely synthetic profiles for one target entity.

Target entity:
{entity_name}

Time-series variables for tasks where this target entity is the primary entity are:
{all_time_series_variables}

Please assign {k_profiles} different profiles to this target entity.

Please return JSON with this schema:
{
  "entity_name": "...",
  "profiles": [
    {
      "profile_id": "1",
      "name": "...",
      "details": {
        "...": "...",
      }
    }
  ]
}

Requirements:
1. The profiles must be realistic and detailed enough to sound like plausible real-world descriptions for the target entity.
2. The profiles must be purely synthetic. Do not copy or infer any factual background from the input tasks beyond the entity category itself.
3. The profiles must not contain any information that could affect forecasting for any of the listed time-series variables, either directly or indirectly.
4. Avoid predictive signals such as events, trends, operations, demand, weather, outages, anomalies, promotions, policies, schedules, maintenance, failures, holidays, traffic shifts, or any other factor that may influence the time series.
5. Keep the profiles as stable background descriptions only, the details should be realistic and just background information will be enough.
6. Generate exactly {k_profiles} profiles for the target entity.
7. The returned entity_name must match the target entity exactly.
8. Make sure the name and details are REALISTIC as if they were describing a real-world entity, but do not include any information that could be used to predict the time-series variables associated with the entity.
9. Only return a JSON object with the specified schema. Do not include any additional text in the response.
\end{minted}
\end{tcolorbox}
\vspace{5pt}

\begin{tcolorbox}[title=Profile Assignment / Rewritten Background (Stage 3), breakable]
\begin{minted}{markdown}
Original context: ```{original_context}```

Source task ID: ```{source_task_id}```

Primary entity: ```{primary_entity_name}```

Assigned primary entity profile: ```{primary_entity_profile}```

Ancestor entity chain from nearest parent to farthest ancestor: ```{ancestor_chain_profiles}```

---

You will rewrite one original context into a sample-level profiled version.

Rewrite instructions:

1. Replace the original entity mention with the assigned profile in a way that matches the original information density.
2. If the original context only mentions the entity as a name/category, only replace it with the generated profile name.
3. If the original context gives slightly richer but still stable identity/background information, you may use a small amount of profile detail, but do not increase the information density beyond the original context.
4. If ancestor entities are present in the original context, replace each mentioned ancestor using the assigned ancestor profiles as needed, again matching the original information density.
5. You can make the slight modifications needed to ensure the rewritten context is fluent and coherent, but do not change any of the original information beyond the entity profile replacement.

Please return JSON with this schema:
{
  "source_task_id": "...",
  "rewritten_context": "...",
  "background_rewritten_context": "...",
  "applied_profiles": {
    "primary_entity": {
      "entity_name": "...",
      "profile_id": "..."
    },
    "ancestor_entities": [
      {
        "entity_name": "...",
        "profile_id": "..."
      }
    ]
  }
}

Requirements:
1. The returned source_task must match the given source task ID exactly.
2. If there are no ancestor entities, return an empty list for applied_profiles.ancestor_entities.
3. Just replace the entity identity wording with the assigned profiles. Do not change any of remaining original context.
4. After rewriting, the reader should still be able to clearly identify the original primary entity type ({primary_entity_name}) from the rewritten context. If the entity type cannot be easily inferred from the context, you may explicitly mention the entity type at the relevant position(s) in the text.
5. Just replace the entity identity wording with the assigned profiles. Do not change any of remaining original context.
6. The `background_rewritten_context` should contain only the background information of the primary entity from the rewritten context, you shoudl mention the entity profile in it. No events, scenarios, or other information that is not background information of the primary entity.
7. The rewritten content must be fully synthetic and must not contain any real-world identifiable information (names, institutions, locations, events, etc.).
8. Only return a JSON object with the specified schema. Do not include any additional text in the response.
\end{minted}
\end{tcolorbox}

\subsubsection{Ground-Truth Evidence and Reasoning Chain}

\begin{tcolorbox}[title=Ground-Truth Evidence Extraction, breakable]
\begin{minted}{markdown}
```{context}```

Extract the independent event(s) and the causal graph (each causal chain) from the text above, including any associated `time_start` and `time_end` (formatted as `YYYY-MM-DD HH:MM:SS`, if available). If a timestamp is not available, set its value to "null".

Requirements:
* Do not add any facts or context not explicitly stated in the text.
* When a causal relationship exists, represent the cause and the effect as two semantically complete events. Do not split discourse markers or partial clauses into separate events.
* Combine all background information, constraints, or other information with no causal relationship into a single event ("E1"). 
* Each event must contain the complete relevant original text span.
* Ensure that the extracted events collectively cover all the original information (no information is missing), and that no information is duplicated or overlapping.
* The provided content may include task instructions, often appearing in the last sentence. If present, ignore these instructions.
* If multiple statements describe successive states, value changes, or different aspects of the same underlying entity, process, or analogy, combine them into a single event (but not the background event) rather than splitting each state or timestamped change into separate events.
* An event should represent a complete semantic unit describing a state, action, or causal situation. Avoid splitting short phrases, discourse markers, or connective fragments (e.g., "Due to", "After", "resulting in") into separate events.
* The causal graph and causal context may be empty if the text contains no causal relationships. In this case, the objective is simply to decompose the original content into events and their corresponding text spans.
* Each element in the `causal_context` list must correspond one-to-one with an entry in the `causal_graph` list, and both lists must have the same length.
* The `causal_context` should only contain the original context covered by the events in the corresponding causal graph entry, do not add any other information, so different causal contexts should not be the same. But each causal context should be self-contained, expressing a complete meaning.
* If an event's `time_start` or `time_end` can be logically inferred from other events in the text, fill them in accordingly; otherwise set them to "null".
* If there is a gap between time intervals, do not merge time ranges across non-contiguous periods, represent them as separate events rather than combining them into a single continuous range. Make sure each event has a self-contained context.
* Do not treat system descriptions, design properties, or static characteristics (e.g., "is designed to", "scales linearly") as effects in a causal relationship unless they are explicitly caused by another event.
* Put only stable background or non-causal contextual information into E1. If the context states constraints on the time-series variable (e.g., bounds, maxima, minima, capacities, or other operating limits), extract those constraints as a separate self-contained event rather than merging them into E1.
* If an extracted event contains explicit timestamps or clearly timestamped transitions, do not leave its time fields as "null"; use the earliest and latest explicit timestamps covered by that event span, and split the event if merging would make the timing unclear.

Output format:
Return a valid JSON object with the following structure (and no additional text):

{
  "extracted_events": [
    {
      "event_id": "E1",
      "event": "complete relevant original text span, self-contained and semantically complete",
      "time_start": "YYYY-MM-DD HH:MM:SS or null",
      "time_end": "YYYY-MM-DD HH:MM:SS or null"
    },
    ...
  ],
  "causal_graph": [
    {
      "cause_event_id": "E1",
      "effect_event_id": "E2",
      "relation": "cause"
    },
    ...
  ],
  "causal_context": [
    "Natural language description of the causal relation using the original text, self contained, without omitting any information",
    ...
  ]
}
\end{minted}
\end{tcolorbox}
\vspace{5pt}

\begin{tcolorbox}[title=Reasoning Chain Expansion, breakable]
\begin{minted}{markdown}
You are a reasoning-chain refactoring assistant.

Your task is to refactor the given input into a full reasoning chain by introducing exactly {hops} new contexts.

Item type:
- {item_kind}
- source ids: {source_event_ids}

Goal:
Construct a coherent, step-by-step reasoning chain that preserves the original meaning while preventing shortcut retrieval.
Also identify the materially real source-grounded details in the input and track where each one is preserved in the expanded chain.

The final chain must:
- be logically valid and sequential,
- require step-by-step reasoning to traverse,
- avoid shortcut reasoning and retrieval,
- be locally obvious but globally non-obvious.

Core requirements:
1. Introduce exactly {hops} new contexts.
2. Preserve the original meaning of the input.
3. The output must be the full final chain (not only new contexts).
4. Original input contexts may be preserved, rewritten, split, or repositioned, but semantics must remain unchanged; if the input uses descriptive wording to characterize a specific state or transition, preserve that descriptive information for that same state or transition, and do not merge multiple descriptively distinct changes into one broader statement.
5. Each context must be specific, standalone, and reasoning-relevant; do not use generic bridge steps, and ensure every intermediate context preserves or advances the original causal specificity rather than replacing it with a vague statement.
6. Adjacent contexts must have a clear causal (reasoning) connection.
7. The chain must be a single linear path (no cycles, no branching).

Shortcut prevention rules:
8. No single context or adjacent pair of contexts should be sufficient to reconstruct the final outcome.
9. Avoid placing critical or distinctive information at the beginning of the chain.
10. Distribute key information across NON-ADJACENT contexts.
11. Avoid repeating or clustering distinctive details in nearby contexts.
12. Ensure information can only be accessed through sequential traversal.

Identifier and anchor control:
13. Treat identifiers (timestamps, names, unique values, distinctive phrases) as controlled anchors.
14. Anchors must not be repeated, paraphrased, or redistributed in ways that enable shortcut matching.
15. Anchors must not allow any single edge (adjacent pair) to act as a shortcut.
16. Structured fields from input (e.g., time_start, time_end) are boundary constraints and must not be freely propagated.

Detail distribution and rewriting:
17. Separate identifiers from descriptive details when possible.
18. Spread different details across NON-ADJACENT contexts.
    When direct separation would make the chain unnatural or incoherent, introduce intermediate context(s) to preserve realism while ensuring that such details remain SEPARATED by at least one reasoning step.
19. Rewrite original content with strong expression divergence:
    - All essential information must be preserved,
    - but the expression should be rewritten to be as different as possible in wording, phrasing, format (e.g., timestamps), and structure,
    - minimize lexical overlap and avoid distinctive or recognizable patterns,
    - the rewritten context should not be easily matched to the original through surface-level similarity,
    - prioritize maximal expression divergence while maintaining full semantic equivalence.
20. Do not produce contexts that closely mirror the original input.

Quality constraints:
21. Maintain realism and plausibility, and preserve temporal and logical consistency with the input, including the original ordering, duration, persistence, reversals, and phase transitions of events or states.
22. Preserve all essential information across the full chain.
23. Avoid trivial steps, redundancy, or artificial decomposition.

Interpretation rule:
- Use relation = "cause" for all edges (as general reasoning dependency).

Item-specific rules:
{item_specific_rules}

Output format:
Return either:
1. JSON with:
   - "source_detail_inventory"
   - "contexts"
   - "relation_graph"
   - "detail_coverage"
2. null

Alignment:
- `relation_graph` must contain exactly `len(contexts) - 1` edges.
- Each edge in `relation_graph` must include exactly one non-empty `relational_context`, so there must be exactly `len(contexts) - 1` relational contexts in total.
- Each edge in `relation_graph` must also include `contains_source_grounded_details` and `grounded_detail_ids`.
- Use `grounded_detail_ids` on an edge only when that edge preserves a real source-grounded relation/detail rather than bridge-only reasoning.
- `detail_coverage[*].covered_by_edge_ids` must use the exact edge ids from `relation_graph`, formatted as `R1->R2`, `R2->R3`, ... . Do not use any other separator.

Context rules:
- Use IDs R1, R2, ..., in order.
- `relation_graph` must use only these `context_id` values.
- Do not place original source ids directly inside `relation_graph`.
- The final chain must include both:
  - contexts newly introduced by the model, and
  - contexts derived from the original input.
- Each original source id must be represented by at least one context in the final chain.
- Each context must include:
  - context_id
  - context
  - origin: new | original_preserved | original_refactored
  - type_count: string in the form `<origin>_<n>`, where `n` is the 1-based occurrence index within that origin type as contexts appear in `contexts`
  - contains_source_grounded_details: true | false
  - grounded_detail_ids: list of `detail_id` values preserved in this context
- If origin is original_*, include source_context_id.
- If origin is new, do not include source_context_id.
- Count `new`, `original_preserved`, and `original_refactored` separately when assigning `type_count`. For example, if the first two `new` contexts appear at `R2` and `R5`, their `type_count` values must be `new_1` and `new_2`.
- Exactly {hops} contexts must have origin = "new".
- For identifiers originating from the input (e.g., timestamps, unique profiles, or distinctive references), their associated semantic content and the identifier itself should be separated across different, non-adjacent contexts whenever possible. 
- The identifier and its corresponding descriptive information must not be co-located in a single context or adjacent contexts in a way that enables direct retrieval.
- However, the chain must still allow the original information to be reconstructed through sequential reasoning across multiple steps.
- Add `source_detail_inventory` before `contexts`.
- Add `detail_coverage` after `relation_graph`.
- Each context and each edge must indicate whether it contains any real source-grounded detail, and if so, which `detail_id` values it preserves.
- Every materially real source-grounded detail from the input must appear in `source_detail_inventory` and must be covered at least once in `detail_coverage`.
- Do not label inferred bridge reasoning as if it were a real source-grounded detail from the input.
- To maximize reasoning difficulty, distribute source-grounded details as sparsely as possible across different, non-adjacent contexts or edges instead of clustering multiple details in one context whenever fidelity and coherence allow.
- If a source-grounded detail is most naturally preserved in the transition between two contexts rather than in a single context, it may appear only on the corresponding `relation_graph` edge; avoid duplicating the same detail in a context unless necessary for faithfulness or coherence.

Fallback:
Return null if no meaningful, consistent reasoning chain can be constructed.

Output JSON schema:
{
  "source_detail_inventory": [
    {
      "detail_id": "D1",
      "detail": "source-grounded detail text"
    }
  ],
  "contexts": [
    {
      "context_id": "R1",
      "context": "...",
      "origin": "new | original_preserved | original_refactored",
      "type_count": "original_refactored_1",
      "source_context_id": "E1",
      "contains_source_grounded_details": true,
      "grounded_detail_ids": ["D1"]
    }
  ],
  "relation_graph": [
    {
      "source_context_id": "R1",
      "target_context_id": "R2",
      "relational_context": "...",
      "relation": "cause",
      "contains_source_grounded_details": false,
      "grounded_detail_ids": []
    }
  ],
  "detail_coverage": [
    {
      "detail_id": "D1",
      "covered_by_context_ids": ["R1"],
      "covered_by_edge_ids": []
    }
  ]
}

Input:
{context}
\end{minted}
\end{tcolorbox}

\subsubsection{Supporting Document Generation}

\begin{tcolorbox}[title=Source Metadata Reconstruction, breakable]
\begin{minted}{markdown}
{context}

You are given one causal evidence bundle from a forecasting task.

Your job is to reconstruct exactly one plausible standalone real-world source that could directly contain this information.

Requirements:

- Produce exactly one source metadata object.
- Use only information directly supported by the provided bundle.
- Do NOT add inferred background, broader causes, forecasts, or unstated implications.
- The source must be realistic, self-contained, and independently plausible.
- Prefer a source that naturally contains all of the information in this bundle with minimal extra scope.
- `provides` should contain self-contained atomic facts.
- `source_content_length` should be a realistic approximate target length for the final reconstructed source content.
For calibration, typical ranges include: short alerts or log entries (~80-150 words), emails or internal chat summaries (~40-120 words), internal memos or brief reports (~300-500 words), news articles or public bulletins (~500-800 words), technical or operational reports (~550-600 words), policy documents or official statements (~400-600 words), and longer formal reports, audits, or multi-section documents (~700+ words); choose an appropriate range based on the source type and issuer rather than minimizing length.

Return valid JSON only with exactly this schema:

{
  "name": "...",
  "type": "...",
  "issuer": "...",
  "provides": ["..."],
  "notes": "Scope of this source and what it does NOT include",
  "source_content_length": "? words"
}
\end{minted}
\end{tcolorbox}
\vspace{5pt}

\begin{tcolorbox}[title=Metadata-to-Document Content, breakable]
\begin{minted}{markdown}
{source_metadata}

You are a generator that reconstructs realistic source content from structured metadata.

Your task is to generate the original content of this source as it would appear in the real world.

---

Requirements:

- The content must match the real-world form of the source type (e.g., report, log, memo, webpage, dataset description)
- Choose the most appropriate format automatically (e.g., plain text, structured log, table, markdown, or HTML if applicable)
- Use realistic tone, structure, and formatting conventions
- All information in `provides` must be explicitly included
- Do NOT introduce new facts, assumptions, or inferred details
- If a "Reference background for consistency only" block is provided, use it only to keep entity/background details consistent with the metadata, and do NOT write any content that is not already supported by the structured metadata above
- If `source_content_length` is provided, use it as the target length guidance for the reconstructed source content
- The length of the content should be appropriate and realistic for the given source type and any provided `source_content_length`
- You MAY add minimal connective language for clarity and coherence
- The result must be self-contained and read like an authentic standalone artifact

---

Constraints:

- Do not mention the metadata or reconstruction process
- Do not add explanations outside the content

---

Output Format (JSON ONLY)

{
  "content": "full reconstructed source content"
}
\end{minted}
\end{tcolorbox}

\subsubsection{Distractor Document Generation}

\begin{tcolorbox}[title=Distractor Content Generation (Off-variable / Irrelevant / Off-entity / Off-window), breakable]
\begin{minted}{markdown}
You are a distractor generation assistant.

Your task is to generate realistic but misleading or non-useful contexts (distractors) based on the given input context.

Input:
```{context}```
{supplemental_context_block}

Goal:
Generate {num_distractors} distractor contexts of type `{distractor_type}` that:
- appear highly similar and plausible,
- share strong surface and structural similarity with the input,
- but do NOT support the correct reasoning path or final answer.

General requirements:
1. Distractors must be realistic and natural.
2. They must not be obviously wrong or nonsensical.
3. They must not allow recovery of the correct answer.
4. Distractors must not reuse or preserve any key contextual details from the input that are directly useful for reasoning or prediction.
   - This includes events, time periods, details, causal relationships, causal_contexts, critical conditions, and outcome-related signals.
   - Distractors must not share the same underlying cause, mechanism, or predictive factors as the input.

5. Task relevance constraint (CRITICAL):
   - Distractors must be irrelevant or insufficient for solving the target task.
   - For forecasting tasks, distractors must not provide valid predictive signals about the future behavior or outcome of the target entity.
   - They may appear related, but should either:
     - lack key predictive information, or
     - contain misleading or non-applicable signals.

6. A human reader should be able to determine that the distractor is not useful or is insufficient.
7. The distractor must be fully self-contained:
   - it must read as a standalone document fragment,
   - it must not refer back to the original context, original event, or original time series,
   - and it must not use phrases like "the same", "as before", "the original event", or similar cross-references.

High-confusability requirement (IMPORTANT):
8. Distractors should preserve as much surface similarity as possible:
   - keep the same entities (e.g., names, locations, variables),
   - reuse similar wording and phrasing where appropriate (e.g., timestamps, structure, style),
   - maintain similar sentence patterns or structure.
   - Use plain text, no formatting, the content should be directly usable as a document body without any post-processing. Don't follow the format of the Input.

9. However, despite this similarity, the distractor must NOT become valid evidence.

Distractor type for this call:
- {distractor_type}: {distractor_type_description}

Type-specific requirements:
{distractor_type_requirements}

Diversity requirement:
- Avoid generating distractors that are too similar to each other.
- Vary the phrasing, emphasis, and document focus while keeping the same distractor type.

Critical constraint:
- No distractor should independently or jointly form a correct reasoning path.
- Every distractor in this call must use `distractor_type` = `{distractor_type}` exactly.

Forecasting-specific guidance:
- If the task involves prediction:
  - Distractors may include plausible but incorrect forecasts,
  - or correct patterns under different conditions,
  - but must not match the true scenario.

Output format:
Return a JSON object. "provides" should be a list of strings describing what to contain in the distractor content, no formatting, just plain text.

{
  "distractors": [
    {
      "provides": ["...", ...],
      "distractor_type": "{distractor_type}",
      "reason": "short explanation of why this is a distractor and how it differs from the original context"
    }
  ]
}
\end{minted}
\end{tcolorbox}
\vspace{5pt}

\begin{tcolorbox}[title=Counterfactual-History Distractor Specification, breakable]
\begin{minted}{markdown}
You are a distractor generation assistant.

Your task is to generate one realistic but misleading time-series distractor specification for a forecasting task.

Input:

The history_values and future_values inputs are provided in direct-prompt time-series format, where each line is `(timestamp, value)`.

history_values:
```{history_values}```

future_values:
```{future_values}```

context:
```{context}```

Goal:
Construct one distractor specification that:
- is realistic and plausible given the context,
- introduces a minimal mischaracterization of one time-series feature, expressed as a plausible but misleading empirical pattern, tendency, or historical-style behavior, without using specific timestamps, and remaining distinguishable from the true history when grounded in the data
- leads to a materially different future pattern,
- can only be identified as incorrect when grounded in the time series.


## Rules


1. Use the SAME context background. Do not change the entity and maybe the event type.
2. Do NOT introduce new causes or external mechanisms.
   - Only change how the history is characterized.
3. The distractor must NOT be easily rejected using text alone.
   - It should require time-series grounding to detect the error.
4. The distractor_history_description and distractor_future_pattern must be independent and self-contained. The distractor should arise naturally from the mischaracterized pattern alone.


## Taxonomy

Valid scope-feature combinations:

- point: value
- segment: value, trend, periodicity
- global: value, trend, periodicity


## Allowed minimal errors by feature

For each selected (scope, feature_type), the distractor must apply ONE of the following minimal modifications:

Value:
- point:
  - misstate the value at a specific time point (e.g., a wrong value, a small deviation, or a different local extremum)
- segment:
  - misstate the average level (e.g., describe as a wrong average value)
  - slightly misidentify local max or min (e.g., claim a peak is higher/lower than it is)
- global:
  - misstate the overall level (similar to segment but applied to the whole history)

Trend:
- segment:
  - misstate the local direction (e.g., describe as rising instead of flat, or flat instead of slightly decreasing)
- global:
  - misstate the overall trend (e.g., describe as increasing when it is roughly stable)

Periodicity:
- segment:
  - misstate whether periodicity is present (e.g., describe weak periodicity as absent, or wrong frequency)
- global:
  - misstate the overall periodic pattern (e.g., claim no periodicity when a weak pattern exists or wrong frequency)


## Procedure


Step 1: Inspect history_values, future_values, and context.

Step 2: Select one valid (scope, feature_type).

Step 3 : Reason about the true history pattern for that feature and how it leads to the true future pattern.
- Inside reasoning, also output suitable experience-style descriptions of the distracting pattern, such as empirical tendencies, observed patterns, or typical behaviors, rather than specific factual statements that could be directly checked against the time series.

Step 4: Write a concise ground-truth description of the history for that feature.

Step 5: Write a concise ground-truth description of the future pattern.

Step 6: Create a distractor history description by making an identifiable minimal error on ONLY the selected (scope, feature_type).
- A self-contained, plausible but misleading characterization of a different pattern narrated as a experience or potential scenario (with no specific timestamps) that would lead to a different future pattern.
- The error must be minimal: misstate that feature slightly, without changing the overall structure of the history.
- The error must be identifiable from the history itself.
- Keep all non-selected properties unchanged.
- Do not invent new patterns or exaggerate weak ones.
- The distractor's modified history description should not be stated as a false factual replacement of the true history. Instead, it should be written as a plausible but misleading description of the observed series, such as an empirical pattern, a typical tendency, or an expected mode of behavior.
- Do not describe the distractor history using specific, checkable facts about the real historical time series, such as exact timestamps, or explicit local events. These would risk creating conflicting information (a lie). Instead, express the distractor history as a plausible but misleading empirical tendency, pattern-level regularity, or conditional behavior (e.g., how the series typically behaves when values stay within a certain range or under a certain observed pattern).

Step 7: Derive a distractor future pattern from that incorrect history description.
- The distractor future must be a plausible but incorrect continuation under the same context.
- It must materially contradict the true future time series.
- When the context implies a future effect or consequence, the distractor future should also contradict that implication.
- The difference must be qualitative, not merely a small numeric deviation.


## Output format (STRICT JSON)

{
  "reasoning": "...",
  "scope": "...",
  "feature_type": "...",
  "gt_history_description": "...",
  "gt_future_pattern": "...",
  "distractor_history_description": "...",
  "distractor_future_pattern": "..."
}
\end{minted}
\end{tcolorbox}
\vspace{5pt}

\begin{tcolorbox}[title=Counterfactual-History Distractor Item, breakable]
\inputminted[fontsize=\scriptsize, breaklines, breaksymbolleft={}, breaksymbolright={}]{markdown}{Prompts/time_series_distractor_item.txt}
\end{tcolorbox}

\subsection{LLM Judge Prompts}
\label{app:prompts-judges}

\begin{tcolorbox}[title=Entity Judge, breakable]
\inputminted[fontsize=\scriptsize, breaklines, breaksymbolleft={}, breaksymbolright={}]{markdown}{Prompts/judge_entity_extraction.txt}
\end{tcolorbox}
\vspace{5pt}

\begin{tcolorbox}[title=Ground-Truth Evidence Judge, breakable]
\inputminted[fontsize=\scriptsize, breaklines, breaksymbolleft={}, breaksymbolright={}]{markdown}{Prompts/judge_gt_evidence.txt}
\end{tcolorbox}
\vspace{5pt}

\begin{tcolorbox}[title=Reasoning Chain Judge, breakable]
\inputminted[fontsize=\scriptsize, breaklines, breaksymbolleft={}, breaksymbolright={}]{markdown}{Prompts/judge_reasoning_chain.txt}
\end{tcolorbox}
\vspace{5pt}

\begin{tcolorbox}[title=Supporting Documents Judge, breakable]
\inputminted[fontsize=\scriptsize, breaklines, breaksymbolleft={}, breaksymbolright={}]{markdown}{Prompts/judge_supporting_documents.txt}
\end{tcolorbox}
\vspace{5pt}

The \texttt{\{distractor\_type\_specific\_rules\}} placeholder adds one extra rule for off-entity distractors (permitting same-scenario entity swaps) and is a no-op for the other generic types; counterfactual-history distractors use the separate prompt below.

\begin{tcolorbox}[title=Distractor Documents Judge, breakable]
\inputminted[fontsize=\scriptsize, breaklines, breaksymbolleft={}, breaksymbolright={}]{markdown}{Prompts/judge_distractor_documents.txt}
\end{tcolorbox}

\subsection{Task-Difficulty Labeling Prompts}
\label{app:prompts-difficulty}

\begin{tcolorbox}[title=Certainty, breakable]
\begin{minted}{markdown}
You are an expert evaluator of causal reasoning chains. Your task is to classify each step of a causal chain by its certainty -- the epistemic status of the information it contains: how confident one can be that the information is accurate, precise, and reliable. Certainty arises from the method of measurement or estimation and the use of probabilistic language.

Levels:
  Certain: values are directly observed or measured with negligible error margins. No probabilistic language is used.
  Likely: at least one of the following applies: the step uses probabilistic language or confidence intervals (e.g. "likely", "70% probability"); values are modelled or estimated with acknowledged but bounded uncertainty.
  Uncertain: multiple sources of uncertainty compound -- for instance, a probabilistic projection with significant measurement error. Explicit uncertainty reasoning is required, or a judgment call about how much to trust the information must be made and justified.

Output only valid JSON, no explanation.
\end{minted}
\end{tcolorbox}
\vspace{5pt}

\begin{tcolorbox}[title=Domain Knowledge, breakable]
\inputminted[fontsize=\scriptsize, breaklines, breaksymbolleft={}, breaksymbolright={}]{markdown}{Prompts/difficulty_domain_knowledge.txt}
\end{tcolorbox}
\vspace{5pt}

\begin{tcolorbox}[title=Explicitness, breakable]
\inputminted[fontsize=\scriptsize, breaklines, breaksymbolleft={}, breaksymbolright={}]{markdown}{Prompts/difficulty_explicitness.txt}
\end{tcolorbox}
\vspace{5pt}

\begin{tcolorbox}[title=Temporal Complexity, breakable]
\inputminted[fontsize=\scriptsize, breaklines, breaksymbolleft={}, breaksymbolright={}]{markdown}{Prompts/difficulty_temporal_complexity.txt}
\end{tcolorbox}

\subsection{Final Agent Audit Prompt}
\label{app:prompts-audit}

\begin{tcolorbox}[title=Final Agent Audit, breakable]
\inputminted[fontsize=\scriptsize, breaklines, breaksymbolleft={}, breaksymbolright={}]{markdown}{Prompts/coding_agent_audit.txt}
\end{tcolorbox}

\clearpage

\end{document}